\documentclass[sigconf,screen,9pt,authorversion]{acmart}

\usepackage[english]{babel}
\usepackage{blindtext}

\usepackage{booktabs}       % For formal tables
\usepackage{enumitem}
\usepackage{outlines}
\usepackage{xspace}
\usepackage{listings}
\usepackage{color}
\usepackage{colortbl}
\usepackage{graphicx}
\usepackage{amsmath, amsfonts}
\usepackage{subcaption}
\usepackage{algorithm}      % pseudocode
\usepackage{algorithmic}    % pseudocode
\usepackage{url}
\usepackage[utf8]{inputenc}
\usepackage{makecell}       % newline in table cell

\copyrightyear{2019}
\acmYear{2019}
\acmConference{SIGCOMM '19}{August 19-23, 2019}{Beijing, China}
\acmBooktitle{SIGCOMM '19, August 19-23, 2019, Beijing, China}
\acmPrice{15.00}
\acmDOI{10.1145/3341302.3342080}
\acmISBN{978-1-4503-5956-6/19/08}

\usepackage[compact, small]{titlesec}

\usepackage[subtle]{savetrees}

\usepackage{times}

\usepackage{balance}

\usepackage{dsfont}

\usepackage[labelfont=bf, textfont=rm, font=small, belowskip=0pt]{caption}

\definecolor{darkred}{rgb}{0.7,0,0}
\definecolor{darkgreen}{rgb}{0,0.5,0}
\hypersetup{colorlinks=true,
        linkcolor=darkred,
        citecolor=darkgreen}
\newcommand\code[1]{\lstinline{#1}}

\usepackage{listings}

\def\compactify{\itemsep=0pt \topsep=0pt \partopsep=0pt \parsep=0pt}
\let\latexusecounter=\usecounter

\newenvironment{CompactEnumerate}
  {\def\usecounter{\setlength{\leftmargin}{1.5em}\compactify\latexusecounter}
   \begin{enumerate}}
  {\end{enumerate}\let\usecounter=\latexusecounter}

\newcommand{\eg}{{e.g.},\xspace}
\newcommand{\ie}{{i.e.},\xspace}

\newcommand{\Graphene}{Graphene\tinystar\xspace}

\newcommand{\one}{({\em i}\/)}
\newcommand{\two}{({\em ii}\/)}
\newcommand{\three}{({\em iii}\/)}
\newcommand{\four}{({\em iv}\/)}
\newcommand{\five}{({\em v}\/)}

\newcommand{\tinystar}{\textsuperscript{$\ast$}}

\newcommand{\update}[1]{{{#1}}}

\newcommand{\para}[1]{\medskip\noindent{\bf #1}}

\newcommand{\name}{{Decima}\xspace}

\begin{document}
\title{Learning Scheduling Algorithms for Data Processing Clusters}

%\titlenote{Produces the permission block, and copyright information}
%\subtitle{Extended Abstract}

%\author{Paper \#306, 12+7 pages}
% \author{Firstname Lastname}
% \authornote{Note}
% \orcid{1234-5678-9012}
% \affiliation{%
%   \institution{Affiliation}
%   \streetaddress{Address}
%   \city{City}
%   \state{State}
%   \postcode{Zipcode}
% }
% \email{email@domain.com}

\author{\fontfamily{phv}\selectfont\Large
	\mbox{Hongzi Mao, Malte Schwarzkopf, Shaileshh Bojja Venkatakrishnan, Zili Meng\textsuperscript{$\star$}, Mohammad Alizadeh}\\ \smallskip
  MIT Computer Science and Artificial Intelligence Laboratory\; \textsuperscript{$\star$}Tsinghua University\\ \smallskip
}
\email{{hongzi,malte,bjjvnkt,alizadeh}@csail.mit.edu, mengzl15@mails.tsinghua.edu.cn}
\newcommand\authorlist{Hongzi Mao, Malte Schwarzkopf, Shaileshh Bojja Venkatakrishnan, Zili Meng and Mohammad Alizadeh}

%\author{\normalsize{\textsf{Hongzi Mao, Malte Schwarzkopf, Shaileshh Bojja Venkatakrishnan, Zili Meng, Mohammad Alizadeh}}\\
%\vspace{-0.05in}
%\normalsize{\textsf{MIT Computer Science and Artificial Intelligence Laboratory}}\\
%\normalsize{\textmd{\{hongzi,\,malte,\,bjjvnkt,\,zili,\,alizadeh\}@csail.mit.edu}}}

%The default list of authors is too long for headers}
\renewcommand{\shortauthors}{H. Mao et al.}

% \noindent
% \textbf{Abstract.}
\begin{abstract}
Efficiently scheduling data processing jobs on distributed compute
clusters requires complex algorithms.
Current systems use simple, generalized heuristics and ignore
workload characteristics, since developing and tuning a scheduling policy for
each workload is infeasible.
In this paper, we show that modern machine learning techniques can generate
highly-efficient policies automatically.
\name uses reinforcement learning (RL) and neural networks to learn
workload-specific scheduling algorithms without any human instruction
beyond a high-level objective, such as minimizing average job completion
time.
However, off-the-shelf RL techniques cannot handle the complexity and
scale of the scheduling problem.
To build \name, we had to develop new representations for jobs' dependency
graphs, design scalable RL models, and invent RL training
methods for dealing with continuous stochastic job arrivals.

Our prototype integration with Spark on a 25-node cluster shows that \name
improves average job completion time by at least 21\% over hand-tuned
scheduling heuristics, achieving up to 2$\times$ improvement during periods of
high cluster load.
%
%Our prototype integration with Spark on a 25-node cluster shows that \name's
%policies are particularly effective during periods of high cluster load,
%improving average JCT by up to 2$\times$ over state-of-the-art heuristics.
%
%outperforms several heuristics, including hand-tuned ones, by at least 21\%.
%
%Further experiments with an industrial production workload trace demonstrate
%that \name
%can execute the same workload with 28\% fewer machines
%delivers up to a 17\% reduction in average job completion time
%and scales to large clusters.
%

{\small


%\smallskip\noindent{\bf ACM Reference format:} Hongzi Mao, Malte Schwarzkopf, Shaileshh Bojja Venkatakrishnan, Zili Meng, Mohammad \mbox{Alizadeh} MIT Computer Science and Artificial Intelligence Laboratory, Tsinghua University. 2019. Learning Scheduling Algorithms for Data Processing Clusters. In \textit{Proceedings of SIGCOMM '19, August 19-23, 2019, Beijing, China}, 19 pags.

%\noindent
%DOI: https://dx.doi.org/00.0000/0000000.0000000
}
\end{abstract}

\maketitle

\section{Introduction}
\label{s:intro}

Efficient utilization of expensive compute clusters matters for
enterprises: even small improvements in utilization can save
millions of dollars at scale~\cite[\S1.2]{wsc-book-2ed}.
Cluster schedulers are key to realizing these savings.
A good scheduling policy packs work tightly to reduce
fragmentation~\cite{graphene, tetris, job-packing-google}, prioritizes jobs
according to high-level metrics such as user-perceived
latency~\cite{borg}, and avoids inefficient configurations~\cite{jockey}.
Current cluster schedulers rely on heuristics that prioritize generality,
ease of understanding, and straightforward implementation over achieving the
ideal performance on a specific workload.
By using general heuristics like fair scheduling~\cite{hadoop-fair-sched, drf},
shortest-job-first, and simple packing strategies~\cite{tetris},
current systems forego potential performance optimizations.
For example, widely-used schedulers ignore readily available information
about job structure (\ie internal dependencies) and efficient parallelism
for jobs' input sizes.
Unfortunately, workload-specific scheduling policies that use this
information require expert knowledge and significant effort to devise,
implement, and validate.
For many organizations, these skills are either unavailable, or uneconomic
as the labor cost exceeds potential savings.
In this paper, we show that modern machine-learning techniques can help side-step this trade-off by \emph{automatically learning} highly efficient,
workload-specific scheduling policies.
We present \name
\footnote{In Roman mythology, \name measures threads of life and decides their destinies.},
a general-purpose scheduling service for data processing jobs with dependent
stages.
Many systems encode job stages
and their dependencies as directed acyclic graphs
(DAGs)~\cite{spark, tez, dryad, flumejava}. Efficiently scheduling
DAGs leads to hard algorithmic problems whose optimal solutions
 are intractable~\cite{graphene}.
%
% GAINS
%
Given only a high-level goal (\eg minimize average job completion time),
\name uses existing monitoring information and past workload logs to
automatically learn sophisticated scheduling policies.
For example, instead of a rigid fair sharing policy, \name learns to give  jobs different
shares of resources to optimize overall performance,
and it learns job-specific parallelism levels that avoid wasting resources on
diminishing returns for jobs with little inherent parallelism.
%
%\ms{Do we ever prove the choke point and plan-ahead claim?}
%
The right algorithms and thresholds for these policies are workload-dependent,
and achieving them today requires painstaking manual scheduler customization.
%

%it learns to use jobs' dependency structure to plan ahead and avoid
%waiting at ``choke points'';

% %
% % Malte: I don't think below matters here
% We focus on these jobs for two reasons: \one{} many systems encode job stages
% and their dependencies as directed acyclic graphs
% (DAGs)~\cite{spark, tez, dryad, flumejava}; and \two{} scheduling
% DAGs is a hard algorithmic problem whose optimal solutions are intractable
% and difficult to capture in good heuristics~\cite{graphene}.
% %
% % GAINS
% %
% Given only a high-level goal (\eg minimize average job completion time),
% \name uses existing monitoring information and past workload logs to
% automatically learn sophisticated scheduling policies.
% %
% For example, instead of a rigid fair sharing policy, \name learns to give jobs
% shares of resources that optimize overall performance;
% it learns to use jobs' dependency structure to plan ahead and avoid
% waiting at ``choke points'';
% and it learns job-specific parallelism levels that avoid wasting resources on
% diminishing returns for jobs with little inherent parallelism.
% %
% %\ms{Do we ever prove the choke point and plan-ahead claim?}
% %
% The right algorithms and thresholds for these policies are workload-dependent,
% and achieving them today requires painstaking manual scheduler customization.
% %

%
\name learns scheduling policies through experience using modern
reinforcement learning (RL) techniques.
RL is well-suited to learning scheduling policies because
it allows learning from actual workload and operating conditions
without relying on inaccurate assumptions.
\name encodes its scheduling policy in a neural network trained via a large number
of simulated experiments, during which it schedules a workload, observes the
outcome, and gradually improves its policy.
However, \name's contribution goes beyond merely applying off-the-shelf RL algorithms to
scheduling: to successfully learn high-quality scheduling policies, we had to develop
novel data and scheduling action representations, and new RL training techniques.

%, and scalability
%insights applied to RL.
%
%We built \name using neural networks and RL because these techniques have
%achieved remarkable recent success on challenging decision-making tasks, such
%as learning Go and Chess purely through self-play~\cite{alphagozero}.
%
%To apply these techniques to complex cluster scheduling problems, however, we
%had to solve several key problems.
%

First, cluster schedulers must scale to hundreds of jobs and thousands of machines, and must decide among potentially hundreds of configurations per job (\eg different levels of parallelism).
%
%Second, cluster schedulers must scale to many jobs and machines, and decide among
%hundreds of configurations per job (\eg different levels of parallelism).
%
This leads to much larger problem sizes compared to conventional RL applications (e.g., game-playing~\cite{atari, alphago}, robotics control~\cite{ddpg, trpo}), both in the amount of information available to the scheduler (the
\emph{state space}), and the number of possible choices it must consider (the \emph{action
space}).\footnote{For example, the state of the game of Go~\cite{alphagozero} can be represented by
$19\times19 = 361$ numbers, which also bound the number of legal moves per turn.}
We designed a scalable neural network architecture that combines a {\em graph neural network}~\cite{graphspectral, graphcombopt, graphcnn, gcnn_google} to process job and cluster information without manual feature engineering, and a {\em policy network} that makes scheduling decisions. Our neural networks reuse a small set of building block operations to process job DAGs, irrespective of their sizes and shapes, and to make scheduling decisions, irrespective of the number of jobs or machines. These operations are parameterized functions learned during training, and designed for the scheduling domain\,---\,\eg ensuring that the graph neural network can express properties such as a DAG's critical path. Our neural network design substantially reduces model complexity compared to naive encodings of the scheduling problem, which is key to efficient learning, fast training, and low-latency scheduling decisions.

Second, conventional RL algorithms cannot train models with continuous streaming
job arrivals.
The randomness of job arrivals can make it impossible for RL algorithms to tell
whether the observed outcome of two decisions differs due to disparate job arrival
patterns, or due to the quality the policy's decisions.
Further, RL policies necessarily make poor decisions in early stages of training.
Hence, with an unbounded stream of incoming jobs, the policy inevitably accumulates
a backlog of jobs from which it can never recover.
Spending significant training time exploring actions in such situations fails to
improve the policy.
To deal with the latter problem, we terminate training ``episodes'' early in the
beginning, and gradually grow their length.
This allows the policy to learn to handle simple, short job sequences first, and
to then graduate to more challenging arrival sequences.
%Hence, the policy learns to handle simple, short job sequences first, and
%then graduates to more challenging arrival sequences.
%
%This is due to the randomness of job arrivals in streaming, which causes estimating
%the true quality of scheduling policies to be hard for the learning algorithm.
%Consequently training suffers.
%
To cope with the randomness of job arrivals, we condition training feedback on the
actual sequence of job arrivals experienced, using a recent technique for RL in
environments with stochastic inputs~\cite{variance-reduction}.
This isolates the contribution of the scheduling policy in the feedback and makes it
feasible to learn policies that handle stochastic job arrivals.
We integrated \name with Spark and evaluated it in both an experimental testbed
and on a workload trace from \update{Alibaba's production
clusters}~\cite{alibaba-data, alibaba_data_analysis}.\footnote{We used an earlier
version of Alibaba's public \texttt{cluster-trace-v2018} trace.}
Our evaluation shows that \name outperforms existing heuristics
on a 25-node Spark cluster, reducing average job completion time of TPC-H query
mixes by at least 21\%. \name's
policies are particularly effective during periods of high cluster load,
where it improves the job completion time by up to 2$\times$ over existing
heuristics.
\name also extends to multi-resource scheduling of CPU and memory, where it improves
average job completion time by 32-43\% over prior schemes
such as Graphene~\cite{graphene}.

%compared to the state-of-the-art
%algorithms behind Graphene~\cite{graphene}.
%
%; and \three{} using the same trace, \name's core learning framework
%can easily adapt to packing resources in multiple dimensions (CPU and RAM), achieving a
%37\% improvement over meticulously designed heuristics such as
%Graphene's~\cite{graphene}.
%
%Importantly, while the relative gains may appear modest, \name delivers these
%efficiency gains automatically, without burdening cluster operators with
%expensive and brittle hand-tuning of the scheduler.
%
%We also illustrate that our learning innovations are important for \name to
%successfully learn cluster scheduling policies.
%

%
% CONTRIBUTIONS
%
\smallskip
\noindent
In summary, we make the following key contributions:
\begin{CompactEnumerate}
\item A scalable neural network design that can process DAGs of
  arbitrary shapes and sizes, schedule DAG stages, and set
  efficient parallelism levels for each job
  (\S\ref{s:graph}--\S\ref{s:action}).
\item A set of RL training techniques that for the first time enable training a
  scheduler to handle unbounded stochastic job arrival
  sequences~(\S\ref{s:training}).
\item \name, the first RL-based scheduler that schedules complex data processing
  jobs and learns workload-specific scheduling policies without human
  input, and a prototype implementation of it (\S\ref{s:impl}).
\item An evaluation of \name in simulation and in a real Spark cluster,
  and a comparison with state-of-the-art scheduling heuristics (\S\ref{s:eval}).
%
% \item \name's RL design, its representations of cluster state, output actions,
%   and reward signals suitable for learning scheduling
%   policies without human-encoded input (\S\ref{s:design}).
% %
% \item A novel state embedding that converts job DAGs into vectorized inputs
%   suitable for neural networks (\S\ref{s:graph}).
% %
% \item Techniques for scalable RL models that learn complex scheduling
%   decisions, both respecting DAG structure and choosing levels of parallelism
%   (\S\ref{s:v1}--\ref{s:v2}).
% %
% \item A prototype Spark integration of \name, and an training environment that
%   faithfully simulates Spark clusters (\S\ref{s:impl}).
% %
% \item An evaluation of \name on both mixes of TPC-H queries and production job
%   trace from a batch processing cluster at a large company
%   (\S\ref{s:eval}).
% %
\end{CompactEnumerate}
%

%
%\noindent
%\textbf{Ethics statement:} This work raises no ethical issues.
%

\section{Motivation}
\label{s:motivation}

%\begin{figure}
%\centering
%\includegraphics[width=0.3\textwidth]{figures/motivation/motivation_visualization/motivation_visualization.pdf}
%\caption{The average job completion time for 10 randomly sampled TPC-H
%  queries running on a Spark cluster with 50 task slots improves by 69\% over
%  the default FIFO scheduler (\subref{f:motiv-fifo}), and by 7\% over a fair
%  scheduler (\subref{f:motiv-fair}) when taking graph structure into account
%  (\subref{f:motiv-v1}); when additionally setting the right levels of
%  parallelism, the benefit increases to 81\% and 16\%, respectively
%  (\subref{f:motiv-v2}). Different colors correspond to different queries,
%  and vertical red lines indicate job completions.}
%\label{f:motivation-comparison}
%\end{figure}

\begin{figure}
\centering
\includegraphics[height=3.5cm]{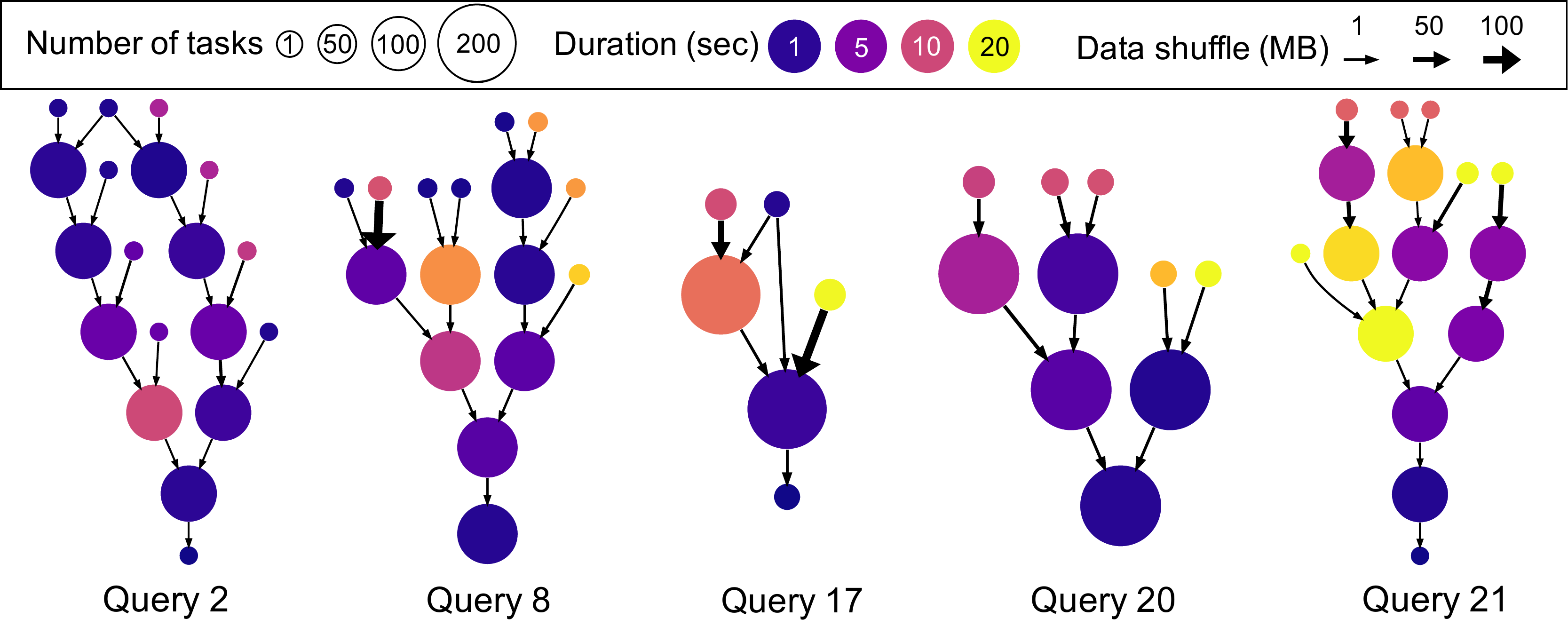}
\vspace{-0.7cm}
\caption{\small Data-parallel jobs have complex data-flow graphs like the ones
  shown (TPC-H queries in Spark), with each node having a distinct number of
  tasks, task durations, and input/output sizes.}
\label{f:tpch-viz}
\vspace{-0.4cm}
\end{figure}

%
%Our goal is to build an efficient cluster scheduler that
%automatically learns to exploit knowledge of
%job structure and workload characteristics to optimize job performance.
%
%We target \emph{batch jobs}, \ie jobs that execute a finite computation to
%completion.
%%
%Batch jobs are an important workload: they make up 80--90\% of the jobs
%and tasks (and, consequently, scheduling decisions) in Google
%clusters~\cite[\S2]{omega}, and other companies run extensive
%infrastructure dedicated to batch data analytics~\cite{scope, graphene}.
%
Data processing systems and query compilers such as Hive, Pig, SparkSQL, and
DryadLINQ create \emph{DAG-structured} jobs, which consist of processing
stages connected by input/output dependencies (Figure~\ref{f:tpch-viz}).
For recurring jobs, which are common in production
clusters~\cite{rope}, reasonable estimates of
runtimes and intermediate data sizes may be available.
Most cluster schedulers, however, ignore this job structure in their
decisions and rely on \eg coarse-grained fair
sharing~\cite{hadoop-fair-sched, drf, h-drf, choosy}, rigid priority
levels~\cite{borg}, and manual specification of each job's parallelism~\cite[\S5]{omega}.
Existing schedulers choose to largely ignore this rich, easily-available job structure
information because it is difficult to design scheduling algorithms that
make use of it.
%
%We illustrate the benefit of using this job-specific information in
%scheduling decisions with two techniques it enables: \one{}
%dependency-aware scheduling, and \two{} automatically choosing the
%right level of parallelism.
%
We illustrate the challenges of using
job-specific information in scheduling decisions with two concrete
examples: \one{} dependency-aware scheduling, and \two{} automatically
choosing the right number of parallel tasks.

%\begin{figure}
%\centering
%\includegraphics[height=0.26\textwidth]{figures/motivation/motivation.pdf}
%\vspace{-0.4cm}
%\caption{By planning ahead, the optimal, DAG-aware schedule avoids the ``choke
%  point'' that occurs for the green node using the critical path heuristic,
%  and improves completion time by 29\%.}
%\label{f:dag-ex}
%\end{figure}
%
\subsection{Dependency-aware task scheduling}
\label{s:motiv-structure}
Many job DAGs in practice have tens or hundreds of stages with different durations
and numbers of parallel tasks in a complex dependency structure.
An ideal schedule ensures that independent stages run in parallel as much as
possible, and that no stage ever blocks on a dependency if there are
available resources.
Ensuring this requires the scheduler to understand the dependency structure
and plan ahead.
This ``DAG scheduling problem'' is algorithmically hard: see, \eg the illustrative example
by Grandl et al.~\cite[\S2.2]{graphene} and the one we describe in
detail in Appendix~\ref{s:graph-hard}.
Theoretical research~\cite{shmoys1994improved, blumofe1999scheduling, chekuri2004multi, leighton1988universal} has focused mostly on simple instances of the problem that do not capture the complexity of real data processing clusters (e.g., online job arrivals, multiple DAGs, multiple tasks per stage, jobs with different inherent parallelism, overheads for moving jobs between machines, etc.). For example, in a recent paper, Agrawal et al.~\cite{dag_schedule_soda} showed that two simple DAG scheduling policies (shortest-job-first and latest-arrival-processor-sharing) have constant competitive ratio in a basic model with one task per job stage. As our results show (\S\ref{s:motivate_visualization}, \S\ref{s:eval}), these policies are far from optimal in a real Spark cluster.    

Hence, designing an algorithm to generate optimal schedules
for all possible DAG combinations is
intractable~\cite{acyclic-jobs-hard, graphene}.
Existing schedulers ignore this challenge: they enqueue tasks
from a stage as soon as it becomes available, or run stages in an arbitrary order.
%

%%%%%%%%%%%%%%%%%%%%%%%%%%%%%%%%%%%%%%%%%%%%%%%%%%%%%%%%%%%%%%%%%%%%%%%%%%%%%%%%%%%

\subsection{Setting the right level of parallelism}
\label{s:motiv-parallelism}
\begin{figure}
 \centering
 \includegraphics[width=0.9\columnwidth]{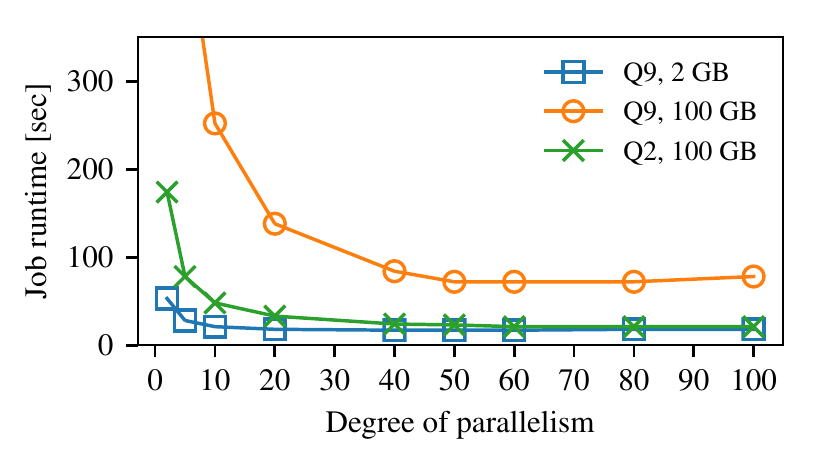}
 \vspace{-0.5cm}
 \caption{\small TPC-H queries scale differently with parallelism: Q9 on a
   100 GB input sees speedups up to 40 parallel tasks, while Q2 stops
   gaining at 20 tasks; Q9 on a 2 GB input needs only 5 tasks.
   Picking ``sweet spots'' on these curves for a mixed workload is
   difficult.}
 \label{f:motivation-parallelism}
 \vspace{-0.2cm}
\end{figure}
\begin{figure*}[t]
\centering
\begin{subfigure}[t]{0.245\textwidth}
  \centering
  \includegraphics[width=\textwidth]{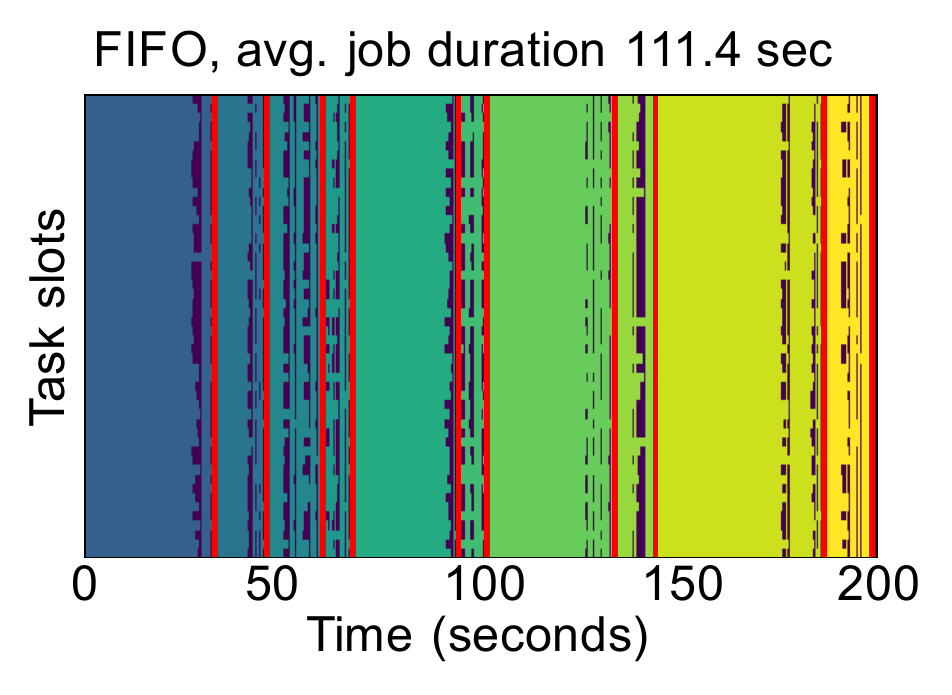}
  \vspace{-0.6cm}
  \caption{FIFO scheduling.}
  \label{f:motiv-fifo}
\end{subfigure}
\begin{subfigure}[t]{0.245\textwidth}
  \centering
  \includegraphics[width=\textwidth]{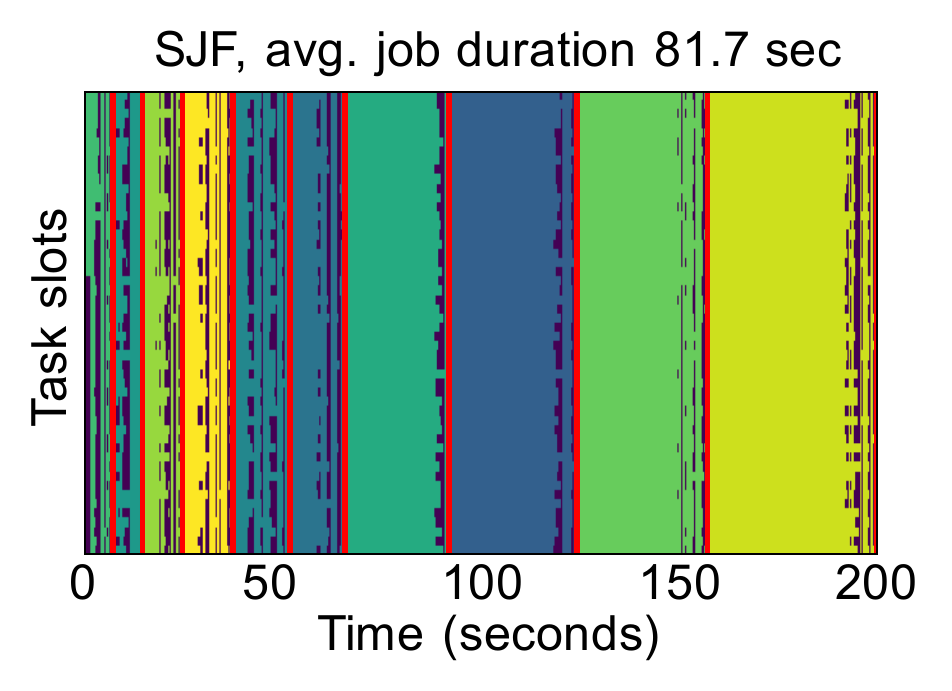}
  \vspace{-0.6cm}
  \caption{SJF scheduling.}
  \label{f:motiv-sjf}
\end{subfigure}
\begin{subfigure}[t]{0.245\textwidth}
  \centering
  \includegraphics[width=\textwidth]{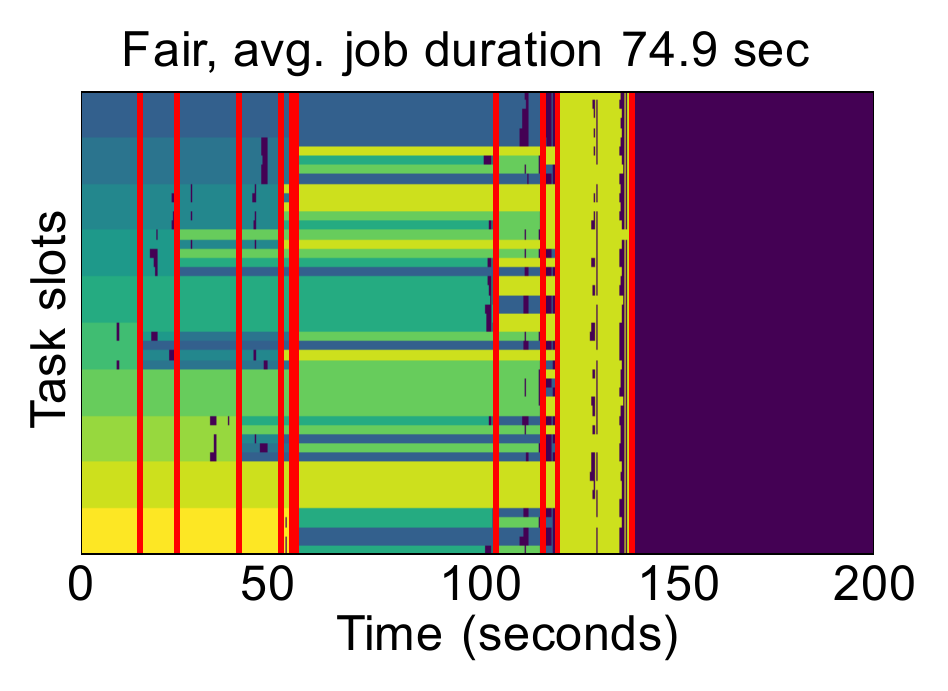}
  \vspace{-0.6cm}
  \caption{Fair scheduling.}
  \label{f:motiv-fair}
\end{subfigure}
\begin{subfigure}[t]{0.245\textwidth}
  \centering
  \includegraphics[width=\textwidth]{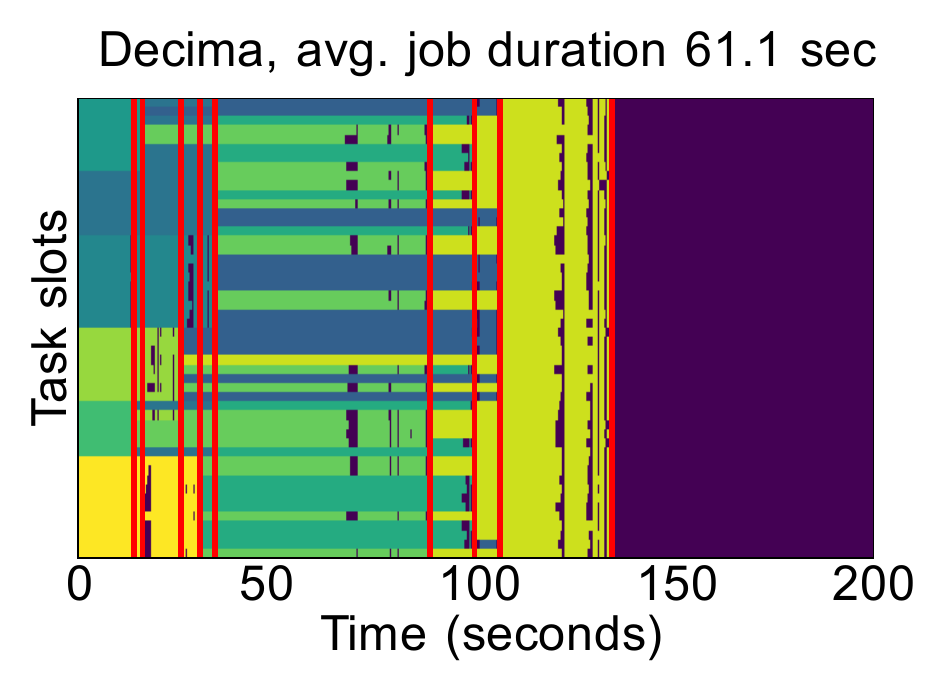}
  \vspace{-0.6cm}
  \caption{\name.}
  % with graph structure and parallelism level setting.}
  \label{f:motiv-v2}
\end{subfigure}
\vspace{-0.3cm}
\caption{Decima improves average JCT of 10 random TPC-H queries by 45\% over
  Spark's FIFO scheduler, and by 19\% over a fair scheduler on a cluster
  with 50 task slots (executors). Different queries in different colors;
  vertical red lines are job completions; purple means idle.}
\label{f:motivation-comparison}
%\vspace{-0.25cm}
\end{figure*}
In addition to understanding dependencies, an ideal scheduler must also
understand how to best split limited resources among jobs.
Jobs vary in the amount of data that they process, and in the amount of parallel
work available.
% when different stages run.
%
A job with large input or large intermediate data can efficiently harness
additional parallelism; by contrast, a job running on small input data, or one
with less efficiently parallelizable operations, sees diminishing
returns beyond modest parallelism.
Figure~\ref{f:motivation-parallelism} illustrates this with the job runtime
of two TPC-H~\cite{tpch} queries running on Spark as they are given additional
resources to run more parallel tasks.
Even when both process 100 GB of input, Q2 and Q9 exhibit widely different
scalability: Q9 sees significant speedup up to 40 parallel tasks, while Q2
only obtains marginal returns beyond 20 tasks.
When Q9 runs on a smaller input of 2 GB, however, it needs no more than ten
parallel tasks.
For all jobs, assigning additional parallel tasks beyond a ``sweet spot'' in
the curve adds only diminishing gains.
Hence, the scheduler should reason about which job will see the largest
marginal gain from extra resources and accordingly pick the sweet spot for each
job.
%
%In principle, this is a straightforward application of Amdahl's
%Law~\cite{amdahl}, but given a mixed workload, the specific curves and sweet
%spots are difficult to predict.
%\hongzi{Given a mixed workload, the specific curve for the diminishing gains are job-dependent and difficult to model since it involves batching and pipelining effects, hard-to-parallelize stages, and intermediate data.}
%

%
Existing schedulers largely side-step this problem.
Most burden the user with the choice of how many parallel tasks to
use~\cite[\S5]{omega}, or rely on a separate ``auto-scaling'' component based on
coarse heuristics~\cite{jockey, spark-dynamic-scaling}.
Indeed, many fair schedulers~\cite{quincy, drf} divide resources without
paying attention to their decisions' efficiency: sometimes, an ``unfair''
schedule results in a more efficient overall execution.

\subsection{An illustrative example on Spark}
\label{s:motivate_visualization}
The aspects described are just two examples of how schedulers can exploit
knowledge of the workload.
To achieve the best performance, schedulers must also respect other
considerations, such as the execution order (\eg favoring short jobs) and
avoiding resource fragmentation~\cite{borg, tetris}.
Considering all these dimensions together\,---\,as \name does\,---\,makes
a substantial difference.
We illustrate this by running a mix of ten randomly chosen
TPC-H~\cite{tpch} queries with input sizes drawn from a long-tailed
distribution on a Spark cluster with 50 parallel task
slots.\footnote{See \S\ref{s:eval} for details of the workload and our
cluster setup.}
Figure~\ref{f:motivation-comparison} visualizes the schedules imposed by
\emph{(\subref{f:motiv-fifo})} Spark's default FIFO scheduling;
\emph{(\subref{f:motiv-sjf})} a shortest-job-first (SJF) policy that strictly
prioritizes short jobs;
\emph{(\subref{f:motiv-fair})} a more realistic, fair scheduler that
dynamically divides task slots between jobs; and
\emph{(\subref{f:motiv-v2})} a scheduling policy learned by \name.
We measure average job completion time (JCT) over the ten jobs.
Having access to the graph structure helps \name improve average JCT by 45\% over
the naive FIFO scheduler, and by 19\% over the fair scheduler.
It achieves this speedup by completing short jobs quickly, as five jobs
finish in the first 40 seconds; and by maximizing parallel-processing
efficiency.
SJF dedicates all task slots to the next-smallest job in order to
finish it early (but inefficiently); by contrast, \name runs jobs near their
parallelism sweet spot.
By controlling parallelism, \name reduces the total time to complete all jobs by
30\% compared to SJF.
Further, unlike fair scheduling, \name partitions task slots non-uniformly across
jobs, improving average JCT by 19\%.
Designing general-purpose heuristics to achieve these benefits is difficult,
as each additional dimension (DAG structure, parallelism, job sizes, etc.)
increases complexity and introduces new edge cases.
%
%It may be feasible to tune a heuristic for a specific workload, but rarely
%happens in practice, since devising, implementing, and testing a scheduling policy
%requires expert knowledge and significant effort.
%
%Hence, the costs of expensive human labor quickly outweigh the performance benefits.
%
\name opens up a new option: using data-driven techniques, it \emph{automatically}
learns workload-specific policies that can reap these gains.
\name does so without requiring human guidance beyond a high-level goal (\eg
minimal average JCT), and without explicitly modeling the system or the workload.

\section{The DAG Scheduling Problem in Spark}
\label{s:context}

\name is a general framework for learning scheduling algorithms for
DAG-structured jobs.
For concreteness, we describe its design in the context of the Spark
system.
%
%This section briefly overviews Spark's job scheduling architecture.
%We later describe how to apply \name to other variants of graph-structured job scheduling problems.

%
A Spark job consists of a DAG whose nodes are the execution \emph{stages} of the job.
Each stage represents an operation that the system runs in parallel over
many shards of the stage's input data.
The inputs are the outputs of one or more parent stages, and each shard
is processed by a single \emph{task}.
A stage's tasks become runnable as soon as all parent stages have completed.
How many tasks can run in parallel depends on the number of \emph{executors}
 that the job holds.
Usually, a stage has more tasks than there are executors, and the tasks
therefore run in several ``waves''.
Executors are assigned by the Spark master based on user requests, and by
default stick to jobs until they finish.
However, Spark also supports dynamic allocation of executors based on
the wait time of pending tasks~\cite{spark-dynamic-scaling}, although moving
executors between jobs incurs some overhead (e.g., to tear down and launch
JVMs).
%\ma{The description of dynamic resource allocation isn't very clear. Does wait time refer to the time that tasks have been waiting for executors?}
%

%
Spark must therefore handle three kinds of scheduling decisions: \one{} deciding
how many executors to give to each job; \two{} deciding which stages' tasks to
run next for each job, and \three{} deciding which task to run next when an executor becomes
idle.
When a stage completes, its job's \emph{DAG scheduler} handles
the activation of dependent child stages and enqueues their tasks with a
lower-level \emph{task scheduler}.
The task scheduler maintains task queues from which it assigns a task every
time an executor becomes idle.
We allow the scheduler to move executors between job DAGs as it sees fit
(dynamic allocation).
\name focuses on DAG scheduling (\ie which stage to run next) and executor
allocation (\ie each job's degree of parallelism).
Since tasks in a stage run identical code and request identical resources,
we use Spark's existing task-level scheduling.

\section{Overview and Design Challenges}
\label{s:design-overview}

\name represents the scheduler as an agent that uses a neural network to make decisions, henceforth referred to as the {\em policy network}. On {\em scheduling events}\,---\,\eg a stage completion (which frees up executors), or a job arrival (which adds a DAG)\,---\,the agent takes as input the current {\em state} of the cluster and outputs a scheduling {\em action}. At a high level, the state captures the status of the DAGs in the scheduler's queue and the executors, while the actions determine which DAG stages executors work on at any given time.

\name trains its neural network using RL through a large number of offline (simulated) experiments. In these experiments, \name attempts to schedule a workload, observes the outcome, and provides the agent with a {\em reward} after each action. The reward is set based on \name's high-level scheduling objective (e.g., minimize average JCT). The RL algorithm uses this reward signal to gradually improve the scheduling policy.
Appendix~\ref{s:background} provides a brief primer on RL.

%, and uses an RL algorithm to gradually improve the scheduling policy. To guide the RL algorithm, \name gives the agent a {\em reward} after each action based on its high-level scheduling objective. The RL algorithm uses this reward signal to

\begin{figure}[t]
\centering
\vspace{-0.4cm}
\includegraphics[height=3.2cm]{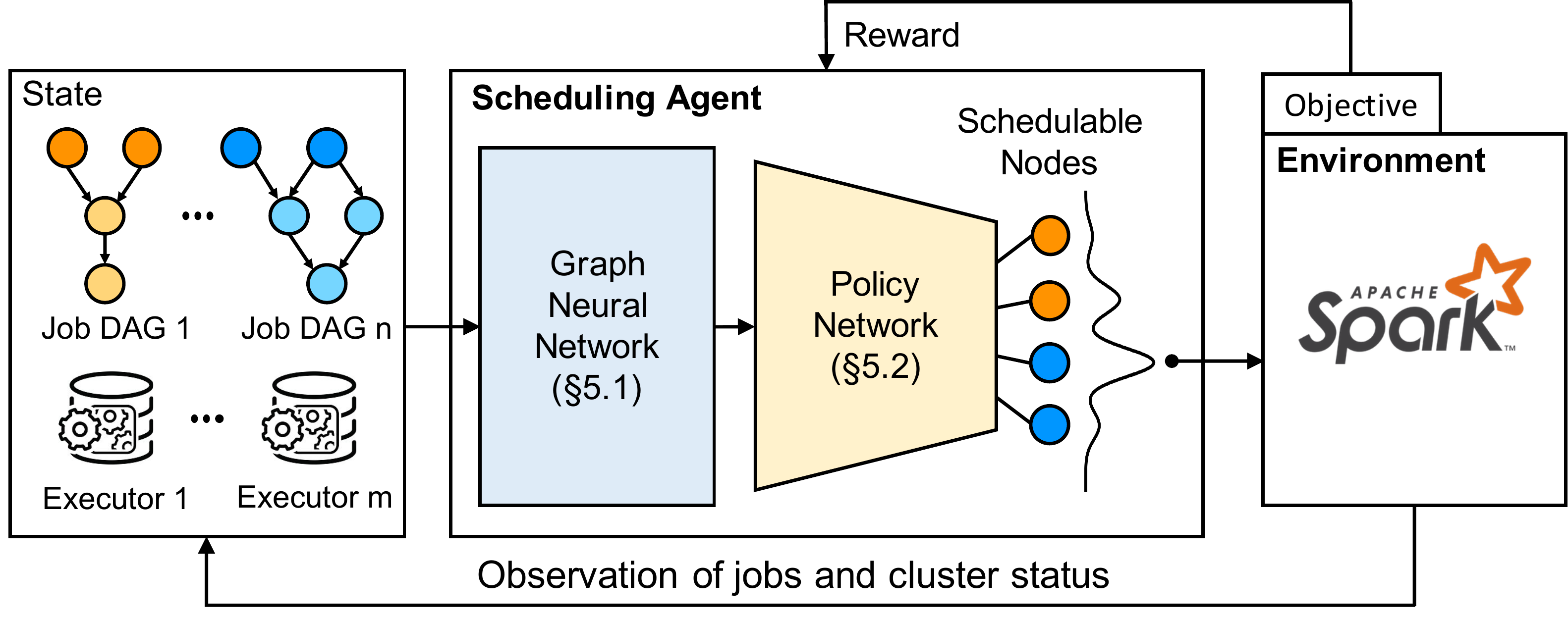}
\vspace{-0.3cm}
\caption{In \name's RL framework, a \emph{scheduling agent} observes the
  \emph{cluster state} to decide a scheduling \emph{action}
  on the cluster \emph{environment}, and receives a \emph{reward} based
  on a high-level objective. The agent uses a \emph{graph neural network} to
  turn job DAGs into vectors for the \emph{policy network}, which outputs actions.}
\vspace{-0.3cm}
\label{fig:rl}
\end{figure}

\name's RL framework (Figure~\ref{fig:rl}) is general and it can be applied to a variety of systems and objectives. In \S\ref{s:design}, we describe the design for scheduling DAGs on a set of identical executors to minimize average JCT. Our results in \S\ref{s:eval} will show how to apply the same design to schedule multiple resources (e.g., CPU and memory), optimize for other objectives like makespan~\cite{makespan}, and learn qualitatively different polices depending on the underlying system (e.g., with different overheads for moving jobs across machines).

% %
% In complex Spark scheduling environments~(\S\ref{s:context}), reinforcement
% learning (RL) is a natural alternative approach to avoid tedious hand-crafted
% heuristic designs.
% %
% At a high level, RL benefits from model-free optimization schemes to automatically
% learn a good scheduling policy from direct interactions with the cluster
% environment~\cite{deeprm, pensieve, alphagozero, doom-rl, rlbook}.
% %
% In our RL framework, the cluster forms the environment, and its state contains
% both job DAGs and the current status of the executors.
% %
% The agent observes the state and picks a scheduling action at scheduling events
% that modify the state\,---\,\eg a stage completion (which frees up executors),
% or a job arrival (which adds a DAG).
% %
% As feedback, the agent receives a reward based on a high-level objective.
% %
% For example, if the objective is to minimize average JCT, \name penalizes the
% agent $-\tau \times J$ at each time step, where $\tau$ is the absolute time (in
% seconds) since last action and $J$ is the number of jobs in the
% system~\cite{deeprm}.
% %
% For a set of jobs, these penalties add up to the sum of the job completion times;
% thus the agent learns to minimize average JCT.

\smallskip
\noindent{\bf Challenges.}
\name's design tackles three key challenges:

\begin{CompactEnumerate}
\item {\bf Scalable state information processing.}
The scheduler must consider a large amount of dynamic information to make scheduling decisions: hundreds of job DAGs, each with dozens of stages, and executors that may each be in a different state (\eg assigned to different jobs).
Processing all of this information via neural networks is challenging, particularly because neural networks usually require fixed-sized vectors as inputs.

%\name uses a novel {\em graph embedding} to represent job DAGs as vectors~(\S\ref{s:graph}). The embedding is based on graph convolutional neural networks~\cite{graphcnn}, but it is customized for  scheduling. The key ingredient is a parametrized non-linear aggregation operation that is performed iteratively on the DAG to derive an embedding for each node based on the embeddings of its children. By reusing aggregation operations within and across DAGs, \name's embedding scales to an arbitrary number of job DAGs with arbitrary sizes and shapes.

\item {\bf Huge space of scheduling decisions.}
The scheduler must map potentially thousands of runnable stages to available executors. The exponentially large space of mappings poses a challenge for RL algorithms, which must ``explore'' the action space in training to learn a good policy.

%Indeed, a naive encoding of scheduling decisions can easily lead to an exceedingly large action space, or require long sequence of actions to schedule jobs, both of which make RL training difficult~\cite{a3c, rlbook}.

%\name addresses this challenge by breaking scheduling decisions into sequences of two-dimensional actions~(\S\ref{s:action}), which output (1) a stage designated to be scheduled next and, (2) a cap on the maximum allowed parallelism for the designated job. We show that decomposing scheduling decisions in this way significantly reduces the learning and inference complexity.
%

\item {\bf Training for continuous stochastic job arrivals}.
%Jobs arrive to the scheduler over time.
It is important to train the scheduler to handle continuous randomly-arriving jobs over time. However, training with a continuous job arrival process is non-trivial because RL algorithms typically require training ``episodes'' with a finite time horizon. Further, we find that randomness in the job arrival process creates difficulties for RL training due to the variance and noise it adds to the reward.

%randomness in the job arrival pattern can lead to large performance variations that make RL training difficult. For example, a period of high load is likely to increase JCT, reducing the agent's rewards compared to a low-load period. But such variations in reward have little to do with scheduling decisions, and therefore can create significant noise for RL algorithms that seek to identify good actions based on the rewards.

% in reward have little to do with scheduling decisions; they're due to the load on the system. We find that the noise introduced by the job arrival process can make learning impossible for standard RL algorithms.

%Inspired by recent RL techniques for ``input-driven'' environments~\cite{variance-reduction}, we use an {\em input-dependent baseline} to condition the reward signal on the actual sequences of job arrivals during training. Input-dependent baselines help account for the variance caused by the arrival process and enable RL training with stochastic job arrivals.

\end{CompactEnumerate}

%%%%%%%%%%%%%%%%%%%%%%%%%%%%%%%%%%%%%%%%%%%%%%%%%%%%%%%%%%%%%%%%%%%%%%%%%%%%%%%%%%%%%%%%%%%%%%%%
%%%%%%%%%%%%%%%%%%%%%%%%%%%%%%%%%%%%%%%%%%%%%%%%%%%%%%%%%%%%%%%%%%%%%%%%%%%%%%%%%%%%%%%%%%%%%%%%
%%%%%%%%%%%%%%%%%%%%%%%%%%%%%%%%%%%%%%%%%%%%%%%%%%%%%%%%%%%%%%%%%%%%%%%%%%%%%%%%%%%%%%%%%%%%%%%%

\section{Design}
\label{s:design}

This section describes \name's design, structured according to how it addresses the three aforementioned challenges: scalable processing of the state information (\S\ref{s:graph}), efficiently encoding scheduling decisions as actions (\S\ref{s:action}), and RL training with continuous stochastic job arrivals (\S\ref{s:training}).

%The following sections describe how \name's design addresses these challenges in detail.

\subsection{Scalable state information processing}
\label{s:graph}

%
% Neural networks cannot directly process the graph-structured job DAGs
% in the state, as they expect flat, numeric vector inputs.
%

%\name uses a novel {\em graph embedding} to represent job DAGs as vectors~(\S\ref{s:graph}). The embedding is based on graph convolutional neural networks~\cite{graphcnn}, but it is customized for  scheduling. The key ingredient is a parametrized non-linear aggregation operation that is performed iteratively on the DAG to derive an embedding for each node based on the embeddings of its children. By reusing aggregation operations within and across DAGs, \name's embedding scales to an arbitrary number of job DAGs with arbitrary sizes and shapes.

On each state observation, \name must convert the state information (job DAGs and executor status) into features to pass to its policy network. One option is to create a flat feature vector containing all the state information. However, this approach cannot scale to arbitrary number of DAGs of arbitrary sizes and shapes. Further, even with a hard limit on the number of jobs and stages, processing a high-dimensional feature vector would require a large policy network that would be difficult to train.

\name achieves scalability using a {\em graph neural network}, which encodes or ``embeds'' the state information (\eg attributes of job stages, DAG dependency structure, etc.) in a set of {\em embedding} vectors.
Our method is based on graph convolutional neural networks~\cite{graphcnn, graphcombopt, gcnn_google} but customized for scheduling.
Table~\ref{t:embedding-notation} defines our notation.

\begin{table}[t]
\footnotesize
\centering
\begin{tabular}{|r|l||r|l|}
\hline
  \bf entity & \bf symbol & \bf entity & \bf symbol \\
\hline
%  job & $i$ & per-node feature vector & $\mathbf{x}^i_v$ \\
%  stage (DAG node) & $v$ & per-node embedding & $\mathbf{e}^i_v$ \\
%  job $i$'s parallelism & $l_i$ & per-job embedding & $\mathbf{y}^i$ \\
%  node score & $q_v^i$ & global embedding & $\mathbf{z}$ \\
%  parallelism score & $w_l^i$ & non-linear functions & $f, g, q, w$ \\
%
  job & $i$ & per-node feature vector & $\mathbf{x}^i_v$ \\
  stage (DAG node) & $v$ & per-node embedding & $\mathbf{e}^i_v$ \\
  node $v$'s children & $\xi(v)$ & per-job embedding & $\mathbf{y}^i$ \\
  job $i$'s DAG & $G_i$ & global embedding & $\mathbf{z}$ \\
  job $i$'s parallelism & $l_i$ & node score & $q_v^i$ \\
  non-linear functions & $f, g, q, w$ & parallelism score & $w_l^i$ \\
%
%  job & $i$ & per-node feature vector & $\mathbf{x}^i_v$ \\
%  stage (DAG node) & $v$ & per-node embedding & $\mathbf{e}^i_v$ \\
%  job $i$'s DAG & $G_i$ & per-job embedding & $\mathbf{y}^i$ \\
%  non-linear functions & $f, g, q, w$ & global embedding & $\mathbf{z}$ \\
%
%  job & $i$ & per-node embedding & $\mathbf{e}^i_v$ \\
%  stage (DAG node) & $v$ & per-job embedding & $\mathbf{y}^i$ \\
%  job $i$'s DAG & $G_i$ & global embedding & $\mathbf{z}$ \\
%  per-node feature vector & $\mathbf{x}^i_v$ &  non-linear functions & $f, g, q, w$ \\
  %
%  job & $i$ &                           per-node raw feature vector & $\mathbf{x}^i_v$ \\
%  total number of executors & $N$ &     per-node vector & $\mathbf{e}^i_v$ \\
%  total number of nodes & $D$ &         per-job vector & $\mathbf{y}^i$ \\
%  stage (DAG node) & $v$ &              global vector & $\mathbf{z}$ \\
%  graph structure & $G_i$ &             non-linear functions & $f, g, q$ \\
\hline
\end{tabular}
\vspace{0.1cm}
\caption{Notation used throughout \S\ref{s:design}.}
\vspace{-0.85cm}
\label{t:embedding-notation}
\end{table}

The graph embedding takes as input the job DAGs whose nodes carry a set of stage attributes (\eg the number of remaining tasks, expected task duration, etc.), and it outputs
three different types of embeddings:
\begin{CompactEnumerate}
  \item per-node embeddings, which capture
    information about the node and its children (containing, \eg aggregated work
    along the critical path starting from the node);
  \item per-job embeddings, which aggregate information across an entire
    job DAG (containing, \eg the total work in the job); and
  \item a global embedding, which combines information from all per-job embeddings into a cluster-level summary (containing, \eg the number of jobs and the cluster
    load).
\end{CompactEnumerate}
Importantly, what information to store in these embeddings is not hard-coded\,---\,\name
automatically learns what is statistically important and how to compute it from
the input DAGs through end-to-end training.
In other words, the embeddings can be thought of as feature vectors that the
graph neural network learns to compute without manual feature engineering.
\name's graph neural network is scalable because it reuses a common set of  operations as building blocks to compute the above embeddings. These building blocks are themselves implemented as small neural networks that operate on relatively low-dimensional input vectors.

\begin{figure}[t]
\centering
\begin{subfigure}[t]{0.22\textwidth}
  \centering
  \includegraphics[height=1.4in]{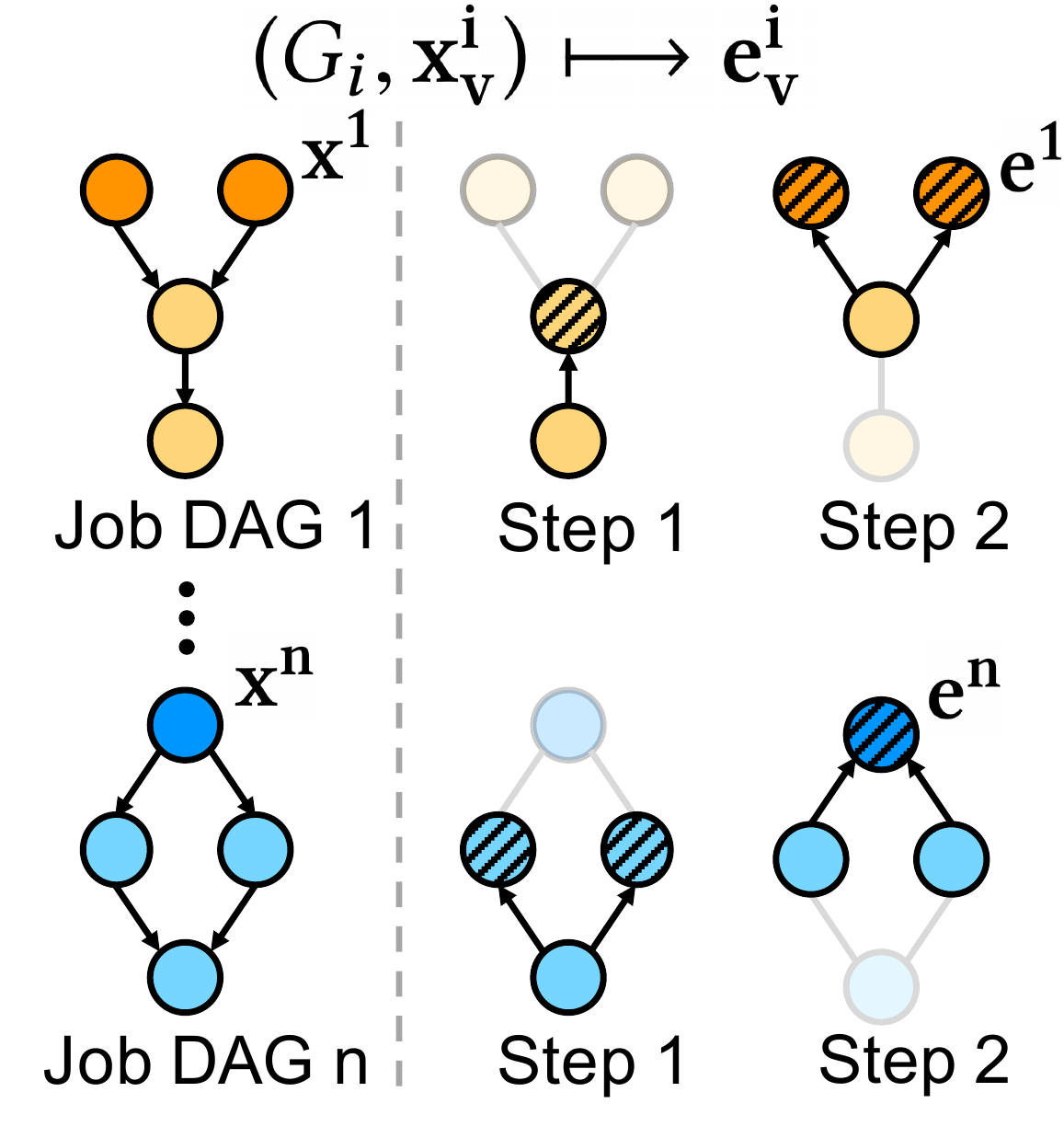}
  \vspace{-0.1cm}
  \caption{Per-node embedding.}
  \label{fig:msg}
\end{subfigure}
\begin{subfigure}[t]{0.245\textwidth}
  \centering
  \includegraphics[height=1.4in]{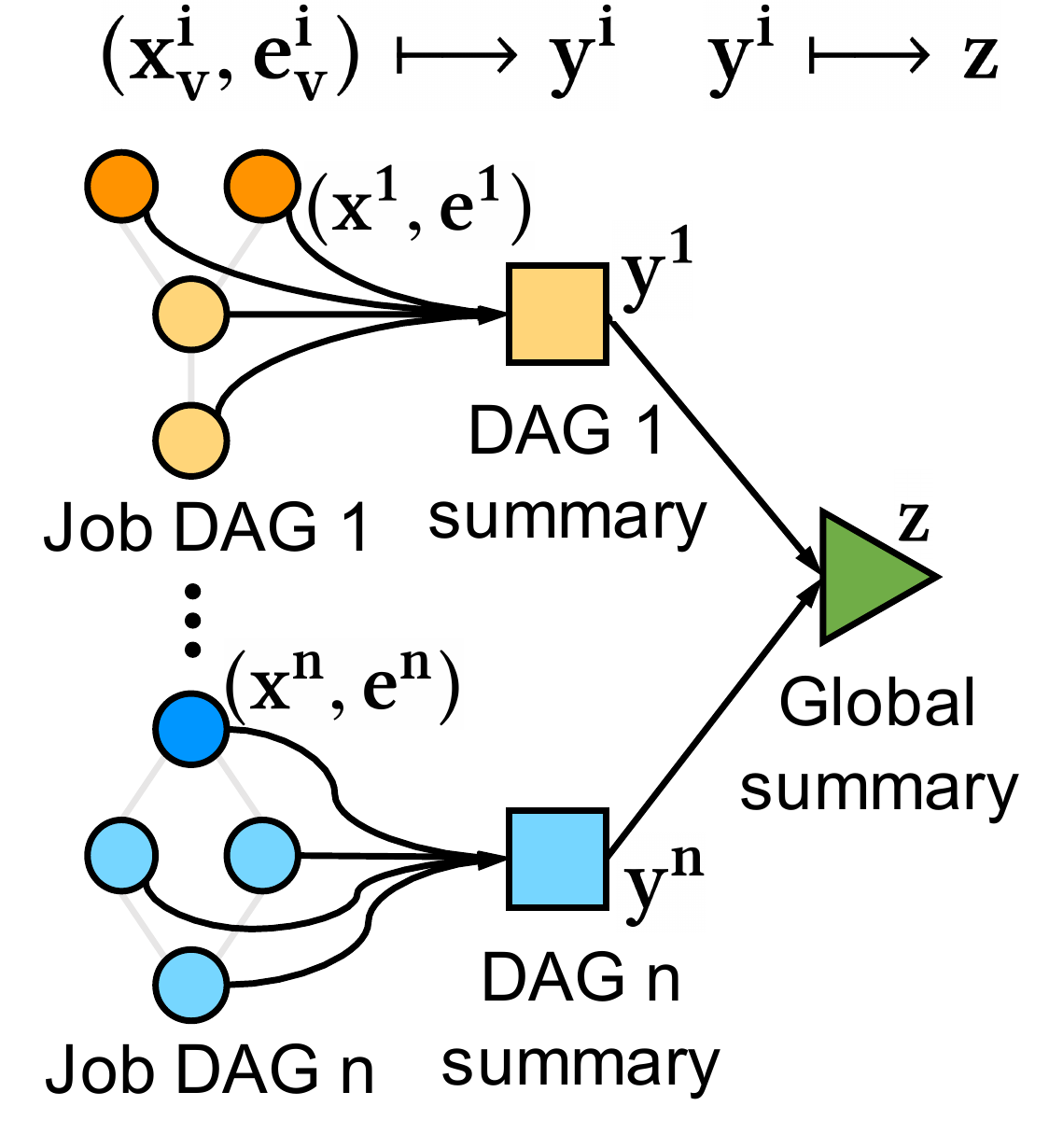}
  \vspace{-0.1cm}
  \caption{Per-job and global embeddings.}
  \label{fig:summ}
\end{subfigure}
\vspace{-0.2cm}
\caption{A \emph{graph neural network} transforms the raw information on each DAG node into a vector representation.
  This example shows two steps of local message passing and two levels of summarizations.}
\vspace{-0.3cm}
\label{fig:graph}
\end{figure}

\para{Per-node embeddings.}
Given the vectors $\mathbf{x}^i_v$ of stage attributes corresponding to the nodes in DAG $G_i$, \name
builds a per-node embedding $(G_i, \mathbf{x}^i_v)
\longmapsto{\mathbf{e}^i_v}$.
The result $\mathbf{e}^i_v$ is a vector (e.g., in $\mathbb{R}^{16}$) that captures information from all nodes reachable from
$v$ (\ie $v$'s child nodes, their children, etc.).
To compute these vectors, \name propagates information from children to parent nodes in
a sequence of \emph{message passing} steps, starting from the leaves of the DAG
(Figure~\ref{fig:msg}).
In each message passing step, a node $v$ whose children have aggregated messages
from all of their children (shaded nodes in Figure~\ref{fig:msg}'s examples)
computes its own embedding as:
\begin{equation}
	\mathbf{e}_v^{i} = g\left[\sum_{u\in \xi(v)}f(\mathbf{e}_u^{i})\right] + \mathbf{x}_v^{i},
	\label{eq:msg}
\end{equation}
%\begin{equation}
%	\mathbf{x}_i \leftarrow g\left[\sum_{j\in \xi(i)}f(\mathbf{x}_j)\right] + \mathbf{x}_i,
%	\label{eq:msg}
%\end{equation}
where $f(\cdot)$ and $g(\cdot)$ are non-linear transformations over vector
inputs, implemented as (small) neural networks, and $\xi(v)$ denotes the set of
$v$'s children.
The first term is a general, non-linear aggregation operation that summarizes
the embeddings of $v$'s children; adding this summary term to $v$'s feature
vector ($\mathbf{x}_v$) yields the embedding for $v$.
\name reuses the same non-linear transformations $f(\cdot)$ and $g(\cdot)$ at all nodes, and in all message passing steps.
%

%The key to a {\em scalable} embedding, i.e., one that works for DAGs of any size and shape, is to use

%
%Equation~\eqref{eq:msg} meets this requirement by using a sum to aggregate information across the children nodes; hence it can be applied to any number of children in $\xi(v)$.

% %
% Equation~\eqref{eq:msg} meets this requirement as it is a \emph{set-function},
% \ie a function that applies over any number of input (\viz children in
% $\xi(v)$).
% %

%Taking a sum across neighbors is a common technique in graph
%embeddings~\cite{graphcnn, graphspectral, graphcombopt}.

Most existing graph neural network architectures~\cite{graphcnn, graphspectral, graphcombopt} use aggregation operations of the form
$\mathbf{e_v} = \sum_{u\in \xi(v)} f(\mathbf{e_u})$ to compute node embeddings. However, we found that adding a second non-linear transformation $g(\cdot)$ in Eq.~\eqref{eq:msg} is critical for learning strong scheduling policies.
The reason is that without $g(\cdot)$, the graph neural network cannot compute certain useful features for scheduling. For example, it cannot compute the critical path~\cite{criticalpath} of a DAG, which requires a max operation across the children of a node during message passing.\footnote{The critical path from node $v$ can be computed as: $\text{cp}(v) = \max_{u \in \xi(v)}\text{cp}(u) + \text{work}(v)$, where work($\cdot$) is the total work on node $v$.} Combining two non-linear transforms $f(\cdot)$ and $g(\cdot)$ enables \name to express a wide variety of aggregation functions. For example, if $f$ and $g$ are identity transformations, the aggregation
sums the child node embeddings; if $f\sim \log(\cdot / n)$,
$g\sim \exp(n \times \cdot)$, and $n\to\infty$, the aggregation computes the
max of the child node embeddings. \update{We show an empirical study of this embedding in Appendix~\ref{s:su}.}

\para{Per-job and global embeddings.}
The graph neural network also computes a summary of all node embeddings for each DAG $G_i$,
$\{(\mathbf{x}^i_v, \mathbf{e}^i_v), v\in G_i\} \longmapsto{\mathbf{y}^i}$; and
a global summary across all DAGs,
% (e.g., for computing the action probability on a node \emph{conditioned} on the status of other DAGs),
$\{\mathbf{y}^1,\mathbf{y}^2,\ldots\} \longmapsto{\mathbf{z}}$.
%
%For example, local message passing cannot capture ``multipath'' structure and
%therefore may aggregate redundant information (e.g., DAG $n$ in
%Figure~\ref{fig:msg}), and it cannot flow information across DAGs.
%\ma{No one will understand what you mean by multipath structure; remove this.}
%
To compute these embeddings, \name adds a summary node to each DAG, which has all the nodes in the DAG as
children (the squares in Figure~\ref{fig:summ}). These DAG-level summary nodes are in turn children of a single global summary node (the
triangle in Figure~\ref{fig:summ}).
The embeddings for these summary nodes are also computed using Eq.~\eqref{eq:msg}. Each level of summarization has its own non-linear transformations $f$ and $g$; in other words, the graph neural network uses six non-linear transformations in total, two for each level of summarization.
%
%This summarization is complementary to local message passing\,---\,each fills
%in omitted information from the other.
%

%%%%%%%%%%%%%%%%%%%%%%%%%%%%%%%%%%%%%%%%%%%%%%%%%%%%%%%%%%%%%%%%%%%%%%%%%%%%%

\subsection{Encoding scheduling decisions as actions}
\label{s:action}
The key challenge for encoding scheduling decisions lies in the learning and
computational complexities of dealing with large action spaces.
As a naive approach, consider a solution, that given the embeddings from \S\ref{s:graph}, returns the assignment for all executors to job stages in one shot. This approach has to choose actions from an exponentially large set of combinations. On the other extreme, consider a solution that invokes the scheduling agent to pick one stage every time an executor becomes available. This approach has a much smaller action space (O(\# stages)), but it requires long sequences of actions to schedule a given set of jobs. In RL, both large action spaces and long action sequences increase sample complexity and slow down training~\cite{rlbook, aicompute}.

\name balances the size of the action space and the number of actions required by decomposing scheduling decisions into a series of \update{two-dimensional} actions, which output \one{} a stage designated to be scheduled next, and \two{} an upper limit on the number of executors to use for that stage's job.

% To efficiently express the scheduling decision (i.e., matching
% between runnable stages and executors), we design a two-dimensional RL action.
% Specifically, \name passes the processed vectors
% from~\S\ref{s:graph} as input to a \emph{policy neural network},
% %to make the actual scheduling decision,
% which outputs of a composite action $\langle v, l_i \rangle$ of a
% stage $v$ and a maximum level of parallelism $l_i$ for $v$'s job $i$. In the below,
% we explain the scheduling window, the stage and parallelism selection, and a
% technique to further speed up training by leveraging Spark domain insight.

%
\para{Scheduling events.}
\name invokes the scheduling agent when the set of runnable
stages\,---\,\ie stages whose parents have completed and which have at least
one waiting task\,---\,in any job DAG changes.
%
%A node of any job is on the frontier if its parent nodes have all completed.
% or do not exist (i.e., root nodes) \ma{can we remove `or do not exist'; unclear what this means.}.
%Hence, frontier nodes are immediately runnable.
%
Such scheduling events happen when \one{} a stage runs out of tasks (\ie needs
no more executors), \two{} a stage completes, unlocking the tasks of one or
more of its child stages, or \three{} a new job arrives to the system.

At each scheduling event, the agent schedules a group of free executors in one or more actions. Specifically, it passes the embedding vectors from~\S\ref{s:graph} as input to the policy network, which outputs a two-dimensional action $\langle v, l_i \rangle$, consisting of a stage $v$ and the parallelism limit $l_i$ for $v$'s job $i$. If job $i$ currently has fewer than $l_i$ executors, \name assigns executors to $v$ up to the limit.
If there are still free executors \update{after the scheduling action,} \name invokes the agent again to select another stage and parallelism limit. This process repeats until all the executors have been assigned, or there are no more runnable stages. \name ensures that this process completes in a finite number of steps by enforcing that the parallelism limit $l_i$ is greater than the number of executors currently allocated to job $i$, so that at least one new executor is scheduled with each action.
%\ma{TODO: maybe explain why this is guaranteed to hold.}

%At any point in time (as long as there are jobs), \name maintains a unique `preferred node', which is a runnable stage with tasks remaining to be scheduled.
%Whenever an executor becomes free a task from the preferred node is assigned to it, which is subsequently run on the executor until completion.
%Scheduling events in \name happen when the preferred node has no more tasks to be scheduled.
%At this point, the scheduling agent is consulted to select a different node, which then becomes the new preferred node.

\begin{figure}[t]
\centering
\includegraphics[height=5cm]{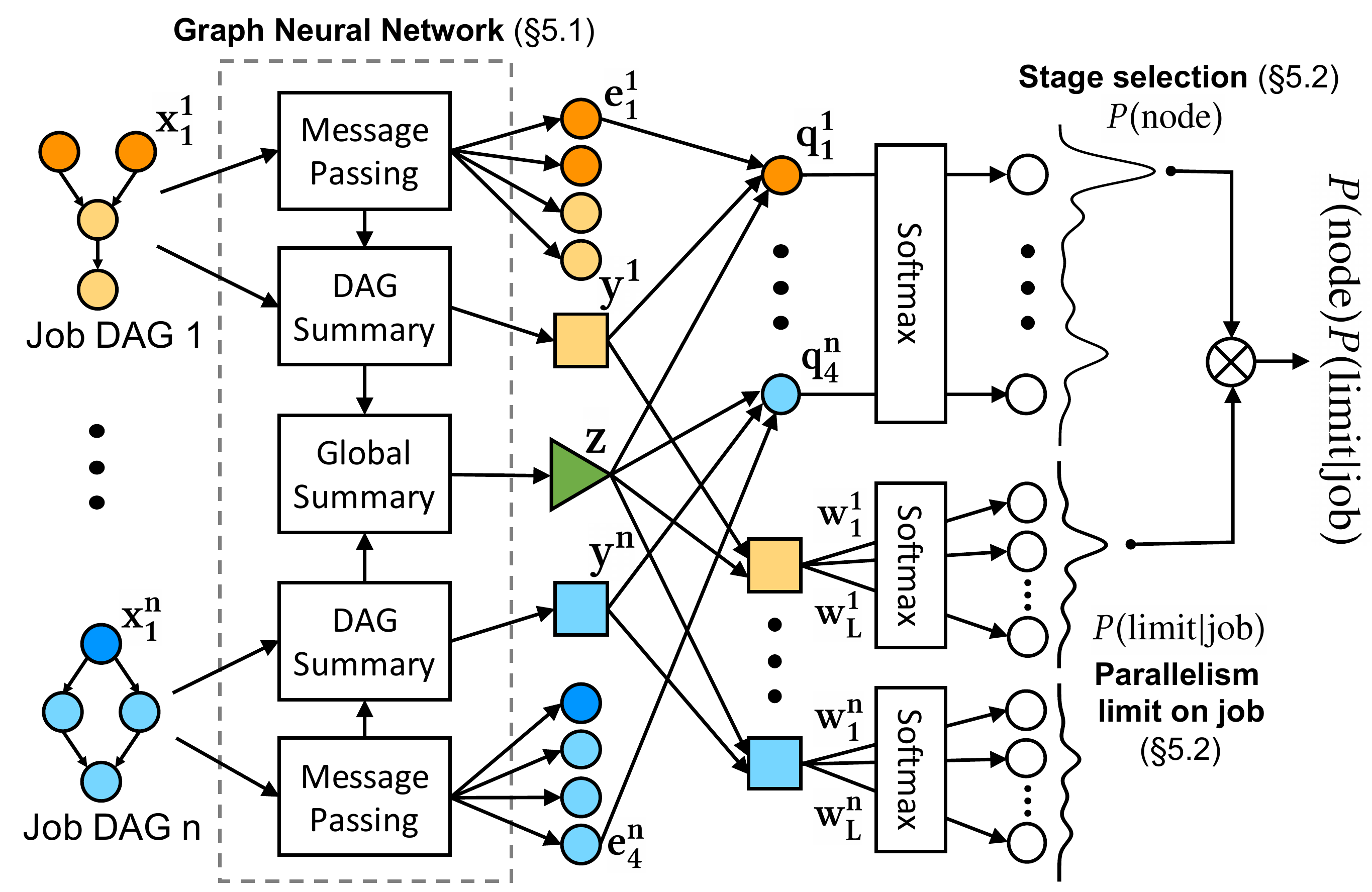}
\vspace{-0.3cm}
\caption{For each node $v$ in job $i$, the \emph{policy network}
  uses per-node embedding $\mathbf{e}^i_v$, per-job embedding $\mathbf{y}^i$ and global embedding $\mathbf{z}$ to compute \one{} the score
  $\mathbf{q}^i_v$ for sampling a node to schedule and \two{} the score
  $\mathbf{w}^i_l$ for sampling a parallelism limit for the node's job.}
\label{f:end-to-end-policy-net}
\vspace{-0.4cm}
\end{figure}

%
\para{Stage selection.}
Figure~\ref{f:end-to-end-policy-net} visualizes \name's policy network. For a scheduling event at time $t$, during which the state is $s_t$, the policy network selects a stage to schedule as follows. For a node $v$ in job $i$, it computes a score $q_v^{i} \triangleq q(\mathbf{e}_v^{i}, \mathbf{y}^{i}, \mathbf{z})$, where $q(\cdot)$ is a {\em score function} that maps the embedding vectors~(output from the graph neural network in \S\ref{s:graph}) to a scalar value.
Similar to the embedding step, the score function is also a non-linear transformation implemented as a neural network.
%and \name reuses the same score function $q(\cdot)$ to all nodes for scaleability.
%
The score $q_v^i$ represents the priority of scheduling node $v$.
\name then uses a softmax operation~\cite{bishop} to compute the probability of selecting node $v$ based on the priority scores:
\begin{equation}
P(\text{node} = v) = \frac{\exp(q_v^i)}{\sum_{u\in\mathcal{A}_t}\exp(q_u^{j(u)})}, \label{eq:softmax}
\end{equation}
where $j(u)$ is the job of node $u$, and $\mathcal{A}_t$ is the set of nodes that can be scheduled at time $t$.
\update{Notice that $\mathcal{A}_t$ is known to the RL agent at each step,
since the input DAGs tell exactly which stages are runnable. Here, $\mathcal{A}_t$ restricts which outputs are considered by the softmax operation. The whole operation is end-to-end differentiable.}

% The priority scores help the agent select an action based on its
% learned policy $\pi: \pi(s_t, a_t) \to [0, 1]$.
% %
% $\pi(s_t, a_t)$ is the probability that action $a_t$ is taken in state
% $s_t$, and \name computes it using a standard ``softmax''
% operation~\cite{bishop}.
% %
% Using the $q(\cdot)$ priority scores, the probability of selecting node
% $v$ in job $j(v)$ thus becomes:

\para{Parallelism limit selection.}
%The stage selection model above can decide the order in which stages are
%scheduled, but it cannot control the level of parallelism for a job.
%
%Choosing the wrong level of parallelism can make jobs inefficient and
%waste resources that others need (\S\ref{s:motiv-parallelism}).
%
%Thus, we augment \name's action space to explicitly include a parallelism limit.
%
%For each job $i$, the limit $l_i$ specifies a bound in $\{1,2,\ldots, N\}$,
%where $N$ is the total number of executors.
%
\update{
Many existing schedulers set a static degree of parallelism for each job: \eg Spark by default takes the number of executors as a command-line argument on job submission.
\name adapts a job's parallelism each time it makes a scheduling decision for that job, and varies the parallelism as different stages in the job become runnable or finish execution.
}

For each job $i$, \name's policy network also computes a score $w^i_l \triangleq w(\mathbf{y}^i, \mathbf{z}, l)$
for assigning parallelism limit $l$ to job $i$, using another score function $w(\cdot)$. Similar to stage selection, \name applies a softmax operation on these scores to compute the probability of selecting a parallelism limit (Figure~\ref{f:end-to-end-policy-net}).
%

% There is one subtlety worth pointing out about how we have defined the score function. A standard method to pick a parallelism limit $l \in \[1,L\]$ is to map the inputs $(\bm{y}^i, \bm{z})$ to $L$ numbers and take the softmax. Typically, this would be achieved by using $L$ separate functions $w_

Importantly, \name uses the same score function $w(\cdot)$ for all jobs and all parallelism limit values. This is possible because the score function takes the parallelism $l$ as one of its inputs. Without using $l$ as an input, we cannot distinguish between different parallelism limits, and would have to use separate functions for each limit. Since the number of possible limits can be as large as the number of executors, reusing the same score function significantly reduces the number of parameters in the policy network and speeds up training~(Figure~\ref{f:eval-learning-curve}).

% %
% This is because otherwise the policy network cannot distinguish between different limit values
% from its inputs (which only contains the embedding vectors) and has to use \emph{separate} score functions for different limits.
% %
% With the limit value explicitly present in the input, \name is able to reuse the same score function $w(\cdot)$ for all the jobs and all the parallelism limit.
% This significantly reduces the number of parameters in the policy network and speeds up the training for \name~(Figure~\ref{f:eval-learning-curve}).
% %
% %This also enables \name to generalize to unseen deployment cluster with larger size.
% %
\update{
\name's action specifies \emph{job-level} parallelism (\eg ten total executors for the entire job), as opposed fine-grained stage-level parallelism.
This design choice} trades off granularity of control for a model that is easier to train.
In particular, restricting \name to job-level parallelism control reduces the space of scheduling policies that it must explore and optimize over during training.

\update{
However, \name still maintains the expressivity to (indirectly) control stage-level parallelism.
On each scheduling event, \name picks a stage $v$, and new parallelism limit $l_i$ for $v$'s job $i$.
The system then schedules executors to $v$ until $i$’s parallelism reaches the limit $l_i$.
Through repeated actions with different parallelism limits, \name can add desired numbers of executors to specific stages.
For example, suppose job $i$ currently has ten executors, four of which are working on stage $v$.
To add two more executors to $v$, \name, on a scheduling event, picks stage $v$ with parallelism limit of 12.}
Our experiments show that \name achieves the same performance with job-level parallelism as with fine-grained, stage-level parallelism choice, at substantially accelerated training (Figure~\ref{f:eval-learning-curve}).

\subsection{Training}
\label{s:training}

The primary challenge for training \name is how to train with continuous stochastic job arrivals. To explain the challenge, we first describe the RL algorithms used for training.

RL training proceeds in {\em episodes}. Each episode consists of multiple scheduling events, and each scheduling event includes one or more actions. Let $T$ be the total number of actions in an episode ($T$ can vary across different episodes), and $t_k$ be the wall clock time of the $k^{\textrm{th}}$ action. To guide the RL algorithm, \name gives the agent a \emph{reward} $r_k$ after each action based on its high-level scheduling objective. For example, if the objective is to minimize the average JCT, \name penalizes the agent $r_k = -(t_k - t_{k-1}) J_k$ after the $k^{th}$ action, where $J_k$ is the number of jobs in the system during the interval $[t_{k-1}, t_k)$. The goal of the RL algorithm is to minimize the expected time-average of the penalties: $\mathbb{E}\left[1/t_T \sum_{k=1}^{T} (t_k - t_{k-1})J_k \right]$. This objective minimizes the average number of jobs in the system, and hence, by Little's law~\cite[\S5]{little}, it effectively minimizing the average JCT.

%Next, we describe the policy gradient methods for training, and how we modify it to handle continuous job arrivals.
%\para{Policy gradient training.}

\name uses a policy gradient algorithm for training. The main idea in policy gradient methods is to learn by
performing gradient descent on the neural network parameters using the rewards observed during training.
Notice that all of \name's operations,
from the graph neural network~(\S\ref{s:graph}) to the policy network~(\S\ref{s:action}), are differentiable. For conciseness, we denote all of the parameters in these operations jointly as $\theta$, and the scheduling policy as $\pi_\theta(s_t, a_t)$\,---\,defined as the probability of taking action $a_t$ in state $s_t$.

Consider an episode of length $T$,
where the agent collects (\emph{state, action, reward}) observations, \ie
$(s_k, a_k, r_k)$, at each step $k$.
The agent updates the parameters $\theta$ of its policy $\pi_\theta(s_t, a_t)$ using the REINFORCE policy gradient algorithm~\cite{reinforce}:
%Let $\theta$ denote the vector of neural network parameters used to simulate the episode, and let $P(a_t|s_t;\theta)$ denote the probability of making action $a_t$ at state $s_t$.
%Then at the end of the episode $\theta$ is updated as
\begin{align}
\vspace{-0.2cm}
\theta \leftarrow \theta + \alpha\sum_{k=1}^{T} \nabla_\theta \log \pi_\theta(s_k, a_k) \left( \sum_{k' = k}^{T} r_{k'} - b_k \right). \label{eq:policygradient}
\vspace{-0.2cm}
\end{align}
Here, $\alpha$ is the learning rate and $b_k$ is a {\em baseline} used to reduce the variance of the policy gradient~\cite{baseline}.
An example of a baseline is a ``time-based'' baseline~\cite{deeprm, greensmith}, which sets $b_k$ to the cumulative reward from step $k$ onwards, averaged over all training episodes.
Intuitively, $\left( \sum_{k'}r_{k'} - b_k \right)$ estimates how much better
(or worse) the total reward is (from step $k$ onwards) in a particular episode compared to the average case;
and $\nabla_\theta \log \pi_\theta(s_k, a_k)$ provides a direction in the parameter space to increase the probability of choosing action $a_k$ at state $s_k$.
As a result, the net effect of this equation is to increase the probability of
choosing an action that leads to a better-than-average reward.\footnote{The update rule in Eq.~\eqref{eq:policygradient} aims to maximize the sum of rewards during an episode. To maximize the time-average of the rewards, \name uses a slightly modified form of this equation. See Appendix~\ref{s:background} for details.}
%
%The training procedure follows DeepRM's~\cite{deeprm} and uses the same
%time-based baseline technique.
%We implement \name's training framework using TensorFlow~\cite{tensorflow}
%and we defer the implementation details and hyperparameter settings to Appendix~\ref{s:pg}.
%
%%%%%%%%%%%%%%%%%%%%%%%%%%%%%%%%%%%%%%%%%%%%%%%%%%%%%%%%%%%%%%%%%%%%%%%%%%%%%

\para{Challenge \#1: Training with continuous job arrivals.}
To learn a robust scheduling policy, the agent has to experience ``streaming'' scenarios, where jobs arrive continuously over time, during training.
Training with ``batch'' scenarios, where all jobs arrive at the beginning of an episode, leads to poor policies in streaming settings (e.g., see Figure~\ref{f:eval-ubench-component-breakdown}).
However, training with a continuous stream of job arrivals is non-trivial.
In particular, the agent's initial policy is very poor (\eg as the initial parameters are random).
Therefore, the agent cannot schedule jobs as quickly as they arrive in early training episodes, and a large queue of jobs builds up in almost every episode.
\update{Letting the agent explore beyond a few steps in these early episodes wastes training time, because the overloaded cluster scenarios it encounters will not occur with a reasonable policy.}

To avoid this waste, we terminate initial episodes early so that the agent can reset and quickly try again from an idle state. We gradually increase the episode length throughout the training process. Thus, initially, the agent learns to schedule short sequences of jobs. As its scheduling policy improves, we increase the episode length, making the problem more challenging. The concept of gradually increasing job sequence length\,---\,and therefore, problem complexity\,---\,during training realizes curriculum learning~\cite{curlearn} for cluster scheduling.

One subtlety about this method is that the termination cannot be deterministic. Otherwise, the agent can learn to predict when an episode terminates, and defer scheduling certain large jobs until the termination time. This turns out to be the optimal strategy over a fixed time horizon: since the agent is not penalized for the remaining jobs at termination, it is better to strictly schedule short jobs even if it means starving some large jobs. We found that this behavior leads to indefinite starvation of some jobs at runtime (where jobs arrive indefinitely). To prevent this behavior, we use a \emph{memoryless} termination process. Specifically, we terminate each training episode after a time $\tau$, drawn randomly from an exponential distribution. As explained above, the mean episode length increases during training up to a large value (e.g., a few hundreds of job arrivals on average).
%\footnote{Training \name in a ``batch'' setting without jobs arriving over time does not generalize to continuous job arrivals; see Figure~\ref{f:eval-batch-in-streaming} in Appendix~\ref{s:batch_in_streaming}.}
%
\begin{figure}[t]
\centering
\includegraphics[height=3.4cm]{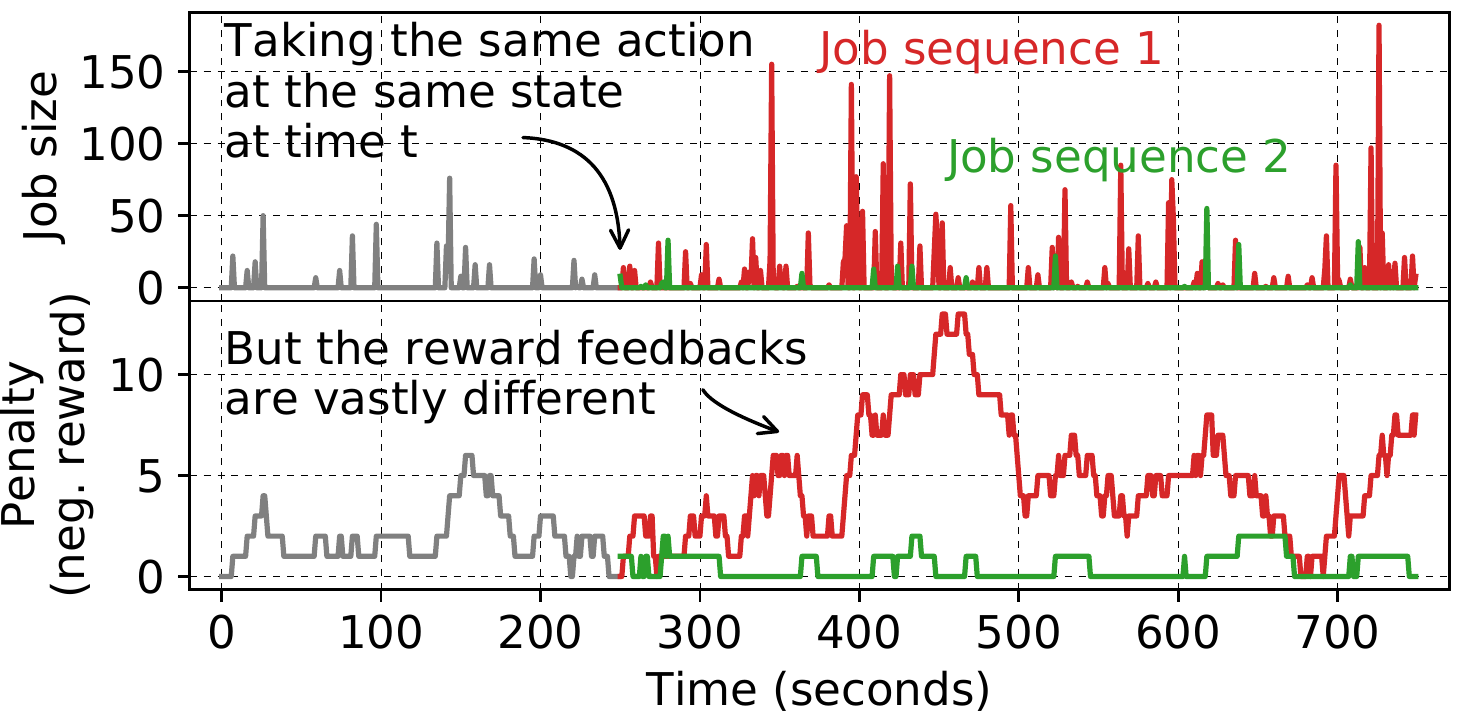}
\vspace{-0.3cm}
\caption{Illustrative example of how different job arrival sequences can lead to vastly different rewards. After time $t$, we sample two job arrival sequences, from a Poisson arrival process (10~seconds mean inter-arrival time) with randomly-sampled TPC-H queries.}
\vspace{-0.4cm}
\label{f:design-var}
\end{figure}

\para{Challenge \#2: Variance caused by stochastic job arrivals.}
%
%The third challenge occurs uniquely due to the stochastic nature of job arrival during streaming.
Next, for a policy to generalize well in a streaming setting, the training episodes must include many different job arrival patterns.
This creates a new challenge: different job arrival patterns have a large impact on performance, resulting in vastly different rewards.
Consider, for example, a scheduling action at the time $t$ shown in Figure~\ref{f:design-var}. If the arrival sequence following this action consists of a burst of large jobs (e.g., job sequence 1), the job queue will grow large, and the agent will incur large penalties. On the other hand, a light stream of jobs (e.g., job sequence 2) will lead to short queues and small penalties. The problem is that this difference in reward has nothing to do with the action at time $t$\,---\,it is caused by the randomness in the job arrival process. Since the RL algorithm uses the reward to assess the goodness of the action, such variance adds noise and impedes effective training.

%Importantly, this difference in rewards has little to do with the scheduling action; it is a consequence of the random job arrival process. Since this variance is due to randomness in the job arrival process,  \emph{not} the quality of scheduling decisions, it adds significant noise to
%the reward feedback and distorts the policy gradient
%estimation in Equation~\eqref{eq:policygradient}.
%

To resolve this problem, we build upon a
recently-proposed variance reduction technique for ``input-driven''
environments \cite{variance-reduction}, where an exogenous, stochastic input process (e.g., \name's job arrival process) affects the dynamics of the system.
The main idea is to fix the {\em same} job arrival sequence in multiple training episodes,
and to compute separate baselines specifically for each  arrival sequence. In particular,
instead of computing the baseline $b_k$ in Eq.~\eqref{eq:policygradient} by averaging over
episodes with different arrival sequences, we average only over episodes with the same
arrival sequence. During training, we repeat this procedure for a large number of
randomly-sampled job arrival sequences \update{(\S\ref{s:eval-real-cluster} and \S\ref{s:multi-dim} describe how we
generate the specific datasets for training)}. This method removes the variance caused by the job
arrival process entirely, enabling the policy gradient algorithm to assess the goodness of
different actions much more accurately~(see Figure~\ref{f:eval-ubench-component-breakdown}).
For the implementation details of our training and the hyperparameter settings used, see Appendix~\ref{s:pg}.

\section{Implementation}
\label{s:impl}
%\end{figure*}
We have implemented \name as a pluggable scheduling service that parallel data
processing platforms
%(\eg Spark, Dryad, or YARN)
can communicate with over an
RPC interface.
%
%The \name service consists of 1389 lines of Python, of which 647 lines comprise
%the TensorFlow model served.
%
In \S\ref{s:impl-spark}, we describe the integration of \name
with Spark.
%
% (mention this is a general purposed platform?)
%
Next, we describe our Python-based training infrastructure which
%consists of 1654 lines of code and
includes an accurate Spark cluster simulator~(\S\ref{s:simulator}).

%%%%%%%%%%%%%%%%%%%%%%%%%%%%%%%%%%%%%%%%%%%%%%%%%%%%%%%%%%%%%%%%%%%%%%%%%%%%%%%
\subsection{Spark integration}
\label{s:impl-spark}

\begin{figure}
 \centering
 \includegraphics[height=3.6cm]{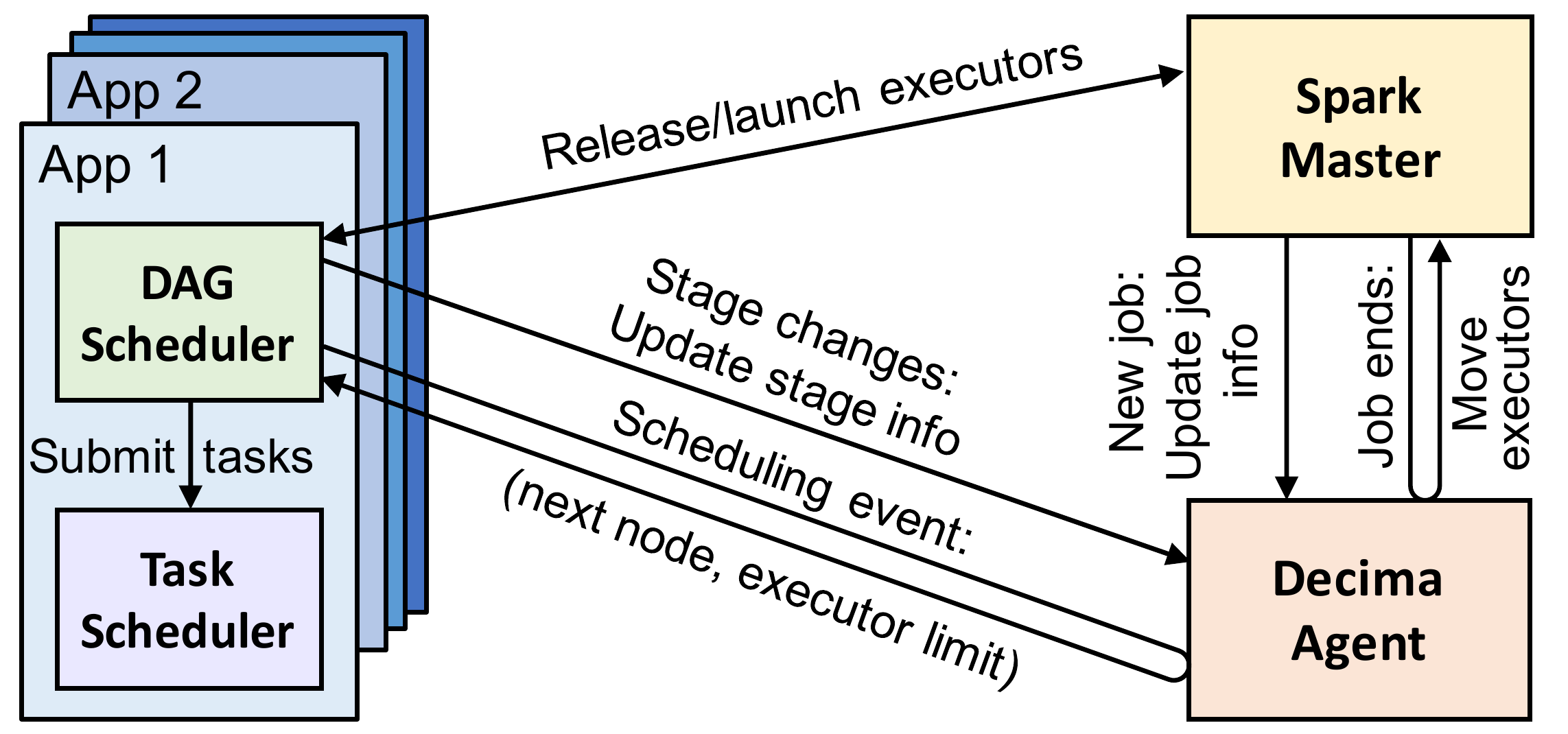}
 \vspace{-0.3cm}
 \caption{Spark standalone cluster architecture, with \name additions
          highlighted.}
 \label{f:impl-spark-architecture}
\vspace{-0.4cm}
\end{figure}

A Spark cluster\footnote{We discuss Spark's ``standalone'' mode of operation
  here (\url{http://spark.apache.org/docs/latest/spark-standalone.html});
  YARN-based deployments can, in principle, use \name, but require modifying
  both Spark and YARN.}
runs multiple parallel \emph{applications}, which contain one or more jobs that together form a DAG of processing stages.
The Spark master manages application execution and monitors the health of many
\emph{workers}, which each split their resources between multiple executors.
Executors are created for, and remain associated with, a specific application,
which handles its own scheduling of work to executors.
%
%Internally to an application, two levels of scheduling exist: the application's
%\emph{DAG scheduler} chooses stages to work on and submits their tasks to a
%lower-level \emph{task scheduler} that maps running stages' fine-grained
%parallel \emph{tasks} to executors.
%
Once an application completes, its executors terminate.
Figure~\ref{f:impl-spark-architecture} illustrates this architecture.
\smallskip
\noindent
To integrate \name in Spark, we made two major changes:
\begin{CompactEnumerate}
 \item Each application's \textbf{DAG scheduler} contacts \name on startup and
       whenever a scheduling event occurs.
       \name responds with the next stage to work on and the parallelism limit (\S\ref{s:action}).
 \item The Spark \textbf{master} contacts \name when a new job arrives to
       determine how many executors to launch for it, and
       \update{aids \name by taking
       executors away from a job once they complete a stage.}
\end{CompactEnumerate}
%
%\todo{scheduling across applications}
%
%Our changes amount to a 901-line patch to Spark v2.2.
%

\para{State observations.}
In \name, the feature vector
$\mathbf{x}^i_v$~(\S\ref{s:graph}) of a node $v$ in job DAG $i$ consists of: \one{} the number of
tasks remaining in the stage, \two{} the average task duration, \three{} the
number of executors currently working on the node, \four{} the number of available
executors, and \five{} whether available executors are local to the job.
\update{We picked these features by attempting to include information necessary to capture
the state of the cluster (\eg the number of executors currently assigned to each stage), as
well as the statistics that may help in scheduling decisions (\eg a stage's average task duration).
These statistics depend on the information available (\eg profiles from past executions of the
same job, or runtime metrics) and on the system used (here, Spark).
\name can easily incorporate additional signals.}

% %
% These features are \emph{raw}, \ie chosen without a human design intention.
% %
% However, they intuitively capture some useful information.
% %
% For example, the task count and average task duration can help estimate the
% remaining work in a stage.
% %
% To learn a good scheduling policy, \name must learn to exploit this raw
% information.

\para{Neural network architecture.}
The graph neural network's six transformation functions $f(\cdot)$ and $g(\cdot)$ (\S\ref{s:graph}) (two each for node-level, job-level, and global embeddings)
and the policy network's two score functions $q(\cdot)$ and $w(\cdot)$ (\S\ref{s:action}) are implemented using
two-hidden-layer neural networks, with 32 and 16 hidden units
on each layer. Since these neural networks are reused for all jobs
and all parallelism limits, \name's model is lightweight\,---\,it consists of  12,736 parameters (50KB) in total. Mapping the cluster state to a scheduling decision takes
less than 15ms (Figure~\ref{f:eval-inference-time}).
%on Intel Xeon E5-2640 CPU

%These features form a compact, yet reasonably accurate, representation of the node state at any time.
%Number of tasks and average task duration are useful to estimate the amount of work remaining in the stage;
%whereas average first wave, rest wave durations and the number of executors assigned to the job are critical for deciding parallelism limits.
%%%%%%%%%%%%%%%%%%%%%%%%%%%%%%%%%%%%%%%%%%%%%%%%%%%%%%%%%%%%%%%%%%%%%%%%%%%%%%%
\subsection{Spark simulator}
\label{s:simulator}
\name's training happens offline using a faithful simulator that has access to
profiling information (\eg task durations) from a real Spark
cluster~(\S\ref{s:eval-real-cluster}) and the job run time characteristics
from an industrial trace~(\S\ref{s:multi-dim}).
To faithfully simulate how \name's decisions interact with a cluster, our
simulator captures several real-world effects:
\begin{CompactEnumerate}
  \item The first ``wave'' of tasks from a particular stage often runs slower
    than subsequent tasks.
    This is due to Spark's pipelined task execution~\cite{ousterhout-nsdi15},
    JIT compilation~\cite{jit} of task code, and warmup costs (\eg making TCP
    connections to other executors).
    \name's simulated environment thus picks the actual runtime of first-wave tasks
    from a different distribution than later waves.
  \item Adding an executor to a Spark job involves launching a JVM process,
    which takes 2--3 seconds.
    Executors are tied to a job for isolation and because Spark assumes them
    to be long-lived.
    \name's environment therefore imposes idle time reflecting the startup delay
    every time \name moves an executor across jobs.
  \item A high degree of parallelism can \emph{slow down} individual Spark tasks,
    as wider shuffles require additional TCP connections and create more work
    when merging data from many shards.
    \name's environment captures these effects by sampling task durations from
    distributions collected at different levels of parallelism if this data is
    available.
\end{CompactEnumerate}
%
%The \name agent is initially unaware of these effects, it learns to anticipate
%them in its decisions durings training.
%
In Appendix~\ref{s:sim-fidelity}, we validate the fidelity of our simulator by
comparing it with real Spark executions.
%~(\eg Figure~\ref{f:motivation-parallelism}).

\section{Evaluation}
\label{s:eval}
We evaluated \name on a real Spark cluster testbed and in
simulations with a production workload from Alibaba.
Our experiments address the following questions:
\begin{CompactEnumerate}
  \item How does \name perform compared to carefully-tuned heuristics in
    a real Spark cluster (\S\ref{s:eval-real-cluster})?
%   We find that \name outperforms all considered schemes in both batch and
%   streaming job arrivals, achieving $21\%$ to $3.1\times$ lower average JCTs.
%   Drill-down analysis also shows the circumstances \name finds most room
%   for improvement during streaming and unveils the corresponding learned policies.
 %
  \item Can \name's learning generalize to a multi-resource setting with
    different machine configurations (\S\ref{s:multi-dim})?
%   In an augmented Spark environment with production workloads,
%   \name uses resources efficiently
%   and outperforms state-of-the-art scheduling heuristics like Graphene~\cite{graphene}
%   by 32\%. In this complex environment, we find that \name better balances different
%   scheduling factors such as resource fragmentation, packing and prioritizing small jobs,
%   to maintain a larger cluster throughput.
 %
  \item How does each of our key ideas contribute to \name's performance;
  how does \name adapt when scheduling environments change; and how fast does \name train and
  make scheduling decisions after training?
  %We answer these questions in details with
  %a series of deep-dive microbenchmark experiments (\S\ref{s:eval-ubenches}).

  % and how robust are its learned policies to differences between training
  % and deployment environments?
  % We find that \name learns a range of qualitatively different policies for
  % different goals, and that all key ideas matter to achieving good performance
  % in practice.
  %, and that its polices are robust,
  % scheduling \eg a 15$\times$ larger cluster than it trained on with an
  % 8$\%$ performance drop
\end{CompactEnumerate}

\begin{figure}[t]
  \centering
\begin{subfigure}[t]{0.29\textwidth}
  \centering
  \includegraphics[width=\textwidth]{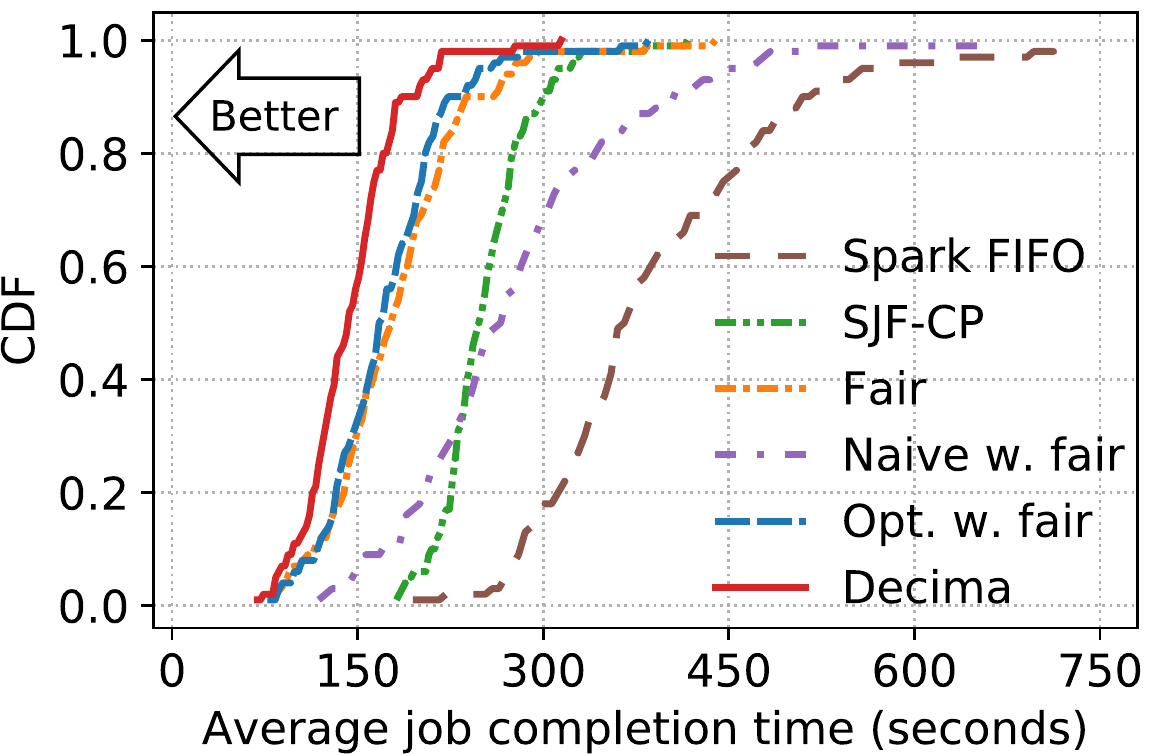}
  \vspace{-0.45cm}
  \caption{Batched arrivals.}
  \label{f:eval-spark-tpch-batch-cdf}
\end{subfigure}
\begin{subfigure}[t]{0.17\textwidth}
  \centering
  \includegraphics[width=\textwidth]{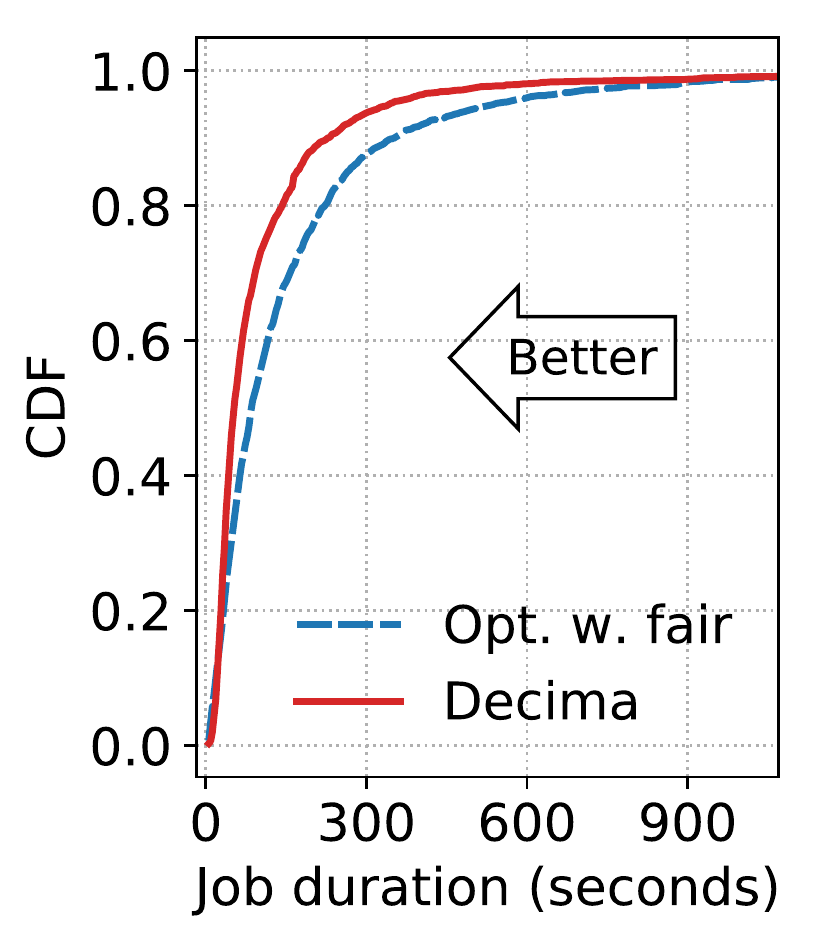}
  \vspace{-0.45cm}
  \caption{Continuous arrivals.}
  \label{f:eval-spark-tpch-stream-cdf}
\end{subfigure}
\vspace{-0.3cm}
\caption{\name's learned scheduling policy achieves 21\%--3.1$\times$ lower average
	job completion time than baseline algorithms for batch and continuous arrivals of TPC-H
    jobs in a real Spark cluster.}
  \label{f:eval-spark-tpch-cdf}
\vspace{-0.5cm}
\end{figure}

%%%%%%%%%%%%%%%%%%%%%%%%%%%%%%%%%%%%%%%%%%%%%%%%%%%%%%%%%%%%%%%%%%%%%%%%%%%%%%%
\subsection{Existing baseline algorithms}
\label{s:eval-comparing-schemes}
In our evaluation, we compare \name's performance to that
of seven baseline algorithms:
\begin{CompactEnumerate}
  \item Spark's default FIFO scheduling, which runs jobs in
  the same order they arrive in and grants as many executors to each job
  as the user requested.
  \item A shortest-job-first critical-path heuristic (SJF-CP), which
  prioritizes jobs based on their total work, and
  within each job runs tasks from the next stage on its critical path.
  \item Simple fair scheduling, which gives each job an equal fair share
  of the executors and round-robins over tasks from runnable stages
  to drain all branches concurrently.
  \item Naive weighted fair scheduling, which assigns executors
  to jobs proportional to their total work.
  \item A carefully-tuned weighted fair scheduling that gives each
  job $T_i^\alpha / \sum_i T_i^\alpha$ of total executors, where $T_i$
  is the total work of each job $i$ and $\alpha$ is a tuning factor.
  Notice that $\alpha=0$ reduces to the simple fair scheme, and
  $\alpha=1$ to the naive weighted fair one. We sweep through
  $\alpha\in \{-2, -1.9, ..., 2\}$ for the optimal factor.
  \item The standard multi-resource packing algorithm from
  Tetris \cite{tetris}, which greedily schedules the stage that
  maximizes the dot product of the requested resource vector and the
  available resource vector.
  \item \Graphene, an adaptation of Graphene~\cite{graphene} for
  \name's discrete executor classes. \Graphene detects and groups
  ``troublesome'' nodes using Graphene's algorithm~\cite[\S4.1]{graphene},
  and schedules them together with optimally tuned parallelism as in (5),
%  but keeps them unscheduled until \emph{all} troublesome nodes are
%  runnable,
  achieving the essence of Graphene's planning strategy.
  \update{We perform a grid search to optimize for the hyperparameters
  (details in Appendix~\ref{s:graphene}).}
\end{CompactEnumerate}
%

%%%%%%%%%%%%%%%%%%%%%%%%%%%%%%%%%%%%%%%%%%%%%%%%%%%%%%%%%%%%%%%%%%%%%%%%%%%%%%%
\subsection{Spark cluster}
\label{s:eval-real-cluster}
\begin{figure}[t]
  \centering
  \includegraphics[width=1.02\columnwidth]{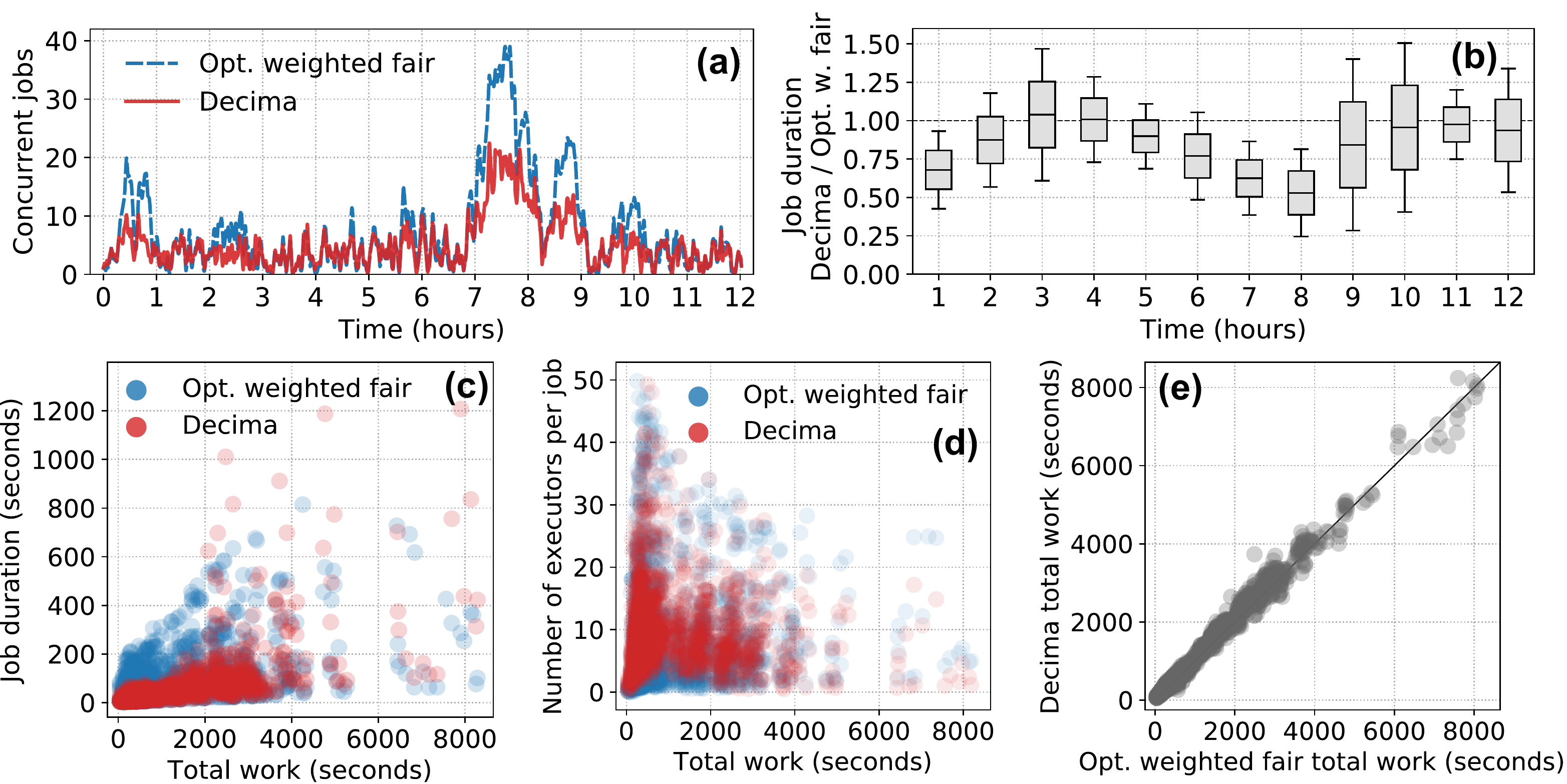}
  \vspace{-0.6cm}
  \caption{Time-series analysis (a, b) of continuous TPC-H job arrivals
  to a Spark cluster shows that \name achieves most performance
  gains over heuristics during busy periods
  (\eg runs jobs $2\times$ faster during hour 8), as it appropriately prioritizes small jobs (c) with more executors (d), while preventing work inflation (e). }
  \label{f:eval-spark-tpch-stream-analysis}
\vspace{-0.3cm}
\end{figure}
We use an OpenStack cluster running Spark v2.2, modified as described in
\S\ref{s:impl-spark}, in the Chameleon Cloud testbed.
\footnote{\url{https://www.chameleoncloud.org}}
The cluster consists of 25 worker VMs, each running two executors on an
\code{m1.xlarge} instance (8 CPUs, 16 GB RAM) and a master VM on an
\code{m1.xxxlarge} instance (16 CPUs, 32 GB RAM).
%
%We run Spark v2.2, modified as per \S\ref{s:impl-spark}, on a 26-VM cluster in
%the Chameleon Cloud testbed\footnote{\url{https://www.chameleoncloud.org}}.
%Each worker VMs runs two executors on an \code{m1.xlarge} instance (8
%CPUs, 16 GB RAM) and the master VM on an \code{m1.xxxlarge} instance (16
%CPUs, 32 GB RAM).
%
Our experiments consider \one{}
\emph{batched} arrivals, in which multiple jobs start
at the same time and run until completion, and \two{} \emph{continuous}
arrivals, in which jobs arrive with stochastic interarrival distributions
or follow a trace.
%
%
%\begin{figure*}
%  \centering
%  \includegraphics[width=2\columnwidth]{figures/eval/real_spark_tpch.pdf}
%  \vspace{-0.4cm}
%  \caption{\name's learned scheduling policy achieves
%    21\%--3.1$\times$ lower average job completion time than baseline
%    algorithms for batch and continuous arrivals of TPC-H
%    jobs in a real Spark cluster.}
%  \label{f:eval-real-spark}
%\end{figure*}

\para{Batched arrivals.}
We randomly sample jobs from six different input sizes (2, 5, 10, 20, 50, and 100
GB) and all 22 TPC-H~\cite{tpch} queries, producing a heavy-tailed distribution:
$23\%$ of the jobs contain $82\%$ of the total work.
A combination of 20 random jobs (unseen in training) arrives as a batch,
%at the start of the experiment,
and we measure their average JCT.

Figure~\ref{f:eval-spark-tpch-batch-cdf} shows a cumulative distribution of the
average JCT over 100 experiments.
There are three key observations from the results.
First, SJF-CP and fair scheduling, albeit simple, outperform the
FIFO policy by 1.6$\times$ and 2.5$\times$ on average.
%
%This is expected, since these heuristics prioritize shorter jobs and
%constrain the parallelism of each job, while FIFO grants the entire cluster.
%
%
Importantly, the fair scheduling policies outperform SJF-CP since they work on
multiple jobs, while SJF-CP focuses all executors exclusively on the shortest job.

Second, perhaps surprisingly, unweighted fair scheduling outperforms
fair scheduling weighted by job size (``naive weighted fair'').
This is because weighted fair scheduling grants small jobs \emph{fewer}
executors than their fair share, slowing them down and increasing average JCT.
Our tuned weighted fair heuristic (``opt.\ weighted fair'') counters this
effect by calibrating the weights for each job \emph{on each experiment}
(\S\ref{s:eval-comparing-schemes}).
The optimal $\alpha$ is usually around $-1$, \ie the heuristic sets the number of executors
inversely proportional to job size.
This policy effectively focuses on small jobs early on, and later shifts to running
large jobs in parallel; it outperforms fair scheduling by $11\%$.
%
%These results illustrate the difficulty of balancing the conflicting goals
%between prioritizing small jobs to improve average JCT, and dividing the cluster
%correctly to run jobs at efficient parallelism levels.
%
% Picking the right
% heuristic and setting the right workload-dependent weights is
% difficult, tedious and results in a brittle policy.

Finally, \name outperforms all baseline algorithms and improves the
average JCT by $21\%$ over the closest heuristic (``opt.\ weighted fair'').
This is because \name prioritizes jobs better, assigns efficient
executor shares to different jobs, and leverages the job DAG structure
(\S\ref{s:eval-ubenches} breaks down the benefit of each of these factors).
\name autonomously learns this policy through end-to-end RL training, while
the best-performing baseline algorithms required careful tuning.
%
%\begin{figure}
%  \centering
%  \includegraphics[width=1.05\columnwidth]{figures/eval/real_spark_tpch_stream_abc.pdf}
%  \vspace{-0.75cm}
%  \caption{Streaming job arrival of 1,000 TPC-H jobs over 12 hours in a real Spark
%    cluster. \name achieves $29\%$ better average JCT than the best heuristic
%    (other heuristics cannot keep up) and has fewer active jobs at most points in
%    time.}
%  \label{f:eval-spark-tpch-stream}
%\end{figure}

\begin{figure}[t]
  \centering
\begin{subfigure}[t]{0.215\textwidth}
  \centering
  \includegraphics[width=\textwidth]{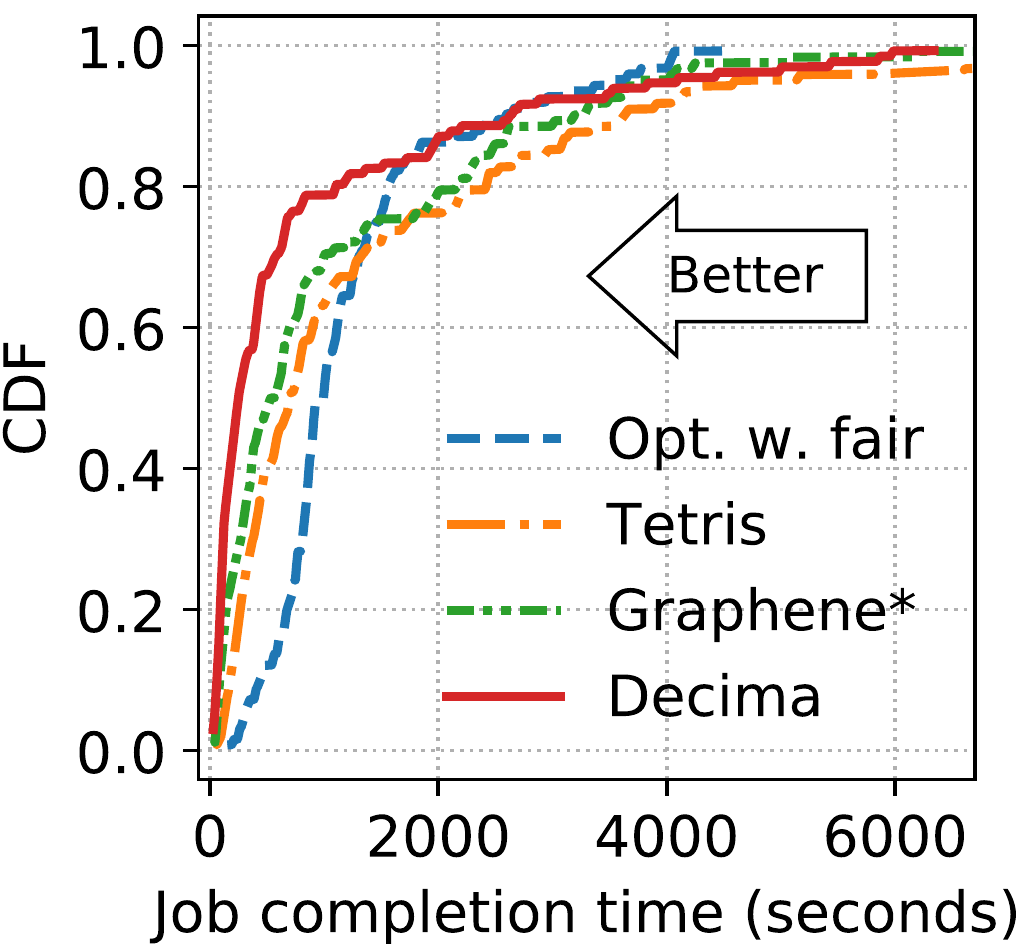}
  \vspace{-0.4cm}
  \caption{Industrial trace replay.}
  \label{f:eval-alibaba-multi-resource-cdf}
\end{subfigure}
\hspace{0.03cm}
\begin{subfigure}[t]{0.215\textwidth}
  \centering
  \includegraphics[width=\textwidth]{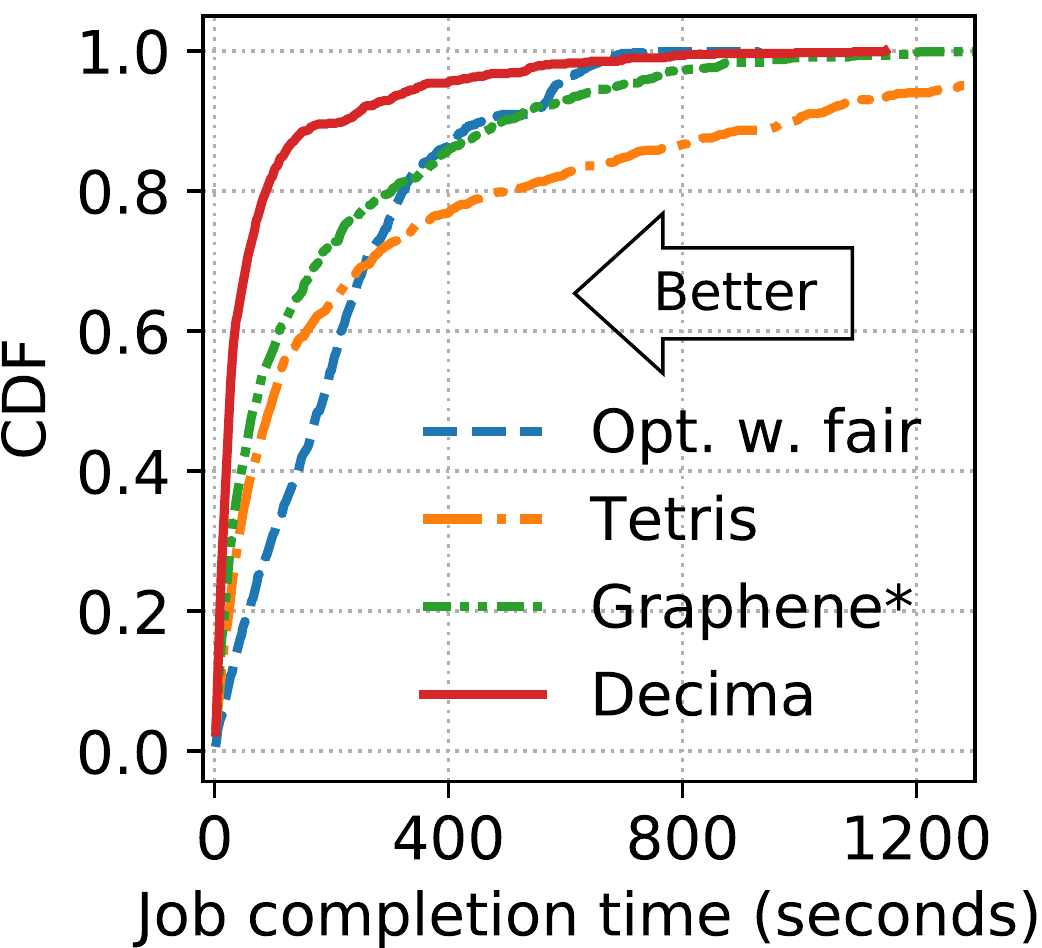}
  \vspace{-0.4cm}
  \caption{TPC-H workload.}
  \label{f:eval-tpch-multi-resource-cdf}
\end{subfigure}
\vspace{-0.3cm}
\caption{With multi-dimensional resources, \name's scheduling policy outperforms
  \Graphene by $32\%$ to $43\%$ in average JCT.}
  \label{f:eval-multi-resource-cdf}
\vspace{-0.4cm}
\end{figure}

\para{Continuous arrivals.}
%
%\name trains with the techniques described in \S\ref{s:training}, and we
%evaluate it with an unseen sequence of job arrivals.
%
We sample 1,000 TPC-H jobs of six different sizes \update{uniformly at random}, and model their
arrival as a Poisson process with an average interarrival time of 45 seconds.
The resulting cluster load is about $85\%$.
%
%We record all job durations, and, in 10-second intervals, the concurrent
%number of jobs and the executor usage.
%
%
%During a busy period 8 hours into the experiment,
At this cluster load, jobs arrive faster than
most heuristic-based scheduling policies can complete them.
Figure~\ref{f:eval-spark-tpch-stream-cdf} shows that \name outperforms the
only baseline algorithm that can keep up (``opt.\ weighted fair''); \name's
average JCT is $29\%$ lower.
In particular, \name shines during busy, high-load periods, where scheduling
decisions have a much larger impact than when cluster resources are abundant.
Figure~\ref{f:eval-spark-tpch-stream-analysis}{\color{darkred}a} shows that
\name maintains a lower concurrent job count than the tuned heuristic
particularly during the busy period in hours 7--9, where \name completes
jobs about $2\times$ faster (Figure~\ref{f:eval-spark-tpch-stream-analysis}{\color{darkred}b}).
\update{
Performance under high load is important for batch processing clusters, which often have long job queues~\cite{yaq}, and periods of high load are when good scheduling decisions have the most impact (\eg reducing the overprovisioning required for workload peaks).}

\name's performance gain comes from finishing small jobs faster, as the
concentration of red points in the lower-left corner of
Figure~\ref{f:eval-spark-tpch-stream-analysis}{\color{darkred}c} shows.
%
%while not
%starving the large jobs.
%
\name achieves this by assigning more executors to the small jobs
(Figure~\ref{f:eval-spark-tpch-stream-analysis}{\color{darkred}d}).
The right number of executors for each job is workload-dependent:
indiscriminately giving small jobs more executors would use cluster
resources inefficiently (\S\ref{s:motiv-parallelism}).
For example, SJF-CP's strictly gives all available executors to the
smallest job, but this inefficient use of executors inflates total
work, and SJF-CP therefore accumulates a growing backlog of jobs.
\name's executor assignment, by contrast, results in similar total
work as with the hand-tuned heuristic.
Figure~\ref{f:eval-spark-tpch-stream-analysis}{\color{darkred}e}
shows this: jobs below the diagonal have smaller total work with
\name than with the heuristic, and ones above have larger total work
in \name.
Most small jobs are on the diagonal, indicating that \name only
increases the parallelism limit when extra executors are still efficient.
Consequently, \name successfully balances between giving small jobs
extra resources to finish them sooner and using the resources
efficiently.

%%
%%\name outperforms the tuned heuristic particularly during the busy period in
%%hours 7--9, maintaining a lower active job count as it completes jobs sooner.
%%
%The key difference in policy is that \name assigns more executors to each job
%during the busy period
%(Figure~\ref{f:eval-spark-tpch-stream-analysis}{\color{darkred}c}).
%%
%In other words, instead of dividing executors between runnable jobs according
%to their fair shares, \name ``unfairly'' prioritizes some jobs and grants them
%more executors.
%%
%As a result, these jobs finish sooner, reducing the average JCT.
%%
%The other, temporarily-starved jobs start later, but receive many executors once
%they do run.
%%
%Consequently, these jobs' final completion time is unchanged, even though their
%start was delayed.
%%
%%Therefore, appropriately prioritizing small jobs significantly reduces average
%%JCT and improves cluster throughput.
%\name sees a 45\% shorter job durations in hour 8 of the busy
%period~(Figure~\ref{f:eval-spark-tpch-stream-analysis}{\color{darkred}d}).
%%
%Therefore, \name learns efficient parallelism levels during periods of high
%load, when getting them right matters the most.
%

%%%%%%%%%%%%%%%%%%%%%%%%%%%%%%%%%%%%%%%%%%%%%%%%%%%%%%%%%%%%%%%%%%%%%%%%%%%%%%%
\subsection{Multi-dimensional resource packing}
\label{s:multi-dim}
\begin{figure}[t]
  \centering
\begin{subfigure}[t]{0.23\textwidth}
  \centering
  \includegraphics[width=\textwidth]{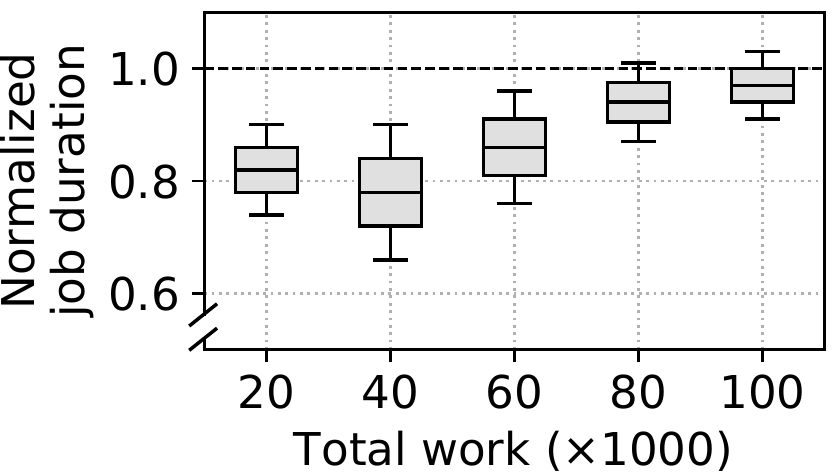}
  \vspace{-0.5cm}
  \caption{Job duration grouped by total work, \name normalized to \Graphene.}
  \label{f:eval-alibaba-multi-resource-job-duration}
\end{subfigure}
\hspace{0.02cm}
\begin{subfigure}[t]{0.23\textwidth}
  \centering
  \includegraphics[width=\textwidth]{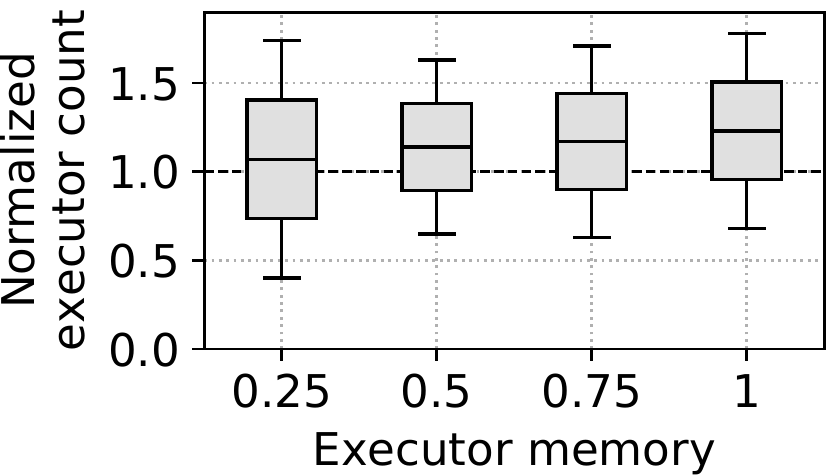}
  \vspace{-0.5cm}
  \caption{Number of executors that \name uses for ``small'' jobs,
  		   normalized to \Graphene.}
  \label{f:eval-tpch-multi-resource-exec-usage}
\end{subfigure}
\vspace{-0.3cm}
\caption{\name outperforms \Graphene with multi-dimensional resources by (a) completing
  small jobs faster and (b) use ``oversized'' executors for small jobs (smallest $20\%$
  in total work).}
  \label{f:eval-multi-resource-cdf-drill-down}
\vspace{-0.4cm}
\end{figure}
%
%\begin{figure*}[t]
%\centering
%%
%\begin{subfigure}[t]{0.488\textwidth}
%  \centering
%  \includegraphics[width=\textwidth]{figures/eval/tpch-multi-resource-stream.pdf}
%  \vspace{-0.6cm}
%  \caption{TPC-H workload.}
%  \label{f:eval-multi-dimension-tpch}
%\end{subfigure}
%%
%\hspace{0.023\textwidth}
%%
%\begin{subfigure}[t]{0.472\textwidth}
%  \centering
%  \includegraphics[width=\textwidth]{figures/eval/alibaba-multi-resource-stream.pdf}
%  \vspace{-0.6cm}
%  \caption{Industrial trace replay.}
%  \label{f:eval-multi-dimension-alibaba}
%\end{subfigure}
%%
%\vspace{-0.2cm}
%%
%\caption{In a multi-resource scheduling environment, \name learns a policy that
%  outperforms \Graphene by $32\%$ to $43\%$ in average JCT. ``Memory usage''
%  (task memory$/$executor memory) indicates how much memory is fragmented;
%  \name performs similarly to \Graphene. The ``slowdown''
%  of a job is the JCT normalized to its JCT on an otherwise idle cluster; \name
%  consistently has the lowest slowdown. ``Small jobs'' means the smallest $20\%$ in
%  total work of all jobs; \name finishes them quicker by more aggressively using executors
%  with large memory.
%  }
%\label{f:eval-multi-dimension}
%\end{figure*}
%
The standalone Spark scheduler used in our previous experiments only provides
jobs with access to predefined executor slots.
More advanced cluster schedulers, such as YARN~\cite{yarn} or
Mesos~\cite{mesos}, allow jobs to specify their tasks' resource
requirements and create appropriately-sized executors.
%
%Many cluster managers allow users to specify task resource requests using
Packing tasks with multi-dimensional resource needs (\eg $\langle$CPU,
memory$\rangle$) onto fixed-capacity servers adds further complexity to the
scheduling problem~\cite{tetris, graphene}.
We use a production trace from Alibaba to investigate if \name can
learn good multi-dimensional scheduling policies with the same core approach.

\begin{figure*}[t]
\centering
\begin{subfigure}[t]{0.325\textwidth}
  \centering
  \includegraphics[width=\textwidth]{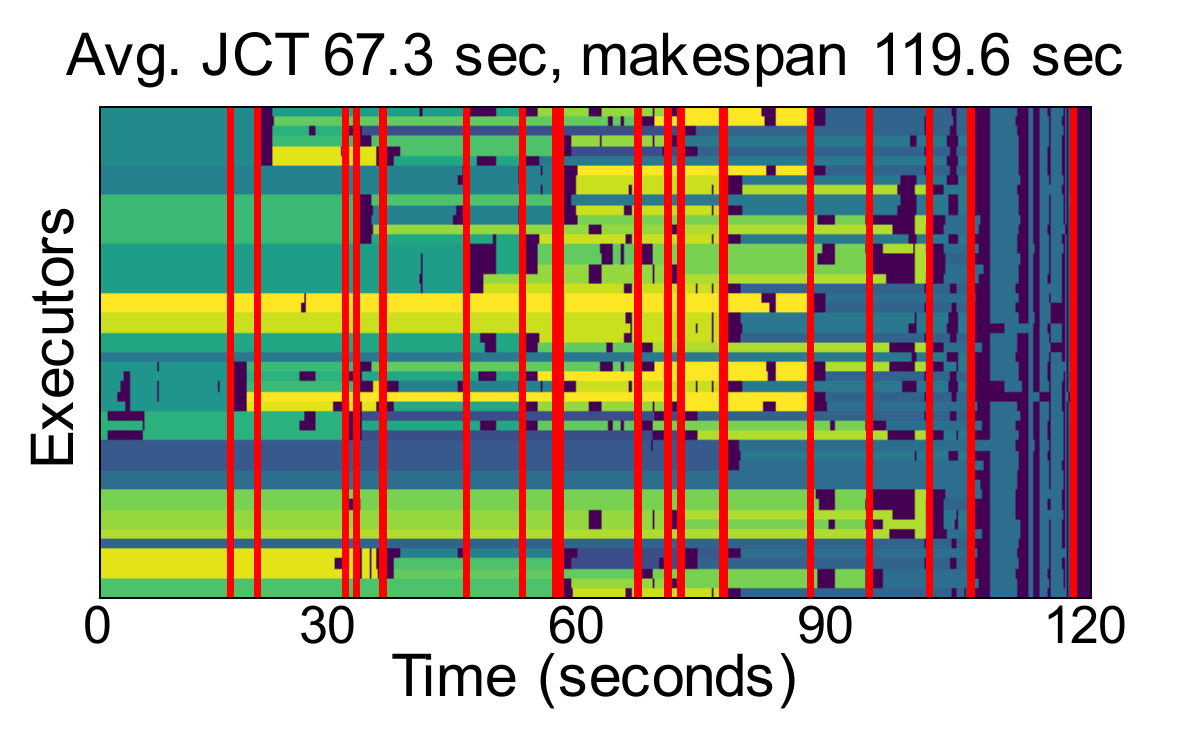}
  \vspace{-0.6cm}
  \caption{Average JCT objective.}
  \label{f:eval-spark-examples-avg-jct}
\end{subfigure}
\begin{subfigure}[t]{0.325\textwidth}
  \centering
  \includegraphics[width=\textwidth]{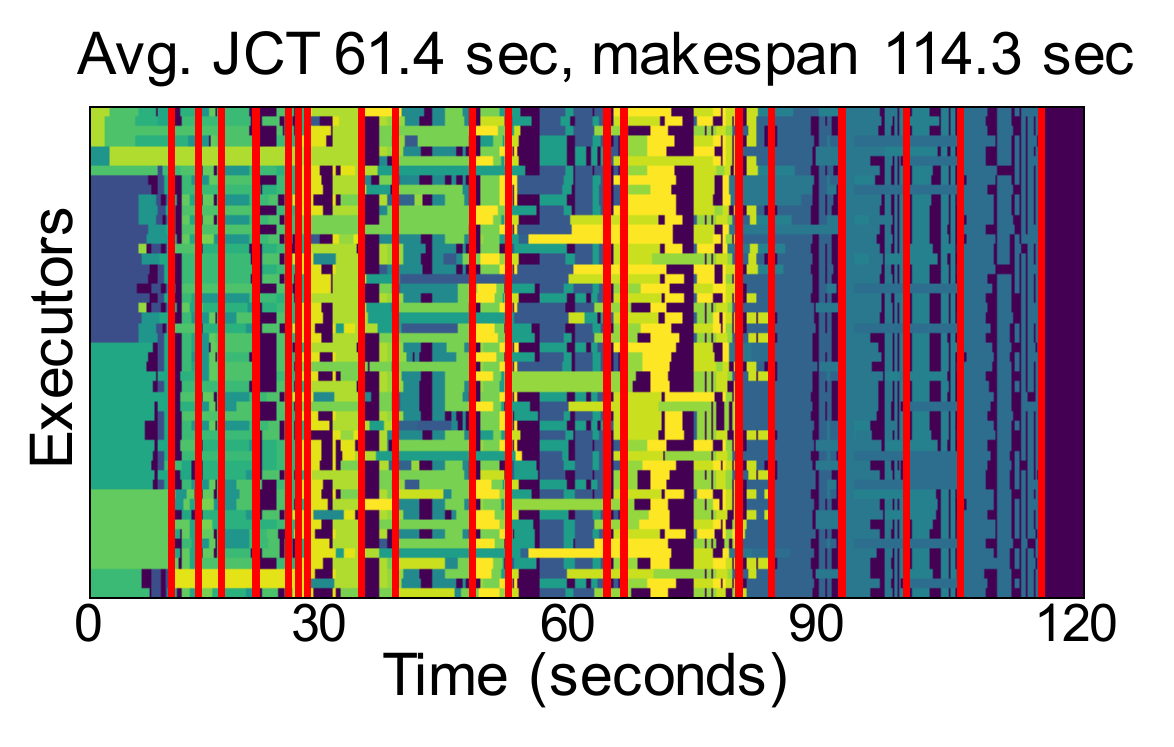}
  \vspace{-0.6cm}
  \caption{Avg.\ JCT, with zero-cost executor motion.}
  \label{f:eval-spark-examples-no-moving-cost}
\end{subfigure}
\begin{subfigure}[t]{0.325\textwidth}
  \centering
  \includegraphics[width=\textwidth]{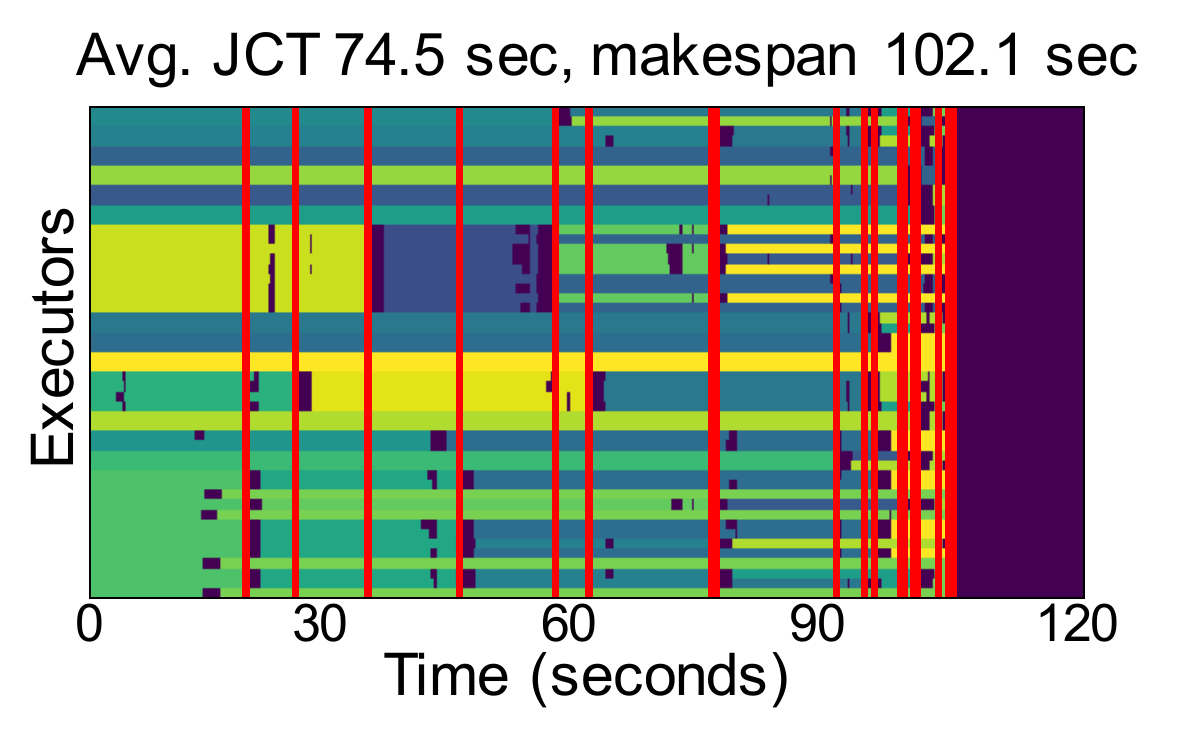}
  \vspace{-0.6cm}
  \caption{Minimal makespan objective.}
  \label{f:eval-spark-examples-makespan}
\end{subfigure}
\vspace{-0.3cm}
\caption{\name learns qualitatively different policies depending on the environment
  (\eg costly (\subref{f:eval-spark-examples-avg-jct}) vs.\ free executor migration
  (\subref{f:eval-spark-examples-no-moving-cost})) and the objective (\eg average JCT
  (\subref{f:eval-spark-examples-avg-jct})  vs.\ makespan
  (\subref{f:eval-spark-examples-makespan})). Vertical red lines indicate job completions,
  colors indicate tasks in different jobs, and dark purple is idle time.}
\label{f:eval-spark-examples}
\vspace{-0.3cm}
\end{figure*}

\para{Industrial trace.}
The trace contains about 20,000 jobs from a production cluster.
Many jobs have complex DAGs: 59\% have four or more stages, and some have hundreds.
We run the experiments using our simulator (\S\ref{s:simulator}) with up to
30,000 executors.
This parameter is set according to the maximum number of concurrent tasks in the trace.
\update{We use the first half of the trace for training and then compare \name's performance with other schemes on the remaining portion.}
%Since the trace contains no information about cluster capacity, we use the
%maximum number of concurrently active tasks (27,986) to guide our setup and
%simulate cluster sizes up to 30,000 executors.
%

\para{Multi-resource environment.}
We modify \name's environment to provide several discrete executor
\emph{classes} with different memory sizes.
Tasks now require a minimum amount of CPU and memory, \ie a task must fit
into the executor that runs it.
Tasks can run in executors larger than or equal to their resource request.
\name now chooses a DAG stage to schedule, a parallelism level, and an
executor class to use.
%
%The discretization to fixed executor classes is needed to maintain \name's
%discrete action space; we find that it imposes only a moderate overhead
%over continuous resource allocations of arbitrary sizes.
%
Our experiments use four executor types, each with $1$ CPU core and $(0.25, 0.5,
0.75, 1)$ unit of normalized memory; each executor class makes up 25\% of total
cluster executors.
\para{Results.}
We run simulated multi-resource experiments on continuous job arrivals according
to the trace.
Figure~\ref{f:eval-alibaba-multi-resource-cdf} shows the results for \name and
three other algorithms: the optimally tuned weighted-fair heuristic, Tetris, and
\Graphene.
\name achieves a $32\%$ lower average JCT than the best competing algorithm
(\Graphene), suggesting that it learns a good policy in the multi-resource
environment.

\name's policy is qualitatively different to \Graphene's.
Figure~\ref{f:eval-alibaba-multi-resource-job-duration} breaks \name's improvement
over \Graphene down by jobs' total work.
\name completes jobs faster than \Graphene for all job sizes, but its gain is
particularly large for small jobs.
The reason is that \name learns to use ``oversized'' executors when they can help
finish nearly-completed small jobs when insufficiently many right-sized executors
are available.
Figure~\ref{f:eval-tpch-multi-resource-exec-usage} illustrates this:
\name uses $39\%$ more executors of the largest class on the jobs
with smallest $20\%$ total work (full profiles in
Appendix~\ref{s:multi-resource-more-analysis}).
In other words, \name trades off memory fragmentation against clearing the job
queue more quickly.
This trade-off makes sense because small jobs \one{} contribute
more to the average JCT objective, and \two{} only fragment resources for
a short time.
By contrast, Tetris greedily packs tasks into the best-fitting executor class
and achieves the lowest memory fragmentation.
\name's fragmentation is within $4\%$--$13\%$ of Tetris's, but \name's average
JCT is $52\%$ lower, as it learns to balance the trade-off well.
%
%However, this can result in a suboptimal packing in time: jobs in need of
%executors from a class in high demand need to wait.
%%
%This increases \emph{job slowdown}, which normalizes the JCT to the job's
%runtime on an idle cluster on which all executors are available
%(Figure~\ref{f:eval-multi-dimension}{\color{darkred}a3} and
%\ref{f:eval-multi-dimension}{\color{darkred}b3}).
%
This requires respecting workload-dependent factors, such as the DAG
structure, the threshold for what is a ``small'' job, and others.
Heuristic approaches like \Graphene attempt to balance those
factors via additive score functions and extensive tuning, while
\name learns them without such inputs.
We also repeat this experiment with the TPC-H workload, using 200
executors and sampling each TPC-H DAG node's memory request from $(0, 1]$.
Figure~\ref{f:eval-tpch-multi-resource-cdf} shows that \name outperforms
the competing algorithms by even larger margins (\eg $43\%$ over \Graphene).
%
%\name consistently performs better on TPC-H workloads than on the industrial
%trace because we
This is because the industrial trace lacks work inflation measurements for
different levels of parallelism, which we provide for TPC-H.
\name learns to use this information to further calibrate executor assignment. %than with the industrial trace.
%

%%%%%%%%%%%%%%%%%%%%%%%%%%%%%%%%%%%%%%%%%%%%%%%%%%%%%%%%%%%%%%%%%%%%%%%%%%%%%%%
\subsection{\name deep dive}
\label{s:eval-ubenches}
Finally, we demonstrate the wide range of scheduling policies \name
can learn, and break down the impact of our key ideas and techniques
on \name's performance.
In appendices, we further evaluate \name's
optimality via an exhaustive search of job orderings~(Appendix~\ref{s:optimality}),
the robustness of its learned policies to
changing environments~(Appendix~\ref{s:generalizability}), and
\name's sensitivity to incomplete information~(Appendix~\ref{s:incomplete-info}).

\para{Learned policies.}
\name outperforms other algorithms because it can learn different policies
depending on the high-level objective, the workload, and environmental conditions.
When \name optimizes for average JCT
(Figure~\ref{f:eval-spark-examples-avg-jct}), it learns to share executors for
small jobs to finish them quickly and avoids inefficiently using too many executors on large
jobs~(\S\ref{s:eval-real-cluster}).
\name also keeps the executors working on tasks from the same job to avoid the
overhead of moving executors (\S\ref{s:impl-spark}).
However, if moving executors between jobs is free\,---\,as is effectively the
case for long tasks, or for systems without JVM spawn overhead\,---\, \name
learns a policy that eagerly moves executors among jobs (cf.\ the
frequent color changes in Figure~\ref{f:eval-spark-examples-no-moving-cost}).
Finally, given a different objective of minimizing the overall \emph{makespan}
for a batch of jobs, \name learns yet another different policy
(Figure~\ref{f:eval-spark-examples-makespan}).
Since only the \emph{final} job's completion time matters for a makespan
objective, \name no longer works to finish jobs early.
Instead, many jobs complete together at the end of the batched workload, which gives
the scheduler more choices of jobs throughout the execution, increasing cluster
utilization.

%
%\name learns these qualitatively different policies automatically and without
%human intervention, purely by interacting with the environment.
%
%Achieving the same generality with heuristics and without extensive hand-tuning
%is tedious.
%

\para{Impact of learning architecture.}
We validate that \name uses all raw information provided in the state
and requires all its key design components by selectively omitting
components.
We run 1,000 continuous TPC-H job arrivals %(as in~\S\ref{s:eval-real-cluster})
on a simulated cluster at different loads, and train five different variants of
\name on each load.
%, removing a different component in each.
%

\begin{figure}
  \centering
  \includegraphics[width=0.9\columnwidth]{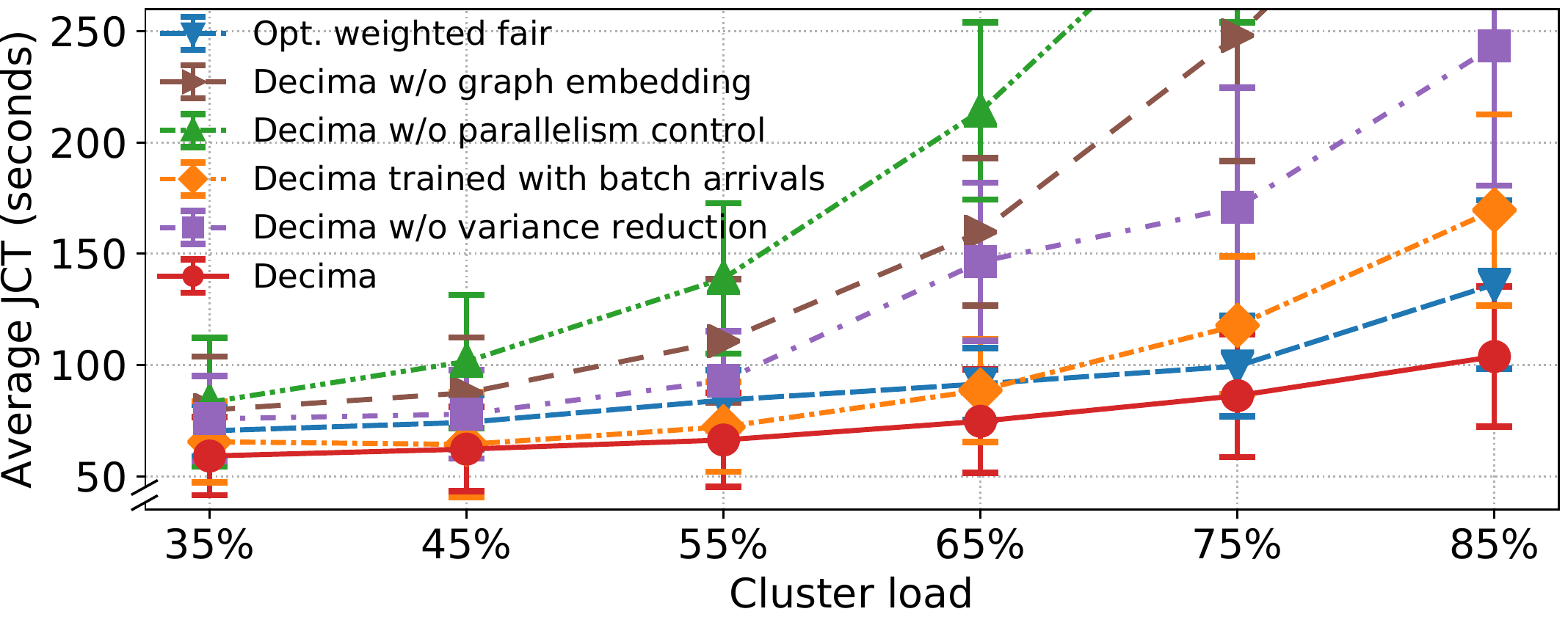}
  \vspace{-0.3cm}
\caption{Breakdown of each key idea's contribution to \name with continuous
  job arrivals. Omitting any concept increases \name's average JCT above that
  of the weighted fair policy.}
  \label{f:eval-ubench-component-breakdown}
  \vspace{-0.3cm}
\end{figure}

%
Figure~\ref{f:eval-ubench-component-breakdown} shows that removing any one
component from \name results in worse average JCTs than the tuned weighted-fair
heuristic at a high cluster load.
There are four takeaways from this result.
First, parallelism control has the greatest impact on \name's performance.
Without parallelism control, \name assigns all available executors to a
single stage at every scheduling event.
Even at a moderate cluster load (\eg 55\%), this leads to an unstable
policy that cannot keep up with the arrival rate of incoming jobs.
Second, omitting the graph embedding (\ie directly taking raw features on
each node as input to the score functions in~\S\ref{s:action}) makes
\name unable to estimate remaining work in a job and to account for other
jobs in the cluster.
Consequently, \name has no notion of small jobs or cluster load, and its
learned policy quickly becomes unstable as the load increases.
Third, using unfixed job sequences across training episodes
%disabling the synchronized termination of workload sequences during training
increases the variance in the reward signal~(\S\ref{s:training}).
As the load increases, job arrival sequences become more varied, which
increases variance in the reward.
At cluster load larger than $75\%$, reducing this variance via synchronized
termination improves average JCT by $2\times$ when training \name, illustrating
that variance reduction is key to learning high-quality policies in
long-horizon scheduling problems.
Fourth, training only on batched job arrivals cannot generalize to
continuous job arrivals.
When trained on batched arrivals, \name learns to systematically defer large
jobs, as this results in the lowest sum of JCTs (lowest sum of penalties).
With continuous job arrivals, this policy starves large jobs indefinitely
as the cluster load increases and jobs arrive more frequently.
Consequently, \name underperforms the tuned weighted-fair heuristic at
loads above $65\%$ when trained on batched arrivals.

%
%\para{Scalability.}
%%
%The inference time of \name's model is crucial to scheduling latency and must
%therefore scale well with cluster size.
%%
%The original model in~\S\ref{s:action}\,---\,without our insight that
%$P(\textit{limit}|\textit{node}) = P(\textit{limit}|\textit{job})$\,---\,has an
%inference time of $O(\#\textit{nodes} \times \#\textit{executors})$.
%%
%Figure~\ref{f:eval-inference-time} shows that this grows to 400ms of scheduling
%delay for a cluster with 30k executors.
%%
%Our domain-specific insight that the parallelism level is independent of node
%choice, however, reduces the inference time to a more scalable
%$O(\#\textit{nodes} + \#\textit{executors})$. As a result, the scheduling delay is
%almost a constant (<50ms) when it scales to 30k executors.
%%

\para{Generalizing to different workloads.}
\begin{table}[t]
\footnotesize
\hspace{0cm}
\scalebox{0.95}{
%\begin{tabular}{|c|c|c|c|c|c|}
\begin{tabular}{|r|l|}
\hline
%  \makecell{\name, trained \\on same\\ workload} & \makecell{\name, trained \\on skewed\\ workload} & \makecell{\name, trained \\on mixed\\ workloads} & \makecell{\name, trained \\on mixed workloads\\ w/ load hints} & \makecell{Opt. weighted \\fair (heuristic)}  \\
  \bf Setup\; (IAT: interarrival time) & \bf Average JCT [sec] \\
\hline
Opt. weighted fair (best heuristic) & $91.2 \pm 23.5$ \\
\hline
\name, trained on test workload \,(IAT: 45 sec) & $65.4 \pm 28.7$ \\
\name, trained on anti-skewed workload \,(IAT: 75 sec) & $104.8 \pm 37.6$ \\
\name, trained on mixed workloads & $82.3 \pm 31.2$ \\
\name, trained on mixed workloads with interarrival time hints & $76.6 \pm 33.4$ \\
\hline
%$65.4 \pm 28.7$  & $104.8 \pm 37.6$ & $82.3 \pm 31.2$ & $76.6 \pm 33.4$ & $91.2 \pm 23.5$\\
%\hline
%
\end{tabular}
}
\vspace{0.1cm}
\caption{\update{\name generalizes to changing workloads. For an unseen workload, \name outperforms the best heuristic by $10\%$ when trained with a mix of workloads; and by $16\%$ if it knows the interarrival time from an input feature.}}
\label{t:generalize_workload}
\vspace{-0.8cm}
\end{table}
\update{
We test \name's ability to generalize by changing the training workload in the TPC-H experiment (\S\ref{s:eval-real-cluster}).
To simulate shifts in cluster workload, we train models for different job interarrival times between
42 and 75 seconds, and test them using a workload with a 45 second interarrival time.
As \name learns workload-specific policies, we expect its effectiveness to depend on whether broad
test workload characteristics, such as interarrival time and job size distributions, match the training
workload.

Table~\ref{t:generalize_workload} shows the resulting average JCT.
\name performs well when trained on a workload similar to the test workload.
Unsurprisingly, when \name trains with an ``anti-skewed'' workload (75 seconds interarrival time), it generalizes
  poorly and underperforms the optimized weighted fair policy.
This makes sense because \name incorporates the learned interarrival time distribution in its policy.
When training with a mixed set of workloads that cover the whole interarrival time range, \name can learn
a more general policy.
This policy fits less strongly to a specific interarrival time distribution and therefore becomes more
robust to workload changes.
If \name can observe the interarrival time as a feature in its state (\S\ref{s:impl-spark}), it
generalizes better still and learns an adaptive policy that achieves a 16\% lower average JCT than the
best heuristic.
These results highlight that a diverse training workload set helps make \name's learned policies robust
to workload shifts; we discuss possible online learning in \S\ref{s:discussion}.
}

\begin{figure}
  \centering
\begin{subfigure}[t]{0.25\textwidth}
  \centering
  \includegraphics[width=\textwidth]{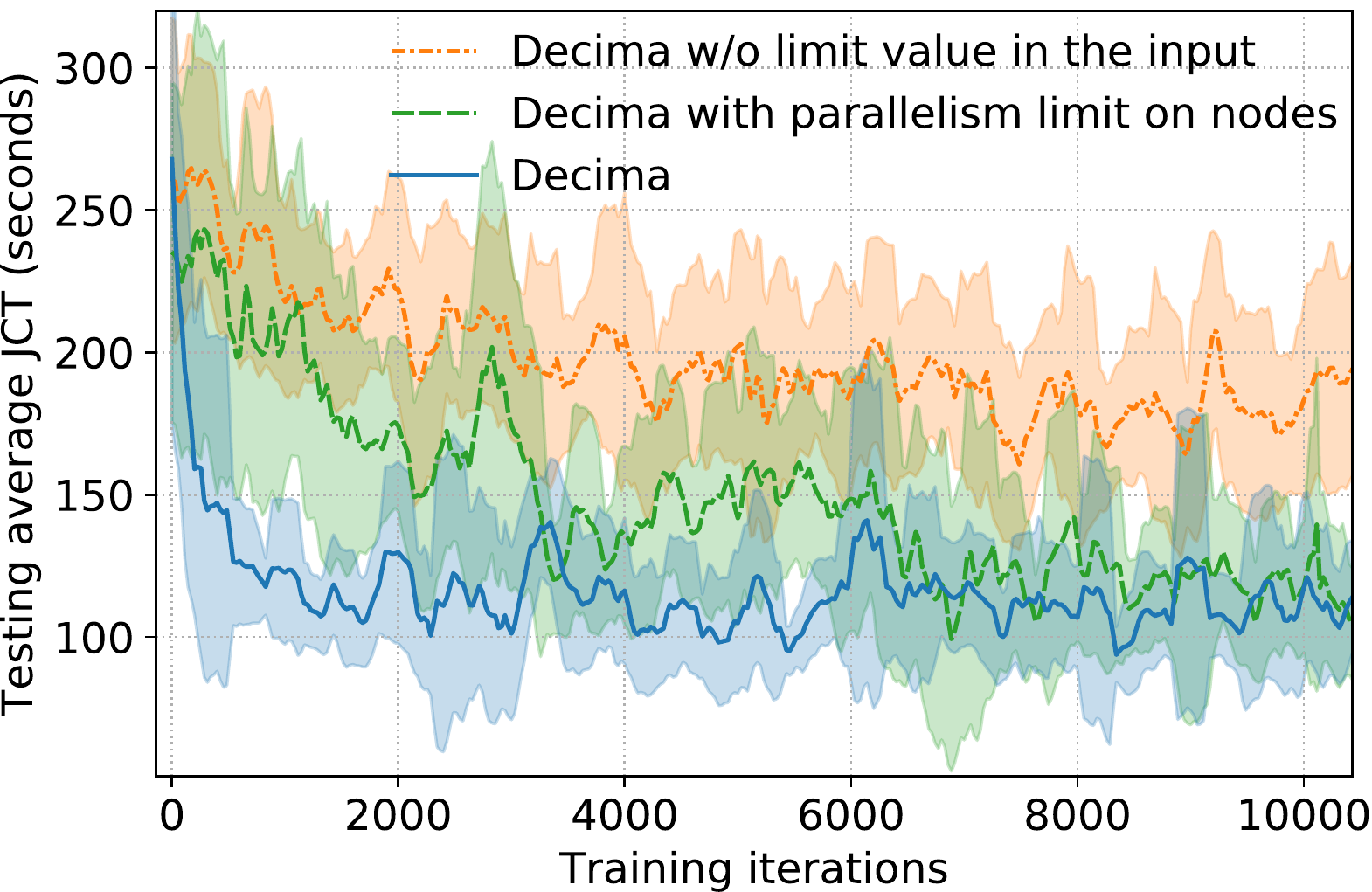}
  \vspace{-0.5cm}
  \caption{Learning curve.}
  \label{f:eval-learning-curve}
\end{subfigure}
\begin{subfigure}[t]{0.18\textwidth}
  \centering
  \includegraphics[width=\textwidth]{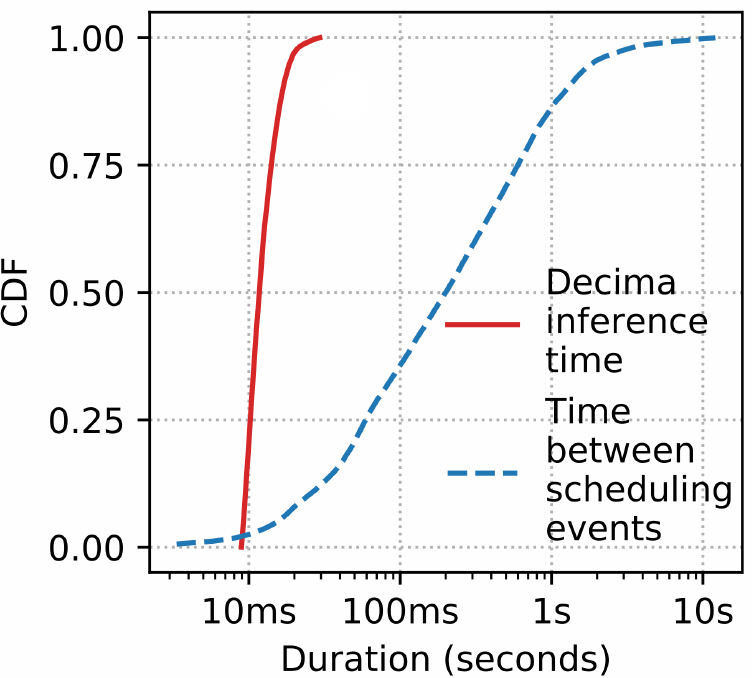}
  \vspace{-0.5cm}
  \caption{Scheduling delay.}
  \label{f:eval-inference-time}
\end{subfigure}
\vspace{-0.3cm}
 \caption{Different encodings of jobs parallelism (\S\ref{s:action}) affect \name's training time.
  \name makes low-latency scheduling decisions: on average, the latency is about 50$\times$ smaller than
  the interval between scheduling events.}
  \label{f:eval-sys-perf}
  \vspace{-0.5cm}
\end{figure}

\para{Training and inference performance.}
Figure~\ref{f:eval-learning-curve} shows \name's learning curve (in blue)
on continuous TPC-H job arrivals (\S\ref{s:eval-real-cluster}),
testing snapshots of the model every 100 iterations on (unseen) job arrival
sequences.
Each training iteration takes about 5 seconds.
\name's design (\S\ref{s:training}) is crucial for training efficiency:
omitting the parallelism limit values in the input (yellow curve) forces \name to use
separate score functions for different limits,
significantly increasing the number of parameters to optimize over;
putting fine-grained parallelism control on nodes (green curve) slows down training
as it increases the space of algorithms \name must explore.

Figure~\ref{f:eval-inference-time} shows cumulative distributions of the time
\name takes to decide on a scheduling action (in red) and the time interval between scheduling events (in blue)
%during the continuous job arrival
in our Spark testbed (\S\ref{s:eval-real-cluster}).
The average scheduling delay for \name is less than 15ms, while the interval between scheduling
events is typically in the scale of seconds.
In less than $5\%$ of the cases, the scheduling interval is shorter than the scheduling delay (\eg when the cluster requests for multiple scheduling actions in a single scheduling event).
%
%The small portion when the scheduling interval is smaller is because the cluster
%requests for consecutive scheduling decisions (\eg when previous action did not use
%all executors).
%
Thus \name's scheduling delay imposes no measurable overhead on task runtimes.
%
%Further optimization in the implementation can potentially reduce this delay since we only prototype \name scheduler in Python.

\section{Discussion}
\label{s:discussion}
\update{
In this section, we discuss future research directions and other potential
applications for \name's techniques.

\smallskip
\noindent
\textbf{Robustness and generalization.}
Our experiments in \S\ref{s:eval-ubenches} showed that \name can learn
generalizable scheduling policies that work well on an unseen workload.
However, more drastic workload changes than interarrival time shifts
could occur.
To increase robustness of a scheduling policy against such changes, it
may be helpful to train the agent on worst-case situations or
adversarial workloads, drawing on the emerging literature on robust
adversarial RL~\cite{rarl}.
Another direction is to adjust the scheduling policy \emph{online} as
the workload changes.
The key challenge with an online approach is to reduce the large sample
complexity of model-free RL when the workload changes quickly.
One viable approach might be to use meta learning~\cite{maml, rl2, mbmpo},
which allows training a ``meta'' scheduling agent that is designed to
adapt to a specific workload with only a few observations.

\smallskip
\noindent
\textbf{Other learning objectives.}
In our experiments, we evaluated \name on metrics related to job duration
(\eg average JCT, makespan).
Shaping the reward signal differently can steer \name to meet other
objectives, too.
For example, imposing a hard penalty whenever the deadline of a job is
missed would guide \name to a deadline-aware policy.
%
%Also, tail performance in job duration can
%be explicitly translated in the reward by empirically computing the
%value from samples of observed job durations.
%
%Other rewards formulations might empirically compute the large
%percentile of rewards from job durations samples to avoid a
%``long tail'' in job durations, or combine signals from batch and
%long-running service workloads.
Alternatively, basing the reward on \eg the 90th percentile of empirical job duration samples, \name can optimize for a tight tail of the JCT distribution.
Addressing objectives formulated as constrained
optimization (\eg to minimize average JCT, but strictly guarantee
fairness) using RL is an interesting further
direction~\cite{constrained_po, mdp_constrained}.

\smallskip
\noindent
\textbf{Preemptive scheduling.}
\name currently never preempts running tasks and can only remove
executors from a job after a stage completes.
This design choice keeps the MDP tractable for RL and results in
effective learning and strong scheduling policies. %(\S\ref{s:eval}).
However, future work might investigate more fine-grained and reactive
preemption in an RL-driven scheduler such as \name.
Directly introducing preemption would lead to a much larger action space
(\eg specifying arbitrary set of executors to preempt) and might require
a much higher decision-making frequency.
To make the RL problem tractable, one potential research
direction is to leverage multi-agent RL~\cite{multi-agent-rl-survey-1, cooperative-multi-agent, berkeley-multi-agent}.
For example, a \name-like scheduling agent might controls which stage to
run next and how many executors to assign, and, concurrently, another
agent might decide where to preempt executors.

\smallskip
\noindent
\textbf{Potential networking and system applications.}
Some techniques we developed for \name are broadly applicable to other networking and computer systems problems.
For example, the scalable representation of input DAGs (\S\ref{s:graph}) has applications in problems over graphs, such as database query optimization~\cite{neo} and hardware device placement~\cite{placeto}.
Our variance reduction technique (\S\ref{s:training}) generally applies to systems with stochastic, unpredictable
inputs~\cite{variance-reduction, abrl}.
}

\section{Related Work}
\label{s:related}
There is little prior work on applying machine learning
techniques to cluster scheduling.
DeepRM~\cite{deeprm}, which uses RL
to train a neural network for multi-dimensional resource
packing, is closest to \name in aim and approach.
\update{However, DeepRM only deals with a basic setting in which each job is a single
task and was evaluated in simple, simulated environments.
DeepRM's learning model also lacks support for DAG-structured jobs, and its training
procedure cannot handle realistic cluster workloads with continuous job
arrivals.}
%
%only works with a simple
%resource-time simulator, and does not support real jobs
%structured as DAGs.
%
%\name supports multi-dimensional resource packing, although it currently relies
%on fixed-size executor sizes.
%
\update{In other applications,} Mirhoseini et al.'s work on learning
device placement in TensorFlow (TF) computations~\cite{tf-device-placement} also
uses RL, but relies on recurrent neural networks to scan through
all nodes for state embedding, rather than a graph neural network.
Their approach use recurrent neural networks to scan through
all nodes for state embedding instead of using a scalable graph neural network.
The objective there is to schedule a single TF
job well, and the model cannot generalize to unseen job
combinations~\cite{hierarchical-tf-device-placement}.

\update{
Prior work in machine learning and algorithm design has combined RL and graph neural networks to optimize complex combinatorial
problems, such as vertex set cover and the traveling salesman problem~\cite{graphcombopt, gcn_comb_search}.
The design of \name's scalable state representation is inspired by this line of work, but
we found that off-the-shelf graph neural networks perform poorly for our problem.
To train strong scheduling agents, we had to change the
graph neural network architecture to enable \name to compute, amongst other metrics, the critical path of a DAG (\S\ref{s:graph}).
}

\update{For resource management systems more broadly,}
Paragon~\cite{paragon} and Quasar~\cite{quasar} use collaborative filtering to
match workloads to different machine types and avoid interference; their goal
is complementary to \name's.
Tetrisched~\cite{tetrisched}, like \name, plans ahead in time, but uses a
constraint solver to optimize job placement and requires the user to supply
explicit constraints with their jobs.
Firmament~\cite{firmament-osdi} also uses a constraint solver and achieves
high-quality placements, but requires an administrator to configure an
intricate scheduling policy.
Graphene~\cite{graphene} uses heuristics to schedule job DAGs, but cannot set
appropriate parallelism levels.
Some systems ``auto-scale'' parallelism levels to meet job
deadlines~\cite{jockey} or opportunistically accelerate jobs using spare
resources~\cite[\S5]{omega}.
\update{
Carbyne~\cite{carbyne} allows jobs to ``altruistically’' give up some of their short-term fair share of cluster resources in order to improve JCT across jobs while guarantee long-term fairness.
\name learns policies similar to Carbyne’s, balancing resource shares and packing for low average JCT, but the current design of \name does not have fairness an objective.
}
%Carbyne~\cite{carbyne} allows jobs to ``altruistically'' give up some of their
%fair share of cluster resources in order to improve JCT across jobs.
%%
%\name learns policies similar to Carbyne's, balancing resource shares and packing
%for low average JCT, but \name does not guarantee long-term fairness.
%

%
General-purpose cluster managers like Borg~\cite{borg}, Mesos~\cite{mesos},
or YARN~\cite{yarn} support many different applications, making
workload-specific scheduling policies are difficult to apply at this level.
However, \name could run as a framework atop Mesos or Omega~\cite{omega}.

\section{Conclusion}
\label{s:concl}
\name demonstrates that automatically learning complex cluster scheduling
policies using reinforcement learning is feasible, and that the learned policies
are flexible and efficient.
\name's learning innovations, such as its graph embedding technique and the training
framework for streaming, may be applicable to other systems processing DAGs
(\eg query optimizers).
We will open source \name, our models, and our experimental infrastructure at
\url{https://web.mit.edu/decima}.
%\name, our models, and our experimental infrastructure are open-source and available
%at:
%\begin{center}
%  \vspace*{-1mm}
%  {\bf\url{https://web.mit.edu/decima}}.
%  \vspace*{-1mm}
%\end{center}
%
%
%\noindent{\bf Ethics statement.}
This work does not raise any ethical issues.

\smallskip
\noindent{\bf Acknowledgments.} \update{We thank our shepherd, Aditya Akella,
and the SIGCOMM reviewers for their valuable feedback. We also thank
Akshay Narayan, Amy Ousterhout, Prateesh Goyal, Peter Iannucci, and Songtao He for
fruitful discussions throughout the development of this project. We appreciate Haiyang Ding and Yihui Feng from Alibaba Cloud Intelligence for sharing the production cluster dataset. This work was funded
in part by the NSF grants CNS-1751009, CNS-1617702, a Google Faculty Research Award,
an AWS Machine Learning Research Award, a Cisco Research Center Award, an Alfred P.
Sloan Research Fellowship, and sponsors of the MIT DSAIL lab.}

\bibliographystyle{ACM-Reference-Format}
\bibliography{paper}

\newpage
\appendix
\noindent
\textbf{\Large Appendices}

\smallskip
\noindent
\update{Appendices are supporting material that has not been peer reviewed.}

\section{An example of dependency-aware scheduling}
\label{s:graph-hard}

\begin{figure}[h]
\centering
\includegraphics[height=0.26\textwidth]{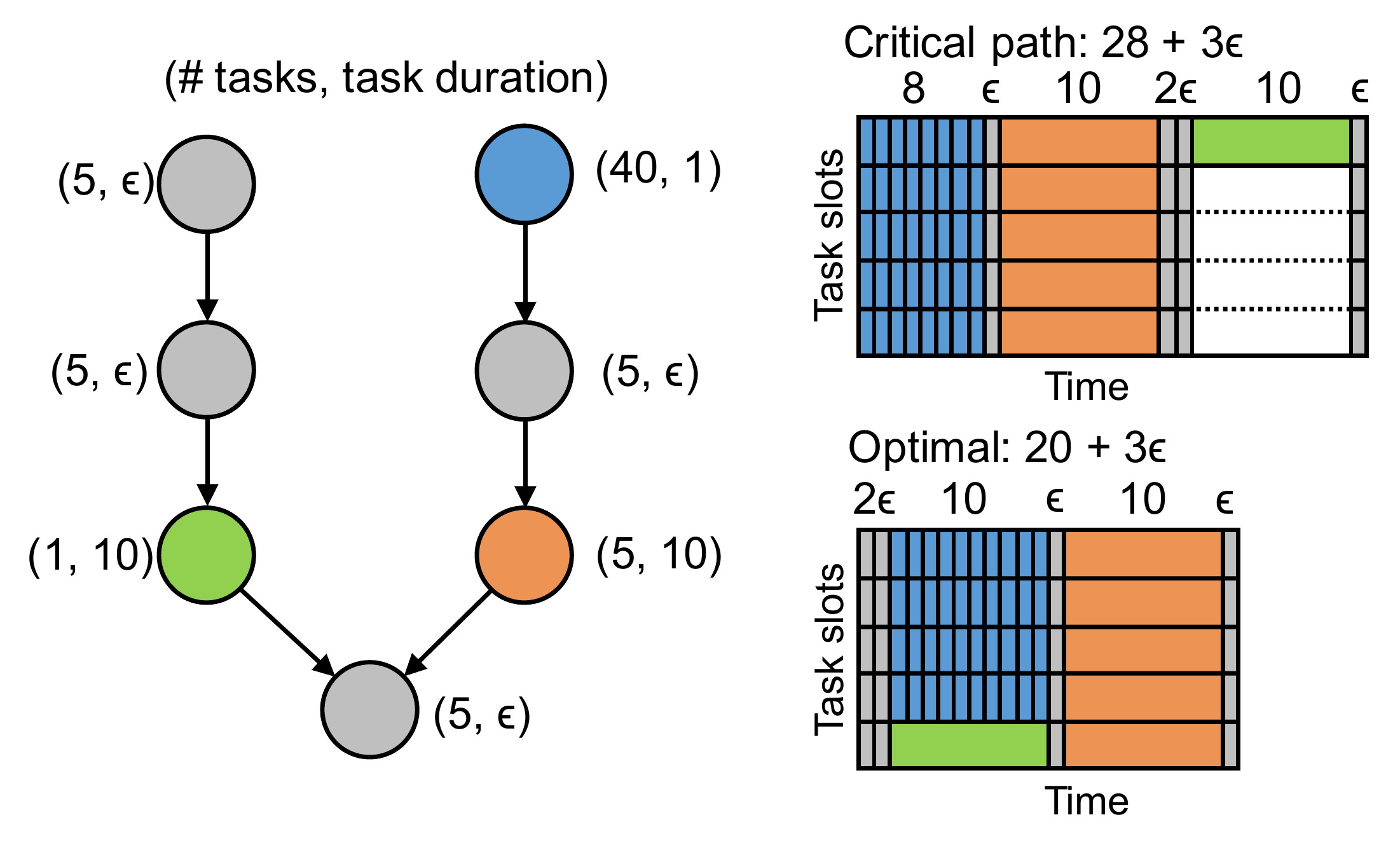}
\vspace{-0.4cm}
\caption{An optimal DAG-aware schedule plans ahead and parallelizes execution
  of the blue and green stages, so that orange and green stages complete at
  the same time and the bottom join stage can execute immediately.
  A straightforward critical path heuristic would instead focus on the right
  branch, and takes 29\% longer to execute the job.}
\label{f:dag-ex}
\end{figure}

Directed acyclic graphs (DAGs) of dependent operators or ``stages'' are
common in parallel data processing applications.
Figure~\ref{f:dag-ex} shows a common example: a DAG with two branches that
converge in a join stage.
A simple critical path heuristic would choose to work on the right branch,
which contains more aggregate work: 90 task-seconds vs.\ 10 task-seconds
in the left branch.
With this choice, once the orange stage finishes, however, the final join
stage cannot run, since its other parent stage (in green) is still
incomplete.
Completing the green stage next, followed by the join stage\,---\,as a
critical-path schedule would\,---\,results in an overall makespan of
$28 + 3\epsilon$.
The optimal schedule, by contrast, completes this DAG in $20 +
3\epsilon$ time, 29\% faster.
Intuitively, an ideal schedule allocates resources such that both branches
reach the final join stage at the same time, and execute it without blocking.
%

%%%%%%%%%%%%%%%%%%%%%%%%%%%%%%%%%%%%%%%%%%%%%%%%%%%%%%%%%%%%%%%%%%%%%%%%%%%%%
\section{Background on Reinforcement Learning}
\label{s:background}
We briefly review reinforcement learning (RL) techniques that we use in this
paper; for a detailed survey and rigorous derivations, see \eg Sutton and
Barto's book~\cite{rlbook}.

% RL algorithms enable an {\em agent} to learn to make better decisions for a task through experience obtained interacting with an {\em environment}.
\begin{figure}[t]
\centering
\includegraphics[height=3.5cm]{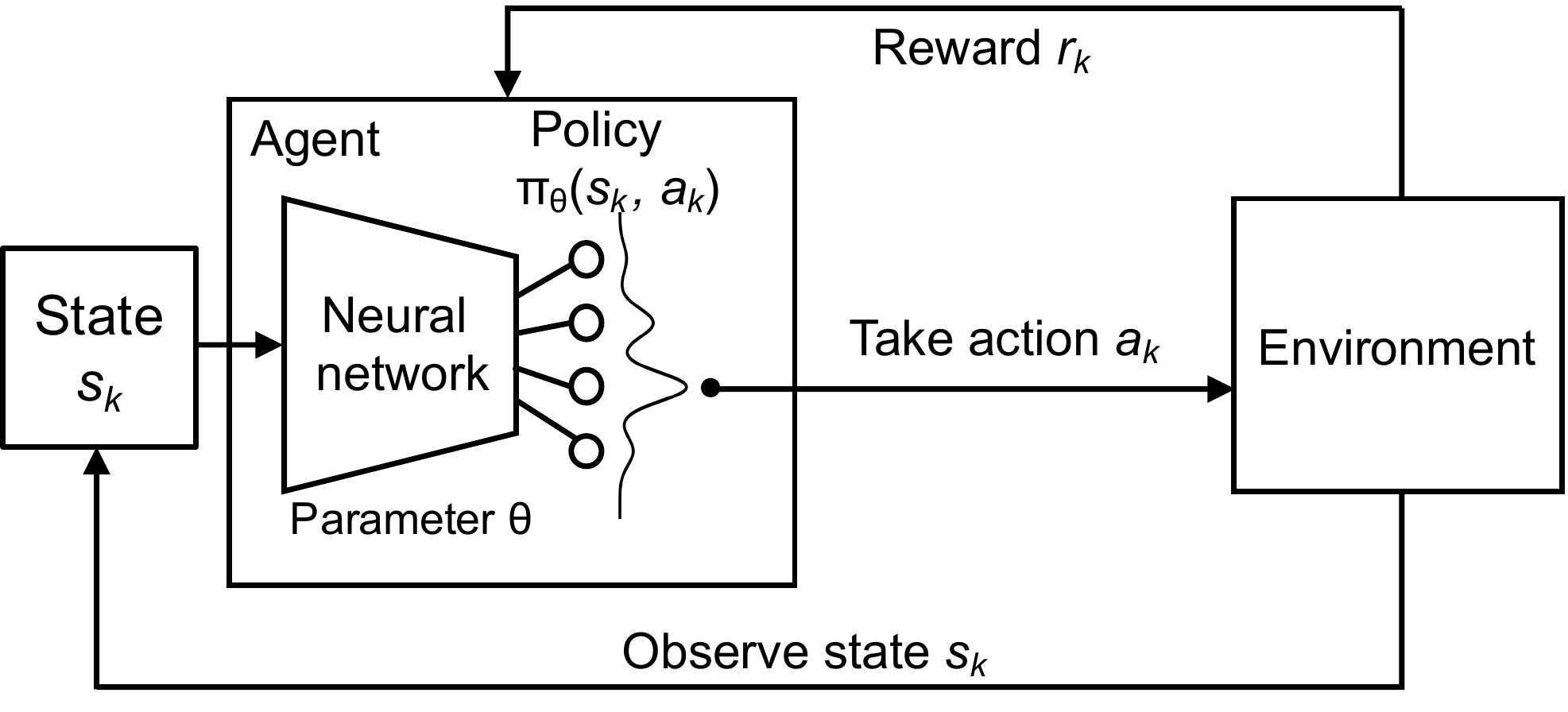}
\vspace{-0.2cm}
\caption{A reinforcement learning setting with neural networks~\cite{rlbook, deeprm}.
The policy is parameterized using a neural network and is trained iteratively
  via interactions with the environment that observe its state and take actions.}
\label{f:background_rl}
\end{figure}

\para{Reinforcement learning.}  Consider the general setting in Figure~\ref{f:background_rl}, where an RL {\em agent} interacts with an {\em environment}. At each step $k$, the agent observes
some state $s_k$, and takes an action $a_k$. Following the action, the state of the environment
transitions to $s_{k+1}$ and the agent receives a reward $r_k$ as feedback. The state transitions
and rewards are stochastic and assumed to be a Markov process: the state transition to $s_{k+1}$ and the reward $r_k$ depend only on the state $s_k$ and the action $a_k$ at step~$k$ (\ie they are conditionally independent of the past).

In the general RL setting, the agent only controls its actions: it has no a priori knowledge of the state transition probabilities
or the reward function. However, by interacting with the environment, the agent can
learn these quantities during training.

For training, RL proceeds in \emph{episodes}. Each episode consists of a sequence of
(state, action, reward) observations\,---\,\ie $(s_k, a_k, r_k)$ at each step
$k \in [0, 1, \ldots, T]$, where $T$ is the episode length .
For ease of understanding, we first describe an RL formulation that maximizes the total reward: $\mathbb{E} \left[ \sum_{k=0}^T r_k \right]$. However, in our scheduling problem, the average reward formulation (\S\ref{s:training}) is more suitable. We later describe how to modify the reward signal to convert the objective to the average reward setting.
%in which can then reuse all the RL training techniques without further modifications.
%
%The goal of learning is to maximize the expected cumulative discounted
%reward: $\mathbb{E} \left[ \sum_{t=0}^\infty \gamma^t r_t \right]$, where $\gamma \in (0,1]$ is a
%factor discounting future rewards.

%In the cluster scheduling problem this paper works on~(\S\ref{s:design-overview}), the environment
%corresponds to the cluster that orchestrates the executors to host data-processing jobs; the state
%captures the status of the DAGs in the scheduler's queue and the status of all executors
%(\S\ref{s:graph}); and the actions assign executors to work on different DAG stages across time
%(\S\ref{s:action}).

% RL schemes aim at improving the policy by interacting with the environment and properly processing the obtained rewards.

\para{Policy.}
The agent picks actions based on a {\em policy} $\pi(s_k, a_k)$, defined as a probability of taking action $a_k$ at state $s_k$.
%$\pi: \pi(s, a) \rightarrow [0,1]$; $\pi(s,a)$ is the probability that action $a$ is taken in state $s$.
In most practical problems, the number of possible $\{$state, action$\}$ pairs is too
large to store the policy in a lookup table.
It is therefore common to use
{\em function approximators}~\cite{ndp,menache2005basis},
with a manageable number of adjustable parameters, $\theta$, to represent the policy as
$\pi_\theta (s_k, a_k)$.
%The justification for approximating the policy is that the agent should take similar actions for ``close-by" states.
%
Many forms of function approximators can be used to represent the policy.
Popular choices include linear combinations of features of the state/action space
(i.e., $\pi_\theta (s_k,a_k) = \theta^T \phi(s_k,a_k)$), and, recently, neural
networks~\cite{nnbook} for solve large-scale RL tasks~\cite{atari, alphagozero}.
An advantage of neural networks is that they do not need hand-crafted features, and that
they are end-to-end differentiable for training.
%
%In this work, we use a novel \emph{graph embedding} technique to represent the policy as a set of
%scalable graph neural network. \S\ref{s:graph} explains the design of such policy network.

\para{Policy gradient methods.}
We focus on a class of RL algorithms that perform training by using {\em gradient-descent} on the policy parameters. Recall that the objective is to maximize the expected total reward; the gradient of this objective is given by:
\begin{equation}
\nabla_\theta \mathbb{E}_{\pi_\theta} \left[ \sum_{k=0}^T r_k \right] = \mathbb{E}_{\pi_\theta} \left[\sum_{k=0}^T \nabla_\theta \log \pi_\theta (s_k,a_k) Q^{\pi_\theta} (s_k, a_k) \right],
\label{eqn:policy_gr}
\end{equation}
where $Q^{\pi_\theta}(s_k, a_k)$ is the expected total discounted reward from (deterministically) choosing action $a_k$ in state $s_k$, and subsequently following policy $\pi_\theta$~\cite[\S13.2]{rlbook}.
The key idea in policy gradient methods is to estimate the gradient using the trajectories
of execution with the current policy. Following the {\em Monte Carlo Method}~\cite{montecarlo},
the agent samples multiple trajectories and uses the empirical total discounted reward,
$v_k$, as an unbiased estimate of $Q^{\pi_\theta}(s_k, a_k)$.
It then updates the policy parameters via gradient descent:
\begin{equation}
\theta \leftarrow \theta + \alpha \sum_{k=0}^T \nabla_\theta \log \pi_\theta (s_k, a_k) v_k,
\label{eqn:reinforce}
\end{equation}
where $\alpha$ is the learning rate. This equation results in the REINFORCE algorithm~\cite{reinforce}. The intuition of REINFORCE is that the direction $\nabla_\theta \log \pi_\theta (s_k, a_k)$ indicates how to change the policy parameters in order to increase $\pi_\theta (s_k, a_k)$ (the probability of action $a_k$ at state $s_k$). Equation~\ref{eqn:reinforce} takes a step in this direction; the size of the step depends on the magnitude of the return $v_k$. The net effect is to reinforce actions that empirically lead to better returns.
Appendix~\ref{s:pg} describes how we implement this training method in practice.

\update{
Policy gradient methods are better suited to our scheduling context than the alternative value-based methods for two reasons.
First, policy-based methods are easier to design if it is unclear whether the neural network architecture used has adequate expressive power.
The reason is that value-based methods aim to find a fixed-point of the Bellman equations~\cite{bellman1966}.
If the underlying neural network cannot express the optimal value function, then a value-based method can have difficulty converging because the algorithm is trying to converge to a fixed point that the neural network cannot express.
With policy-based methods, this issue does not arise, because regardless of the policy network's expressive power, the policy gradient algorithm will optimize for the reward objective over the space of policies that the neural network \emph{can} express.
Second, policy gradient methods allow us to use input-dependent baselines~\cite{variance-reduction} to reduce training variance (challenge \#2 in \S\ref{s:training}).
It is currently unknown whether, and how, this technique can be applied to value-based methods.
}

\para{Average reward formulation.}
For our scheduling problem, an average reward formulation, which maximizes $\lim_{T\to\infty} \mathbb{E} \left[1/T \sum_{k=0}^T r_k \right]$, is a better objective than the total reward we discussed so far.
To convert the objective from the sum of rewards to the average reward, we replace the reward $r_k$ with a {\em differential reward}.
Operationally, at every step $k$, the environment modifies the reward to the agent as
$r_k \leftarrow r_k - \hat{r}$, where
$\hat{r}$ is a moving average of the rewards across a large number of previous steps (across many training episodes).
With this modification, we can reuse the same policy gradient method as in
Equation~\eqref{eqn:policy_gr} and~\eqref{eqn:reinforce} to find the optimal policy.
%Intuitively, such a reward signal `normalizes' the sum completion times of jobs in
%an RL episode by the length of the episode to counter any bias arising due to varying episode lengths.
%The net result is an objective that closely correlates to the average number of concurrent jobs in the system.
%
Sutton and Barto~\cite[\S10.3, \S13.6]{rlbook} describe the mathematical details on how this approach optimizes the average reward objective.

%%%%%%%%%%%%%%%%%%%%%%%%%%%%%%%%%%%%%%%%%%%%%%%%%%%%%%%%%%%%%%%%%%%%%%%%%%%%%
\section{Training implementation details}
\label{s:pg}
Algorithm~\ref{alg:pg} presents the pseudocode for \name's training
procedure as described in~\S\ref{s:training}.
In particular, line 3 samples the episode length $\tau$ from an exponential distribution, with a small initial mean $\tau_\text{mean}$. This step terminates the initial episodes early to avoid wasting training time~(see challenge \#1 in \S\ref{s:training}).
Then, we sample a job sequence (line 4) and use it to collect $N$ episodes of experience (line 5).
Importantly, the baseline $b_k$ in line 8 is computed with the \emph{same} job sequence to reduce the variance caused by the randomness in the job arrival process~(see challenge \#2 in \S\ref{s:training}).
Line 10 is the policy gradient REINFORCE algorithm described in Eq.~\eqref{eq:policygradient}.
Line 13 increases the average episode length (\ie the curriculum learning
procedure for challenge \#1 in \S\ref{s:training}).
Finally, we update \name's policy parameter $\theta$ on line 14.

\begin{algorithm}[t]
\caption{Policy gradient method used to train \name.}
\label{alg:pg}
\begin{algorithmic}[1]
\FOR{each iteration}
\STATE $\Delta \theta \leftarrow 0$
\STATE Sample episode length $\tau \sim \text{exponential}(\tau_\mathrm{mean})$
\STATE Sample a job arrival sequence
\STATE Run episodes $i=1,\hdots,N$:\\
\quad$\{s^i_1, a^i_1, r^i_1, \hdots, s^i_{\tau}, a^i_{\tau},  r^i_{\tau} \} \sim \pi_\theta$
\STATE Compute total reward: $R^i_k = \sum_{k'=k}^{\tau} r^i_{k'}$
\FOR{$k$ = 1 to $\tau$}
\STATE compute baseline: $b_k = \frac{1}{N}\sum_{i=1}^N R^i_k$
\FOR{$i$ = 1 to $N$}
\STATE $\Delta \theta \leftarrow \Delta \theta + \nabla_\theta \log \pi_\theta (s^i_k, a^i_k) (R^i_k -b_k)$
\ENDFOR
\ENDFOR
\STATE $\tau_\mathrm{mean} \leftarrow \tau_\mathrm{mean} + \epsilon$
\STATE $\theta \leftarrow \theta + \alpha\Delta\theta$
\ENDFOR
\end{algorithmic}
\end{algorithm}

Our neural network architecture is described in \S\ref{s:impl-spark}, and
we set the hyperparameters in \name's training as follows.
The number of incoming jobs is capped at 2000, and the episode
termination probability decays linearly from
$5\times10^{-7}$ to $5\times10^{-8}$ throughout training.
The learning rate $\alpha$ is $1\times 10^{-3}$ and we use Adam optimizer~\cite{adamopt} for gradient descent.
For continuous job arrivals, the moving window for
estimating $\hat{r}$ spans $10^5$ time steps (see the average reward formulation in Appendix~\ref{s:background}).
Finally, we train \name for at least 50,000 iterations for all experiments.
We implemented \name's training framework using Tensor\-Flow~\cite{tensorflow}, and we use 16 workers to compute
episodes with the same job sequence in parallel during training.
Each training iteration, including interaction with the simulator, model inference and model update from
all training workers, takes roughly 1.5 seconds on a machine with Intel Xeon E5-2640 CPU and Nvidia Tesla P100 GPU.
All experiments in \S\ref{s:eval} are performed on test job sequences unseen during training (\eg unseen TPC-H job combinations,
unseen part of the Alibaba production trace, etc.).
%

%%%%%%%%%%%%%%%%%%%%%%%%%%%%%%%%%%%%%%%%%%%%%%%%%%%%%%%%%%%%%%%%%%%%%%%%%%%%%

\section{Simulator fidelity}
\label{s:sim-fidelity}

\begin{figure}
  \centering
\begin{subfigure}[t]{0.5\textwidth}
  \centering
  \includegraphics[width=\textwidth]{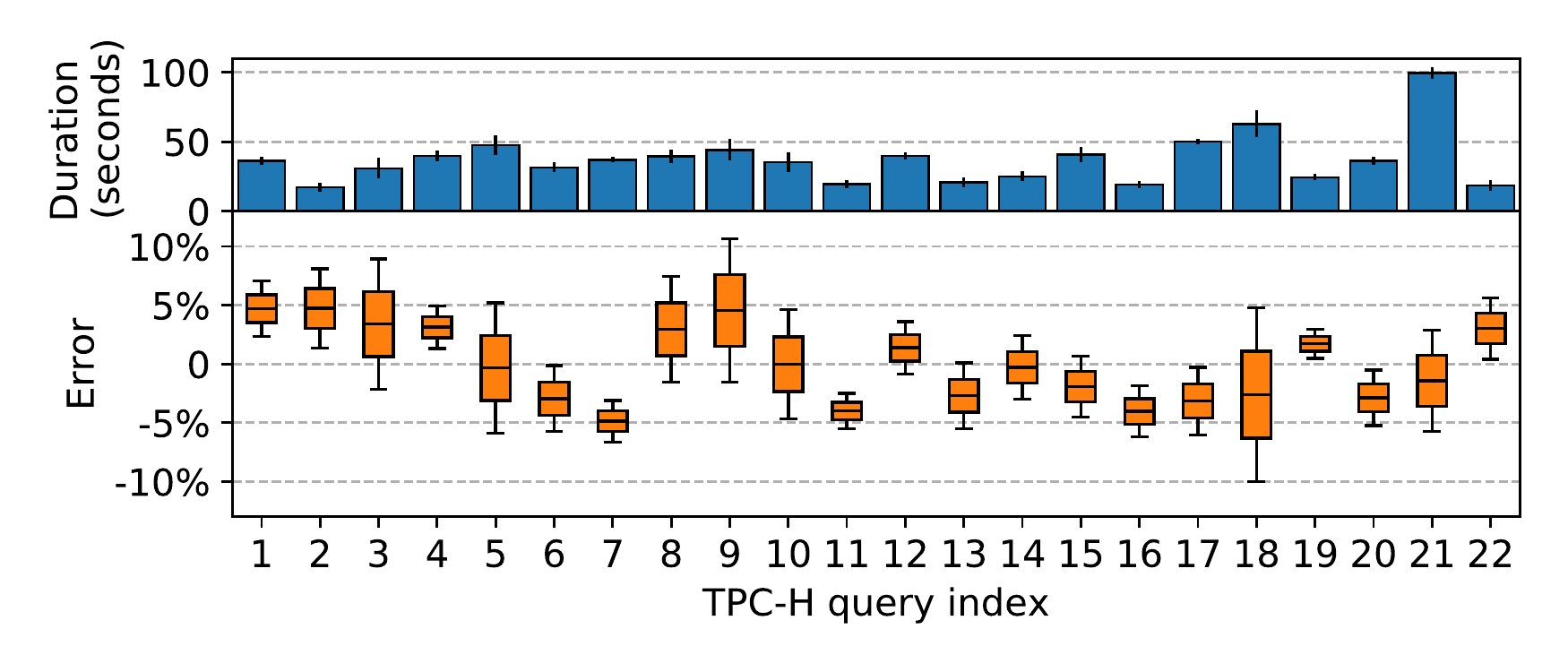}
  \vspace{-0.7cm}
  \caption{Single job running in isolation.}
  \label{f:simulator-fidelity-single}
\end{subfigure}
\begin{subfigure}[t]{0.5\textwidth}
  \centering
  \includegraphics[width=\textwidth]{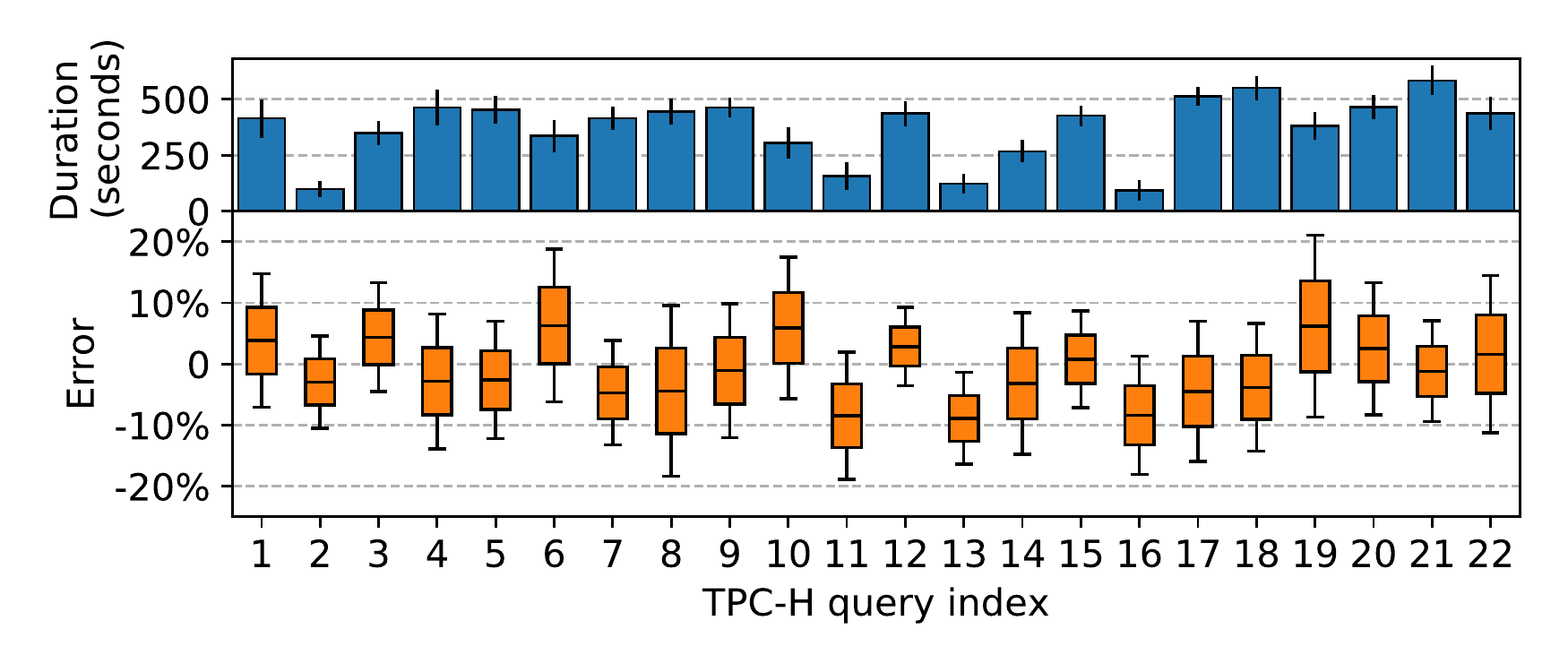}
  \vspace{-0.7cm}
  \caption{Mixture of jobs on a shared cluster.}
  \label{f:simulator-fidelity-mixed}
\end{subfigure}
\vspace{-0.3cm}
\caption{Testing the fidelity of our Spark simulator with \name as a scheduling
  agent.
  Blue bars in the upper part show the absolute real Spark job duration (error
  bars: standard deviation across ten experiments); the orange bars in the
  lower figures show the distribution of simulation error for a $95\%$ confidence
  interval.
  The mean discrepancy between simulated and actual job duration is at most
  $\pm 5\%$ for isolated, single jobs, and the mean error for a mix of all 22
  queries running on the cluster is at most $\pm 9\%$.}
  \label{f:sim-fidelity}
\end{figure}

Our training infrastructure relies on a faithful simulator of Spark job
execution in a cluster.
To validate the simulator's fidelity, we measured how simulated and real
Spark differ in terms of job completion time for ten runs of TPC-H job
sets~(\S\ref{s:eval-real-cluster}), both when jobs run alone and when they
share a cluster with other jobs.
Figure~\ref{f:sim-fidelity} shows the results: the simulator closely matches
the actual run time of each job, even when we run multiple jobs together in
the cluster.
In particular, the mean error of our simulation is within $5\%$ of real
runtime when jobs run in isolation, and within $9\%$ when sharing a cluster
(95\textsuperscript{th} percentile: $\le 10\%$ in isolation, $\le 20\%$
when sharing).
We found that capturing all first-order effects of the Spark environment is
crucial to achieving this accuracy~(\S\ref{s:simulator}).
For example, without modeling the delay to move an executor between jobs, the
simulated runtime consistently underapproximates reality.
Training in such an environment would result in a policy that moves executors
more eagerly than is actually sensible~(\S\ref{s:eval-ubenches}).
Likewise, omitting the effects of initial and subsequent ``waves'' of tasks,
or the slowdown overheads imposed with high degrees of paralllism,
significantly increases the variance in simulated runtime and makes it more
difficult for \name to learn a good policy.
%
%Only a simulator that considers all these factors can we achieve the small
%discrepancy between real and simulation as shown in Figure~\ref{f:sim-fidelity}.
%

%%%%%%%%%%%%%%%%%%%%%%%%%%%%%%%%%%%%%%%%%%%%%%%%%%%%%%%%%%%%%%%%%%%%%%%%%%%%%
\section{Expressiveness of \name's state representation}
\label{s:su}
\begin{figure}
  \centering
  \includegraphics[width=0.85\columnwidth]{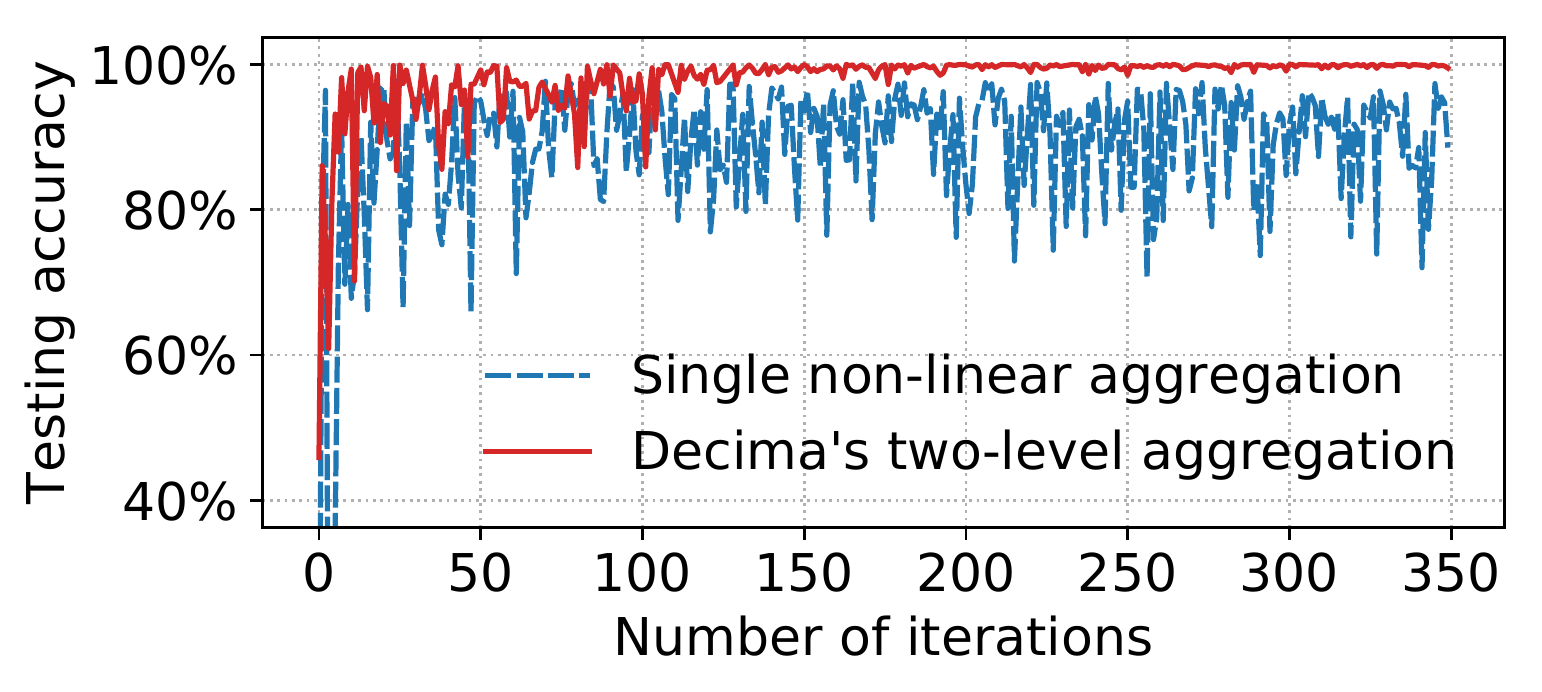}
  \vspace{-0.25cm}
  \caption{\update{Trained using supervised learning, \name's two-level non-linear transformation is able to express the max operation necessary for computing the critical path (\S\ref{s:graph}), and consequently achieves near-perfect accuracy on unseen DAGs compared to the standard graph embedding scheme.}}
  \label{f:su}
\end{figure}
\update{
\name's can only learn strong scheduling policies if its state representation, embedding
scheme, and neural network architecture can express them (\S\ref{s:eval}).
In Equation~\eqref{eq:msg}, combining two non-linear transforms $f(\cdot)$ and $g(\cdot)$ enables
\name to express a wide variety of aggregation functions.
For example, if $f\sim \log(\cdot / n)$, $g\sim \exp(n \times \cdot)$, and $n\to\infty$, the
aggregation computes the maximum of the child node embeddings.
By contrast, a standard aggregation operation of the form
$\mathbf{e_v} = \sum_{u\in \xi(v)} f(\mathbf{e_u})$ without a second non-linear transformation
$g(\cdot)$ is insufficient to express the max operation.
Consequently, such an architecture cannot learn the aggregation (max) required to find the
critical path of a graph.
During development, we relied on a simple sanity check to test the expressiveness of a
graph embedding scheme.
We used supervised learning to train the graph neural network to output the critical path
value of each node in a large number of random graphs, and then checked how accurately the
graph neural network identified the node with the maximum critical path value.
Figure~\ref{f:su} shows the testing accuracy that \name's node embedding with two
aggregation levels achieves on unseen graphs, and compares it to the accuracy achieved by
a simple, single-level embedding with only one non-linear transformation.
\name's node embedding manages to learn the max operation and therefore accurately
identifies the critical path after about 150 iterations, while the standard embedding
is incapable of expressing the critical path and consequently never reaches a stable
high accuracy.}

\section{Multi-resource scheduling heuristic comparison details}
\label{s:graphene}
When evaluating \name's performance in a multi-resource
setting~(\S\ref{s:multi-dim}), we compared with several heuristics.
First, we considered the optimally tuned weighted fair heuristic
from~\S\ref{s:eval-real-cluster}.
This heuristic grants each job an executor share based on the total work in the
job.
Then the heuristic chooses a stage the same way as in the single resource
setting.
Among the available executor types, the heuristic first exhausts the
best-fitting category before choosing any others.
The scheduler ensures that the aggregate allocated resources
(across different executor types) do not exceed the job's weighted fair share.

Second, we compared to the resource-packing algorithm from Tetris~\cite{tetris}.
To maximize resource utilization, we select the DAG node that yields the largest
dot product of the requested resource vector and the available resource vector
for each executor type.
Then, we greedily grant as much parallelism as the tasks in this node need.
The prior two heuristics lack each other's key scheduling ingredients (fairness
and packing), and neither understands the DAG structure.
Finally, we compared to Graphene~\cite{graphene}, whose hybrid heuristic combines
these factors.
However, our multi-resource scheduling environment with discrete executor
classes differs from the original Graphene setting, which assumes continuous,
infinitely divisible resources.
We adapted the Graphene algorithm for discrete executors, but kept its essence:
specifically, we estimate and group the ``troublesome'' nodes the same
way~\cite[\S4.1]{graphene}.
To ensure that troublesome nodes are scheduled at the same time, we dynamically
suppress the priority on all troublesome nodes of a DAG until \emph{all}
of these nodes are available in the frontier.
%
%\ms{I see why we do this, but it seems like this might yield quite a different
%schedule than Graphene's time-based planning. In particular, since this happens
%\emph{per-job}, scheduling decisions in other jobs have an effect on when the
%troublesome tasks are runnable. Do we have reason to believe the schedules end
%up being comparable? Re: Graphene performs the planning for each jobs independently
%first. The planning gives a priority for each node and the run time execution just
%follows this order, which is similar to what we are doing.}
%
We also include parallelism control by sharing the executors according to the
optimally tuned weighted fair partition heuristic; and we pack resources by
prioritizing the executor type that best fits the resource request.
Finally, we perform a grid search on all the hyperparameters (\eg the
threshold for picking troublesome nodes) to tune the heuristic for the best
scheduling performance in each of the experiments (\S\ref{s:multi-dim}).
%

%%%%%%%%%%%%%%%%%%%%%%%%%%%%%%%%%%%%%%%%%%%%%%%%%%%%%%%%%%%%%%%%%%%%%%%%%%%%%
\section{Further analysis of multi-resource scheduling}
\label{s:multi-resource-more-analysis}
\begin{figure}[t]
\centering
\begin{subfigure}[t]{0.22\textwidth}
  \centering
  \includegraphics[width=\textwidth]{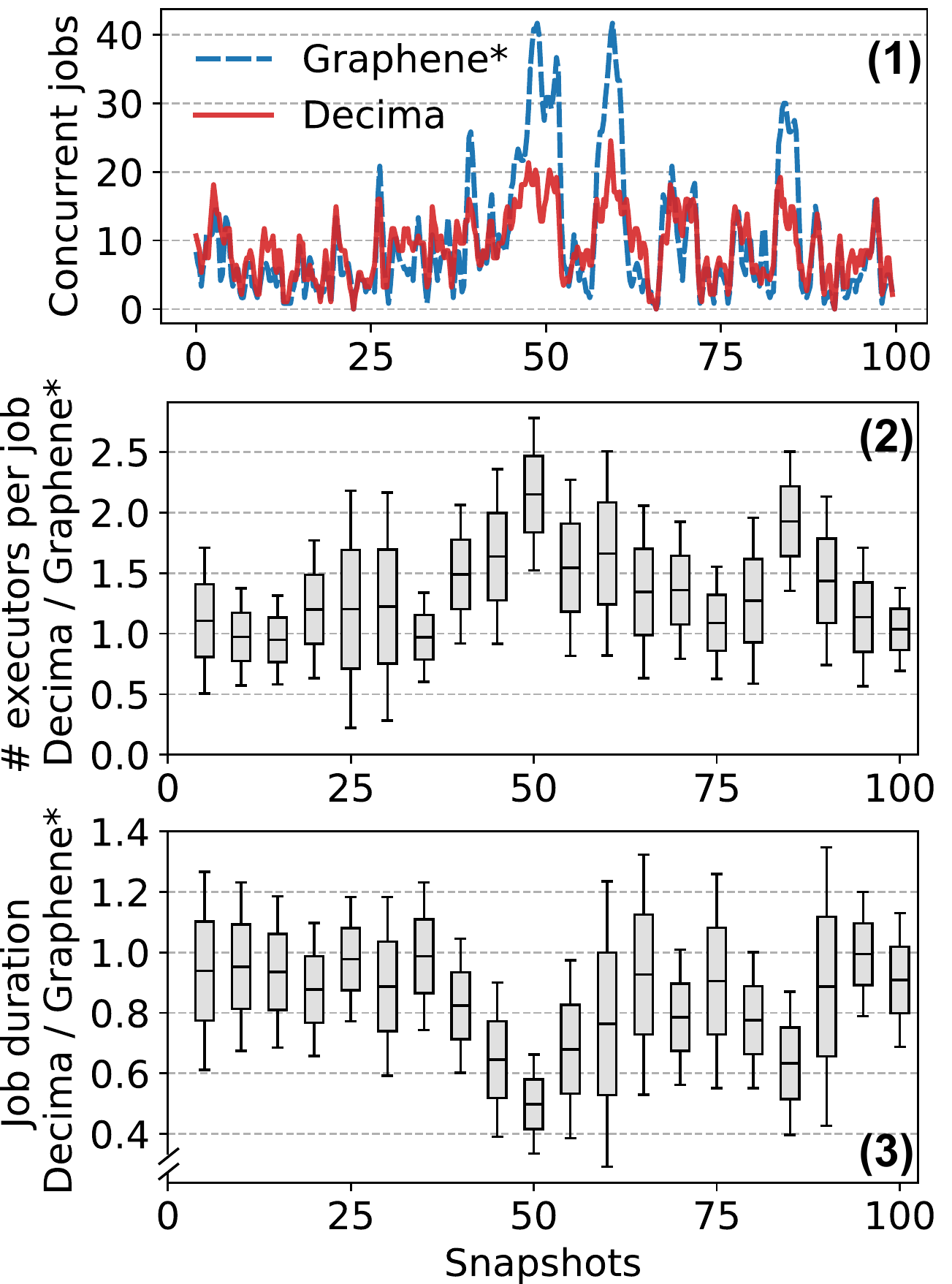}
  \vspace{-0.6cm}
  \caption{TPC-H workload.}
  \label{f:eval-tpch-multi-time-series}
\end{subfigure}
\hspace{0.01\textwidth}
\begin{subfigure}[t]{0.22\textwidth}
  \centering
  \includegraphics[width=\textwidth]{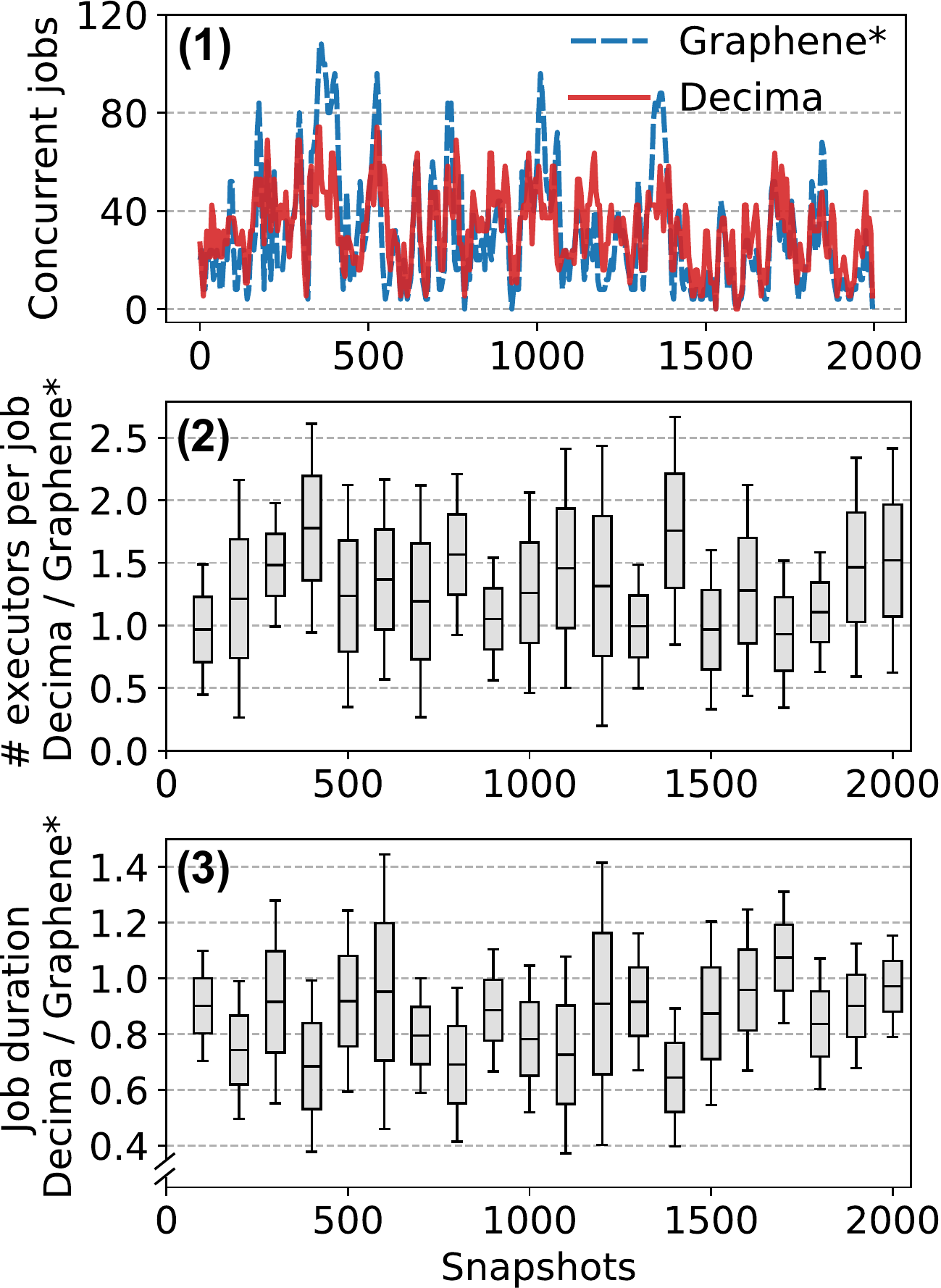}
  \vspace{-0.6cm}
  \caption{Industrial trace replay.}
  \label{f:eval-alibaba-multi-time-series}
\end{subfigure}
\vspace{-0.2cm}
\caption{Timeseries of different statistics in the extended Spark multi-resource environment. We compare \name and \Graphene, the best competing heuristic. During busy periods, \name finishes jobs faster and maintains a lower number of concurrent jobs by using more executors per job.}
\label{f:eval-multi-time-series}
\end{figure}

In \S\ref{s:multi-dim}, we found that \name achieves $32\%-43\%$ lower average
JCT than state-of-the-art heuristics when handling continuous job arrivals in
a multi-resource environment.
\name achieves this by carefully fragmenting cluster memory:
Figure~\ref{f:eval-tpch-multi-resource-exec-usage} illustrated that \name
selectively borrows large executors if they can help finishing short jobs
quickly and increase cluster throughput.
This effect is also evident when examining the timeseries of job duration and
executor usage over a single experiment.
Figure~\ref{f:eval-multi-time-series} shows that \name maintains a smaller
number of concurrent active jobs during periods of high load both for a
synthetic TPC-H workload and for the Alibaba trace.
During busy periods (\eg around snapshot 50 in
Figure~\ref{f:eval-tpch-multi-time-series}{\color{darkred}1}),
\name clears the backlog of jobs in the queue more quickly than the best
competing heuristic, \Graphene.
During these periods, \name assigns more executors to each job than
\Graphene (Figures~\ref{f:eval-tpch-multi-time-series}{\color{darkred}2}
and \ref{f:eval-alibaba-multi-time-series}{\color{darkred}2}), \eg by sometimes
borrowing large executors for jobs that need only smaller ones.
As a consequence, \name achieves lower JCT and higher cluster throughput when
the cluster load is high (Figures~\ref{f:eval-tpch-multi-time-series}{\color{darkred}3}
and \ref{f:eval-alibaba-multi-time-series}{\color{darkred}3}).
%
%the executor assignment on each job is
%larger for \name than that for \Graphene.
%
%\name manages to move more executors to a job by selectively borrows large executors.

\begin{figure}[t]
\centering
\begin{subfigure}[t]{0.2\textwidth}
  \centering
  \includegraphics[width=\textwidth]{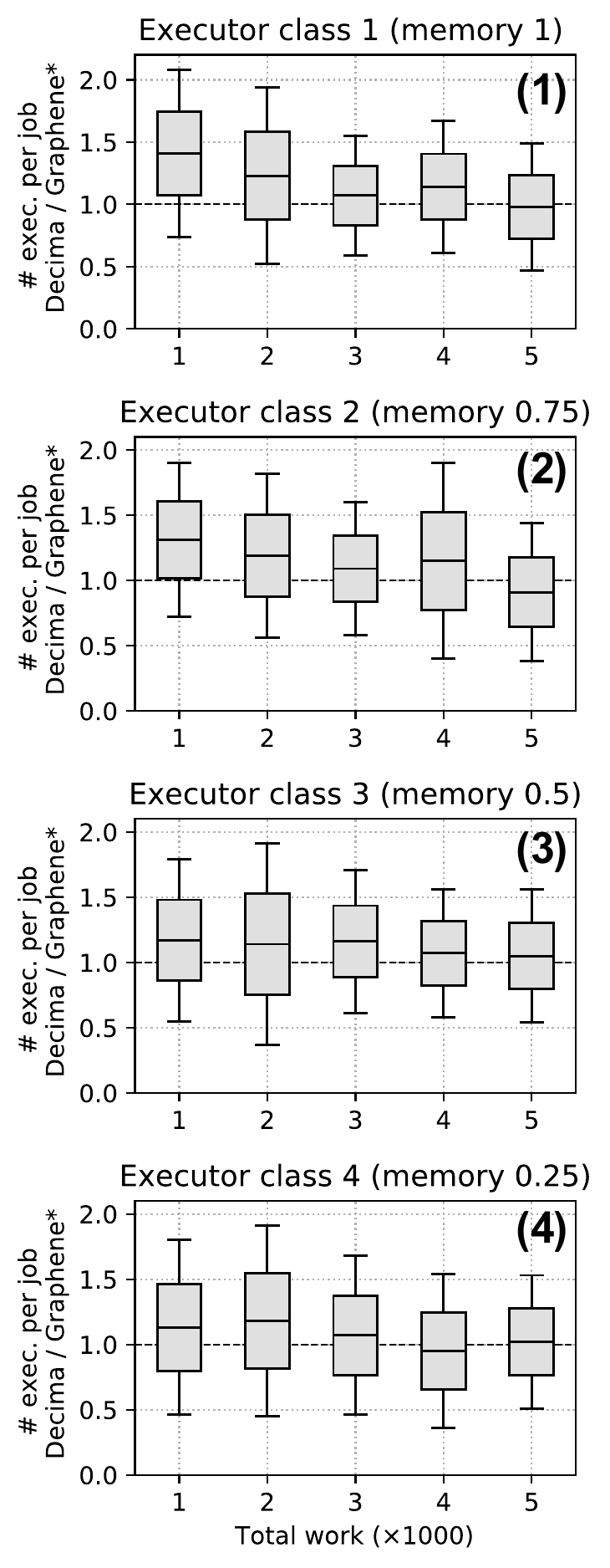}
  \vspace{-0.6cm}
  \caption{TPC-H workload.}
  \label{f:eval-tpch-multi-exec-assignment}
\end{subfigure}
\hspace{0.03\textwidth}
\begin{subfigure}[t]{0.2\textwidth}
  \centering
  \includegraphics[width=\textwidth]{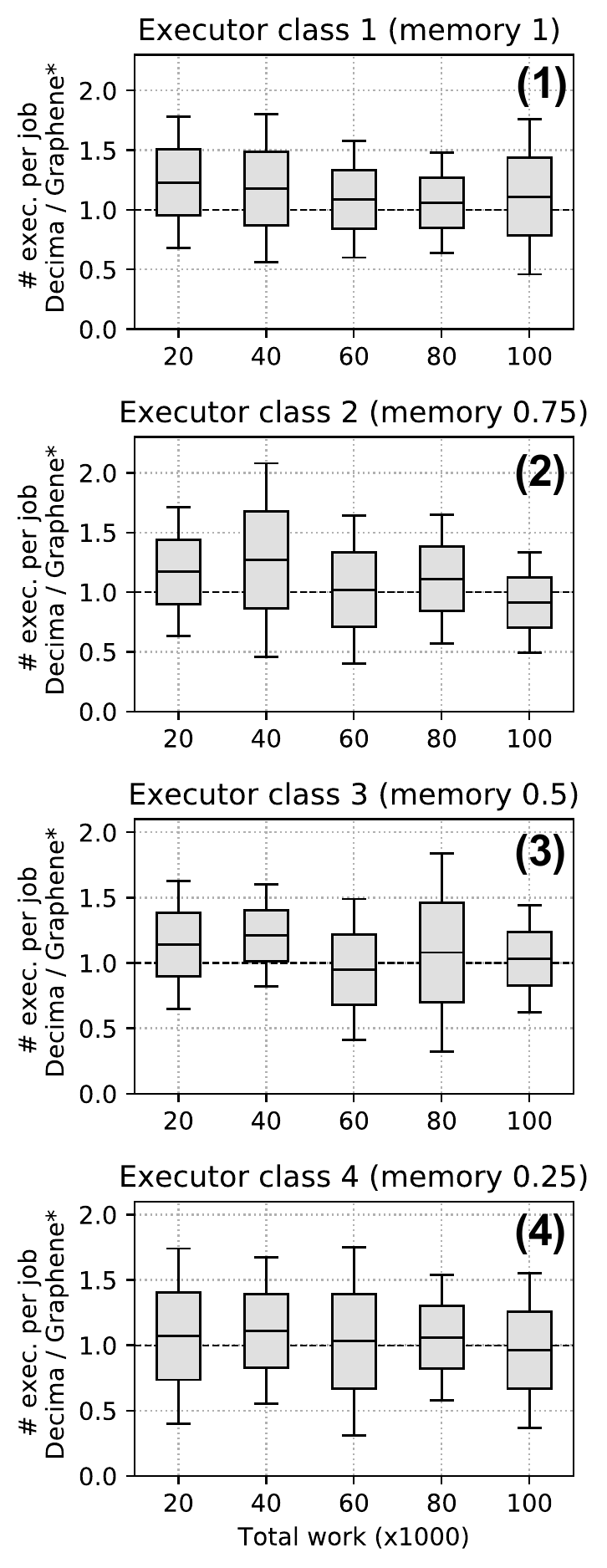}
  \vspace{-0.6cm}
  \caption{Industrial trace replay.}
  \label{f:eval-alibaba-multi-exec-assignment}
\end{subfigure}
\vspace{-0.2cm}
\caption{Profile of executor assignments on jobs with different sizes, \name normalized to
  \Graphene's assignment (>1: more executors in \name, <1: more in \Graphene).
  \name tends to assign more executors.}
\label{f:eval-multi-exec-assignment}
\end{figure}

Figure~\ref{f:eval-tpch-multi-exec-assignment} and \ref{f:eval-alibaba-multi-exec-assignment}
compare the executor assignment between \name
and \Graphene in detail (Figure~\ref{f:eval-tpch-multi-resource-exec-usage} is the first
column of this profile).
On the $x$-axis, we bin jobs according to the total amount of work they contain (in
task-seconds).
The $y$-axis of each graph is the number of executors \name uses normalized to the number of
executors used by \Graphene\,---\,\ie a value above one indicates that \name used more executors.
Overall, \name tends to assign more executors per job compared to \Graphene.
This helps \name complete jobs faster in order to then move on to others, instead of making
progress on many jobs concurrently, similar to the behavior we discussed
in~\S\ref{s:eval-real-cluster}.
Moreover, \name uses more large executors on small jobs.
This aggressive allocation of large executors\,---\,which wastes some memory\,---\,leads to faster
job completion during the busy periods (Figure~\ref{f:eval-tpch-multi-time-series}{\color{darkred}3} and
\ref{f:eval-alibaba-multi-time-series}{\color{darkred}3}), at the expense of leaving some
memory unused.
This trade-off between resource fragmentation and prioritizing small jobs can be tedious to
balance, but \name automatically learns a strong policy by interacting with the environment.
\name may enjoy an advantage here partly because \Graphene is restricted to discrete executor
classes.
In a cluster setting with arbitrary, continuous memory assignment to tasks, large executor
``slots'' could be subdivided into multiple smaller executors, assuming sufficient CPU
capacity exists.
This choice is difficult to express with a finite action space like \name's, and it is an
interesting direction for future work to investigate whether RL with continuous action
could be applied to cluster scheduling.

%%%%%%%%%%%%%%%%%%%%%%%%%%%%%%%%%%%%%%%%%%%%%%%%%%%%%%%%%%%%%%%%%%%%%%%%%%%%%

\section{Optimality of \name}
\label{s:optimality}
In~\S\ref{s:eval}, we show \name is able to rival or outperform existing
scheduling schemes in a wide range of complex cluster environments, including
a real Spark testbed, real-world cluster trace simulations and a multi-resource
packing environment.
However, the optimality of \name in those environments
remains unknown due to the intractability of computing exact optimal
scheduling solutions~\cite{acyclic-jobs-hard, graphene},
or tight lower bounds.\footnote{In our setting (\ie Spark's executor-based
scheduling), we found lower bounds based on total work or the critical path to
be too loose to provide meaningful information.}
To nevertheless get an idea of how close \name comes to an optimal scheduler, we
test \name in simplified settings where a brute-force search over different
schedules is possible.
We consider the Spark scheduling framework simulated in~\S\ref{s:simulator} with
an average JCT objective for a batch of jobs.
To simplify the environment, we turn off the ``wave'' effect, executor startup
delays and the artifact of task slowdowns at high degrees of parallelism.
As a result, the duration of a stage has a strict inverse relation to the number
of executors the stage runs on (\ie it scales linearly with parallel resources),
and the scheduler is free to move executors across jobs without any overhead.
The dominating challenges in this environment are to pack jobs tightly and to
favor short jobs as much as possible.
To find a good schedule for a batch of $n$ jobs, we exhaustively search all
$n!$ possible job orderings, and select the ordering with the lowest average JCT.
To make the exhaustive search feasible, we consider a batch of ten jobs.
For each job ordering, we select the unfinished job appearing earliest in the
order at each scheduling event (\S\ref{s:action}), and use the DAG's critical path
to choose the order in which to finish stages within each job.
%
% At each scheduling event~(\S\ref{s:action}), we \emph{virtually} try
% to schedule each job (we use critical path to choose the order of nodes
% within each job) individually with all available executors. We scan through
% all jobs and check the virtual completion time of each job. Then we greedily
% pick the job with minimum virtual JCT to schedule next.
%The pseudocode in algorithm~\ref{} depicts this heuristic.
%
By considering all possible job orderings, the algorithm is guaranteed to
consider, amongst other schedules, a strict shortest-job-first (SJF) schedule
that yields a small average JCT.
We believe this policy to be close to the optimal policy, as we have empirically
observed that job orderings dominate the average JCT in TPC-H
workloads~(\S\ref{s:eval-ubenches}).
However, the exhaustive search also explores variations of the SJF schedule, \eg
orders that prioritize jobs which can exploit parallelism to complete more quickly
than less-parallelizable jobs that contain smaller total work.

Next, we train an unmodified \name agent in this environment, similar to the setup
in~\S\ref{s:eval-real-cluster}.
We compare this agent's performance with our exhaustive search baseline, a
shortest-job-first critical-path heuristic, and the tuned weighted fair scheduler
(described in~\S\ref{s:eval-real-cluster}).

\begin{figure}
  \centering
  \includegraphics[width=0.85\columnwidth]{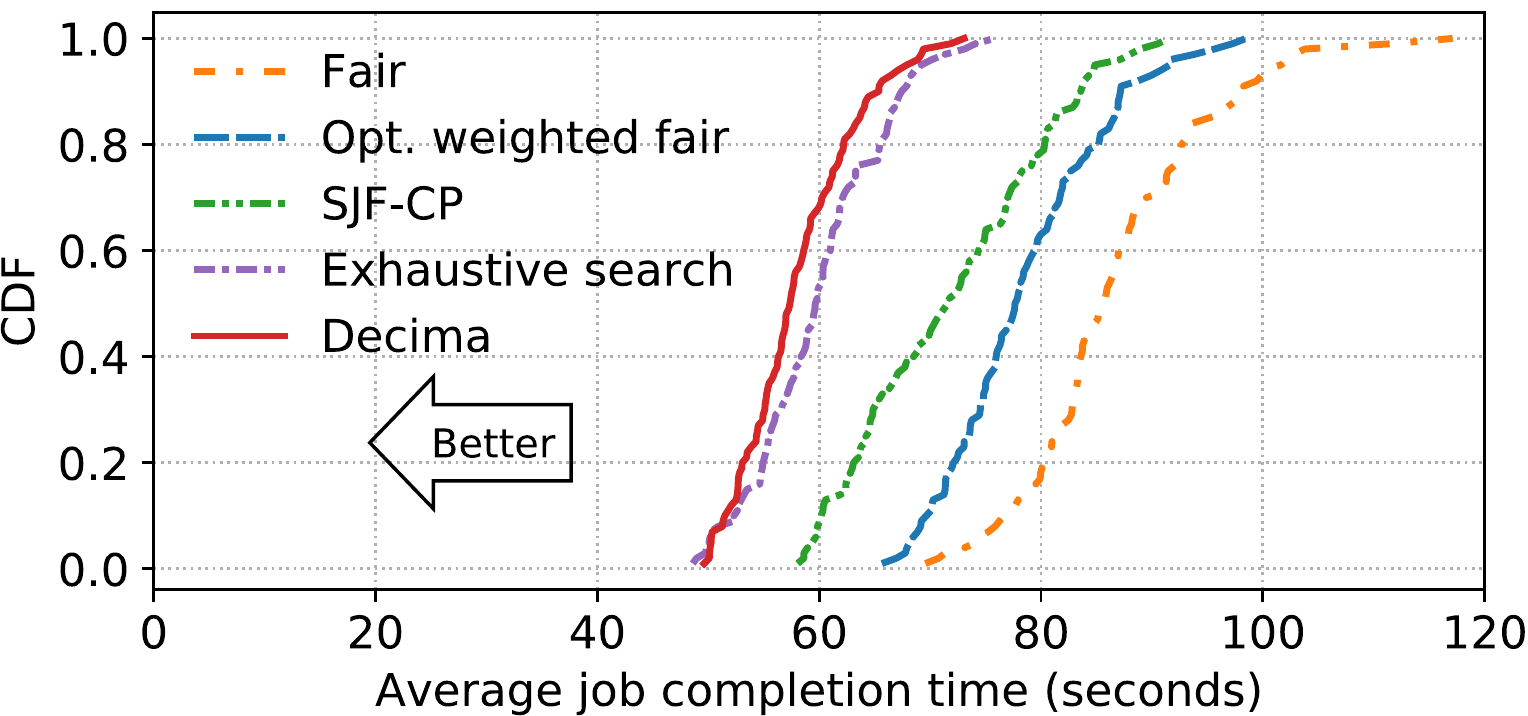}
  \caption{Comparing \name with near optimal heuristics in a simplified scheduling environment.}
  \label{f:near-opt-comparison}
\end{figure}

Figure~\ref{f:near-opt-comparison} shows the results.
We make three key observations.
First, unlike in the real Spark cluster~(Figure~\ref{f:eval-spark-tpch-cdf}),
the SJF-CP scheme outperforms the tuned weighted fair scheduler.
This meets our expectation because SJF-CP strictly favors small jobs to minimize
the average JCT, which in the absence of the complexities of a real-world cluster
is a good policy.
Second, the exhaustive search heuristic performs better than SJF-CP.
This is because SJF-CP strictly focuses on completing the job with the smallest
total work first, ignoring the DAG structure and the potential parallelism it
implies.
The exhaustive search, by contrast, finds job orderings that prioritize jobs which
can execute most quickly given the available executors on the cluster, their DAG
structure, and their total work.
While the search algorithm is not aware of these constraints, by trying out
different job orderings, it finds the schedule that both orders jobs correctly
\emph{and} exploits cluster resources to complete the jobs as quickly as
possible.
%
% non-uniformity of resource utilization of a job\,---\,the number of executors consumed at
% different stage of a job are different;
% and SJF-CP, which only has access to total work of a job and the critical path, does not utilize the information of the resource utilization.
% In contrast, the planning based heuristic dynamically considers the packing of jobs
% by using the information of the whole graph. Therefore, it virtually schedules the job
% with current remaining resources to find a better packing of resource while making
% sure short jobs finish fast.
%
Third, \name matches the average JCT of the exhaustive search or even outperforms it
slightly (by $9\%$ on average).
We found that \name is better at dynamically prioritizing jobs based on their current
structure at runtime (\eg how much work remains on each dependency path), while the
exhaustive search heuristic strictly follows the order determined in an offline
static search and only controls when jobs start.
This experiment shows that \name is able to automatically learn a scheduling
algorithm that performs as well as an offline-optimal job order.
%

%%%%%%%%%%%%%%%%%%%%%%%%%%%%%%%%%%%%%%%%%%%%%%%%%%%%%%%%%%%%%%%%%%%%%%%%%%%%%
\section{Generalizing \name to different environments}
\label{s:generalizability}
\begin{table}[t]
\small
\centering
\begin{tabular}{|l|c|}
\hline
  \bf \name training scenario & \bf average JCT (seconds) \\
\hline
  \name trained with test setting & $3,290\pm680$ \\
  \name trained with 15$\times$ fewer jobs  & $3,540\pm450$ \\
  \hline
  \name trained with test setting & $610\pm90$ \\
  \name trained with 10$\times$ fewer executors & $630\pm70$ \\
\hline
\end{tabular}
\vspace{0.25cm}
\caption{\name generalizes well to deployment scenarios in which the
         workload or cluster differ from the training setting.
         The test setting has 150 jobs and 10k executors.}

\label{t:eval-alibaba-generalization}
\end{table}
%
%\begin{figure}
%  \centering
%\begin{subfigure}[t]{0.22\textwidth}
%  \centering
%  \includegraphics[width=\textwidth]{figures/eval/generalization_jobs.pdf}
%  \caption{$15\times$ more jobs.}
%  \label{f:eval-alibaba-generalization-exec-job}
%\end{subfigure}
%\begin{subfigure}[t]{0.22\textwidth}
%  \centering
%  \includegraphics[width=\textwidth]{figures/eval/generalization_executors.pdf}
%  \caption{$10\times$ more executors.}
%  \label{f:eval-alibaba-generalization-exec}
%\end{subfigure}
%\vspace{-0.3cm}
%\caption{\name generalizes well to deployment scenarios in which the
%         workload or cluster differ from the training setting.}
%  \label{f:eval-alibaba-generalization}
%\end{figure}
%
Real-world cluster workloads vary over time, and the available cluster
machines can also change.
Ideally, \name would generalize from a model trained for a specific load and
cluster size to similar workloads with different parameters.
To test this, we train a \name agent on a scaled-down version of the
industrial workload, using 15$\times$ fewer concurrent jobs and 10$\times$
fewer executors than in the test setting.
Table~\ref{t:eval-alibaba-generalization} shows how the performance of this
agent compares with that of one trained on the real workload and cluster size.
\name is robust to changing parameters: the agent trained with 15$\times$ fewer
jobs generalizes to the test workload with a 7$\%$ reduced average JCT, and an
agent trained on a 10$\times$ smaller cluster generalizes with a 3$\%$ reduction
in average JCT.
Generalization to a larger cluster is robust as the policy correctly limits
jobs' parallelism even if vastly more resources are available.
By contrast, generalizing to a workload with many more jobs is harder, as the
smaller-scale training lacks experiences with complex job combinations.
%

%\hongzi{Generalize to changing environments and workloads: mix training set TPC-H with Alibaba,
%how easy is it to generalize from one dataset to another.}

%%%%%%%%%%%%%%%%%%%%%%%%%%%%%%%%%%%%%%%%%%%%%%%%%%%%%%%%%%%%%%%%%%%%%%%%%%%%%

\section{\name with incomplete information}
\label{s:incomplete-info}
\begin{figure}
  \centering
  \includegraphics[width=0.75\columnwidth]{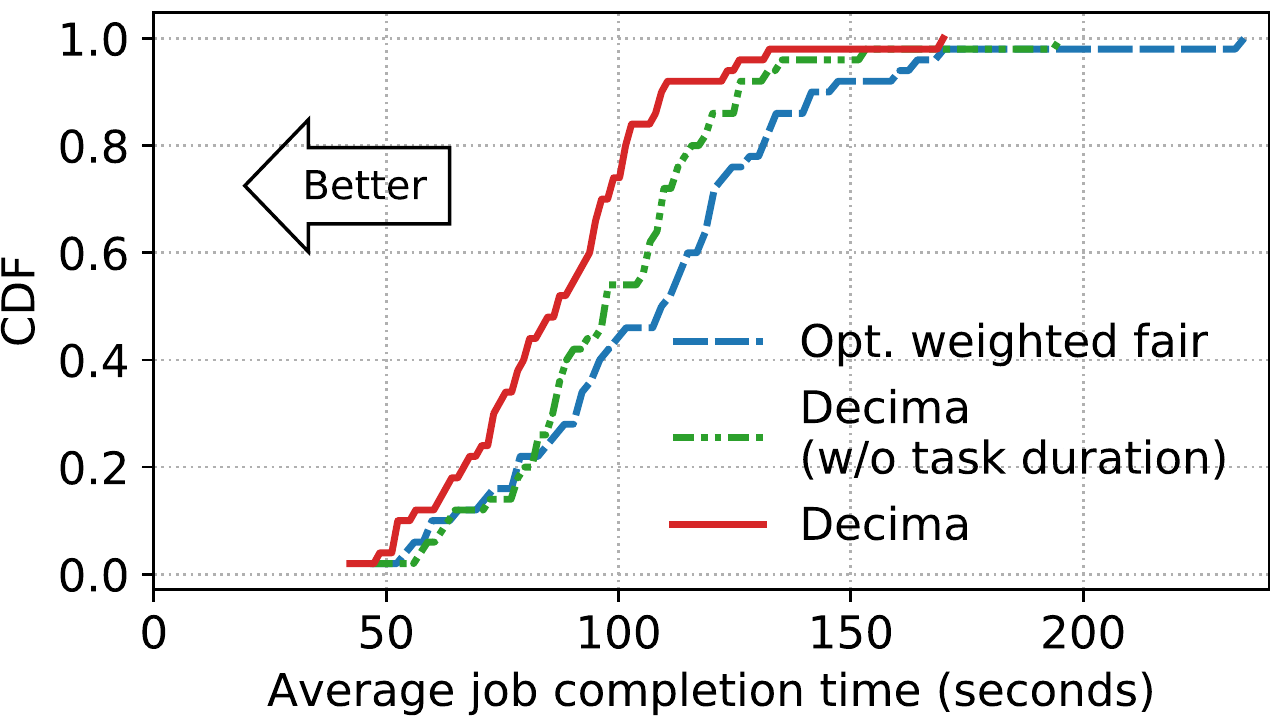}
  \vspace{-0.2cm}
  \caption{\name performs worse on unseen jobs without task duration estimates,
     but still outperforms the best heuristic.}
  \label{f:eval-omit-task-durations}
\end{figure}
In a real cluster, \name will occasionally encounter unseen jobs without
reliable task duration estimates.
Unlike heuristics that fundamentally rely on profiling information (\eg
weighted fair scheduling based on total work), \name can still work with the
remaining information and extract a reasonable scheduling policy.
Running the same setting as in~\S\ref{s:eval-real-cluster},
Figure~\ref{f:eval-omit-task-durations} shows that training without task
durations yields a policy that still outperforms the best heuristic, as
\name can still exploit the graph structure and other information such as
the correlation between number of tasks and the efficient parallelism level.

\balance

\end{document}